\documentclass{article}

\usepackage{arxiv}
\usepackage{graphicx}
\usepackage{amsmath}
\usepackage{amssymb}
\usepackage{subcaption}%
\usepackage{booktabs}

\usepackage{algorithm}
\usepackage[noend]{algpseudocode}

\usepackage{comment}
\usepackage{color}
\usepackage{amsfonts}       
\usepackage{nicefrac}       
\usepackage{microtype}      
\usepackage{xcolor,colortbl}         
\usepackage{tabularx}
\usepackage{epsfig}
\usepackage{lipsum,booktabs}

\usepackage[pagebackref,breaklinks,colorlinks]{hyperref}

\renewcommand{\headeright}{}
\renewcommand{\undertitle}{\vspace{-5ex}}
\renewcommand{\shorttitle}{A Large-scale Robustness Analysis of Video Action Recognition Models}

\hypersetup{
pdftitle={A Large-scale Robustness Analysis of Video Action Recognition Models},
pdfsubject={cs.CV},
pdfauthor={Madeline C.~Schiappa, Naman Biyani, Prudvi Kamtam,Shruti Vyas, Hamid Palangi, Vibhav Vineet, Yogesh S.~Rawat},
pdfkeywords={robustness, video robustness, action recognition, computer vision},
}

\usepackage{wrapfig,lipsum,booktabs}
\newcommand{\abs}[1]{{\color{red} {\bf} #1}}
\newcommand{\rel}[1]{{\color{blue} {\bf} #1}}

\title{
A Large-scale Robustness Analysis of Video Action Recognition Models
} 

\author{ Madeline Chantry Schiappa $^{1}$\thanks{The authors contributed equally as first authors to this paper.} \hspace{1mm}\thanks{Corresponding author:madelineschiappa@knights.ucf.edu} ,	\hspace{1mm}    Naman Biyani $^{2}$\footnotemark[1] ,	\hspace{1mm}   Prudvi Kamtam$^{1}$  \\ \textbf{Shruti Vyas} $^{1}$, \hspace{1mm}
  \textbf{Hamid Palangi} $^{3}$,
  \textbf{Vibhav Vineet} $^{3}$\thanks{The authors contributed equally as supervisors to this paper.} , \hspace{1mm}
  \hspace{1mm}\textbf{Yogesh Rawat} $^{1}$\footnotemark[3] 
 \\ CRCV, University of Central Florida $^{1}$, \hspace{2mm}
IIT Kanpur $^{2}$, \hspace{1mm} and \hspace{1mm}
   Microsoft Research $^{3}$
}  

\begin{document}
\maketitle

\begin{abstract}
        
We have seen a great progress in video action recognition in recent years. There are several models based on convolutional neural network (CNN) and some recent transformer based approaches which provide top performance on existing benchmarks. In this work, we perform a \textbf{large-scale robustness analysis} of these existing models for video action recognition. We focus on robustness against \textbf{real-world distribution shift} perturbations instead of adversarial perturbations. We propose \textbf{four} different benchmark datasets, \textbf{HMDB51-P}, \textbf{UCF101-P}, \textbf{Kinetics400-P}, and \textbf{SSv2-P} to perform this analysis. We study robustness of \textbf{six} state-of-the-art action recognition models against \textbf{90} different perturbations. The study reveals some interesting findings, 1) \textbf{transformer} based models are consistently \textbf{more robust} compared to CNN based models, 
2) \textbf{Pretraining improves robustness} for Transformer based models more than CNN based models, and 3) All of the studied models are \textbf{robust to temporal perturbations} for all datasets but SSv2; suggesting the importance of temporal information for action recognition varies based on the dataset and activities.
Next, we study the role of augmentations in model robustness and present a real-world dataset, \textbf{UCF101-DS}, which contains realistic distribution shifts, to further validate some of these findings. We believe this study will serve as a benchmark for future research in robust video action recognition \footnote[1]{More details available at \url{bit.ly/3TJLMUF}.}.
\vspace{-10pt}

\end{abstract}

    \section{Introduction}
    \label{sec:intro}
\begin{wrapfigure}[18]{r}{0.5\textwidth}
\vspace{-20pt}
\begin{center}
\includegraphics[width=0.9\linewidth]{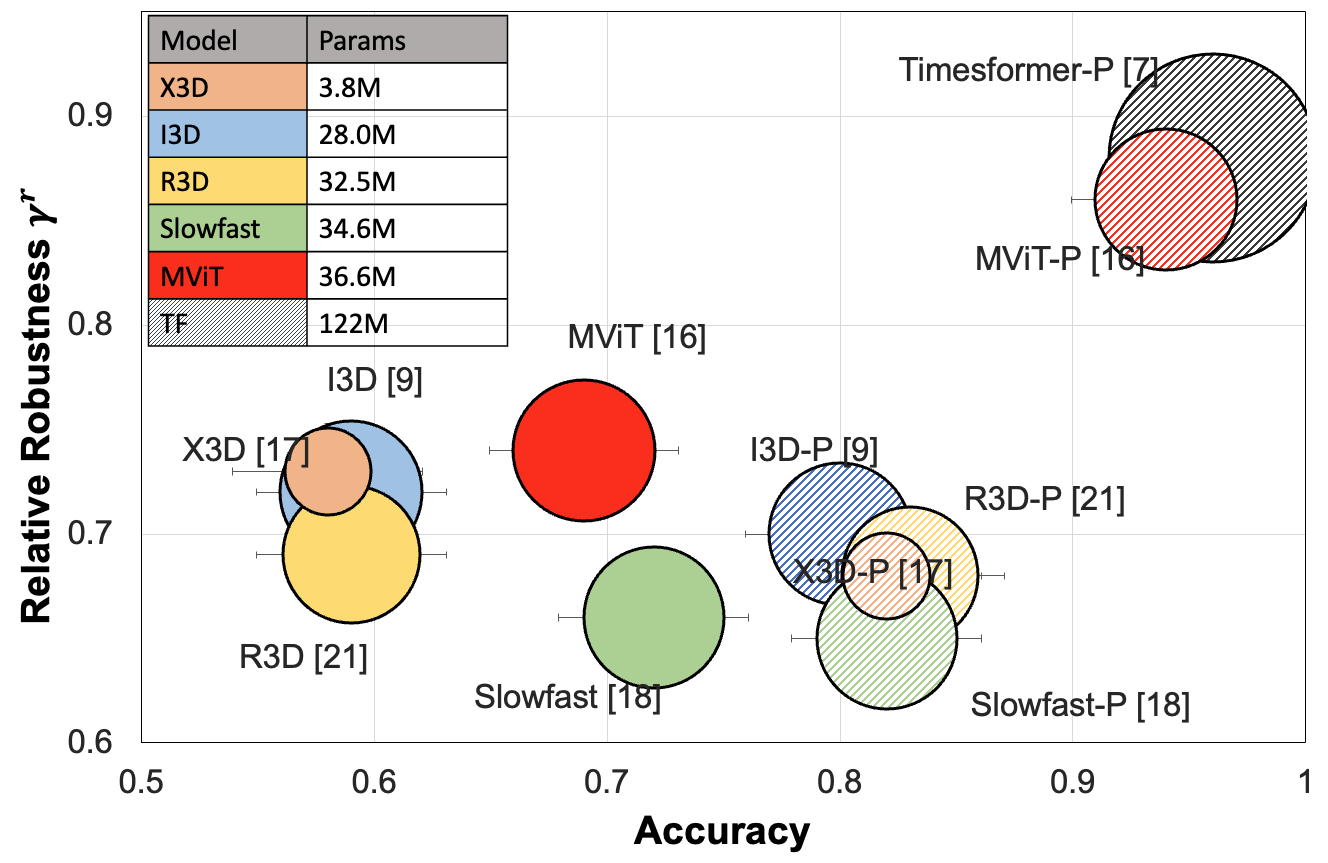}

\end{center}
  \caption{Performance against robustness  of action recognition models on UCF101-P. y-axis: relative robustness $\gamma^r$, x-axis: accuracy on clean videos,  model names appended with P indicate pre-training, and the size of circle indicates FLOPs. 
  }
\label{fig:model_capacity}
\end{wrapfigure}

Robustness of deep learning models against real-world distribution shifts is crucial for various applications in vision, such as medicine \cite{ardulov2021robust},
 autonomous driving \cite{ma2018improved}, environment monitoring \cite{ullo2020advances}, conversational systems \cite{li2021neural}, robotics \cite{yim2007modular} and assistive technologies \cite{asan2020artificial}. 
Distribution shifts with respect to training data can occur due to the variations in environment such as changes in geographical locations, background, lighting, camera models, object scale, orientations, motion patterns, etc. Such distribution shifts can cause the models to fail when deployed in a real world settings \cite{hendrycks2018benchmarking}. 
For example, an AI ball tracker that replaced human camera operators was recently deployed in a soccer game and repeatedly confused a soccer ball with the bald head of a lineman, leading to a bad experience for viewers \cite{soccer_fail}.

Robustness has been an active research topic due to its importance for real-world applications \cite{ullo2020advances,ardulov2021robust,ma2018improved}. However, most of the effort is directed towards images \cite{hendrycks2018benchmarking,bhojanapalli2021understanding,hendrycks2021many}. Video is a natural form of input to the vision systems that function in the real world. Therefore studying robustness in videos is an important step towards developing reliable systems
for real world deployment. In this work we perform a large-scale analysis on robustness of existing deep models for video action recognition against common real world spatial and temporal distribution shifts. 

Video action recognition provides an important test scenario to study robustness in videos given there are sufficient, large benchmark datasets and well developed deep learning models.
Although the existing approaches have made impressive progress in action recognition, there are several fundamental questions that still remain unanswered in the field. Do these approaches enable effective temporal modeling, the crux of the matter for action recognition approaches? 
Are these approaches robust to real-world corruptions like noise (temporally consistent and inconsistent), blurring effects, etc? 
Do we really need heavy architectures for robustness, or are light-weight models good enough? 
%
%
Are the recently introduced transformer-based models, which give state of the art accuracy on video datasets, more robust? Does pretraining play a role in model robustness? This study aims at finding answers to some of these critical questions.

Towards this goal, we present multiple benchmark datasets to conduct robustness analysis in video action recognition. We utilize four different widely used action recognition datasets including HMDB51 \cite{Kuehne11}, UCF101 \cite{soomro2012ucf101}, Kinetics-400\cite{carreira2017quo}, and SSv2 \cite{mahdisoltani2018effectiveness} and propose four corresponding benchmarks; \textit{HMDB51-P}, \textit{UCF101-P}, \textit{Kinetics400-P}, and \textit{SSv2-P}.
In order to create this benchmark, we introduce 90 different common perturbations which include, 20 different noise corruptions, 15 blur perturbations, 15 digital perturbations, 25 temporal perturbations, and 15 camera motion perturbations as a benchmark. The study covers 6 different deep architectures considering different aspects, such as network size (small vs large), network architecture (CNN vs Transformers), and network depth (shallow vs deep). 

This study reveals several interesting findings about action recognition models. We observe that recent \textit{transformer based models are not only better in performance}, but they are also \textit{more robust than CNN models} against most distribution shifts (Figure \ref{fig:model_capacity}). We also observe that \textit{pretraining is more beneficial to transformers compared to CNN} based models in robustness. We find that all the models are very robust against the temporal perturbations with minor drop in performance on the Kinetics, UCF and HMDB. However, on the SSv2 dataset, behavior of the models is different whereas the performance drops on different temporal perturbations. These observations show interesting phenomena about the video action recognition datasets, i.e., \textit{the importance of temporal information varies based on the dataset and activities}.

Next, we study the role of training with data augmentations in model robustness and analyze the generalization of these techniques to novel perturbations. 
To further study the capability of such techniques, we propose a real-world dataset, \textit{UCF101-DS}, which contains realistic distribution shifts without simulation. This dataset also helps us to better understand the behavior of CNN and Transformer-based models under realistic scenarios. We believe such findings will open up many interesting research directions in video action recognition and will facilitate future research on video robustness which will lead to more robust architectures for real-world deployment. 

We make the following contributions in this study, 

\begin{itemize}
\setlength\itemsep{-.15em}
\item A large-scale robustness analysis of video action recognition models to different real-world distribution shifts.

\item 
Provide insights including comparison of transformer vs CNN based models, effect of pre-training, and effect of temporal perturbations on video robustness.

\item Four large-scale benchmark datasets to study robustness for video action recognition along with a real-world dataset with realistic distribution shifts.

\end{itemize}

\section{Related work}
\label{sec:related_work}

\subsection{Action recognition}

Video understanding has made rapid progress with the
introduction of a number of large-scale video datasets such
as Kinetics \cite{carreira2017quo}, Sports1M \cite{KarpathyCVPR14}, Moments-In-Time \cite{monfort2019moments} , SSv2 \cite{goyal2017something}
and YouTube-8M \cite{abu2016youtube}. A number of recent models have emphasized the need to efficiently model spatio-temporal information for video action recognition. Some early approaches, inspired by image classification models \cite{krizhevsky2012imagenet}, utilize 2D-CNN models \cite{KarpathyCVPR14} for video classification. Some recent works  \cite{wang2016temporal,zhou2018temporal,lin2019tsm} have proposed effective ways to integrate image level features for video understanding.
The success of 2D convolution has inspired many 3D convolution based approaches for recognizing actions in videos \cite{chen2021deep,huang2018makes}. For example, C3D\cite{tran2015learning} learns 3D ConvNets, outperforming 2D CNNs through the use of large-scale video datasets. Many variants of 3D-CNNs
are introduced for learning spatio-temporal features such as
I3D  \cite{carreira2018quo} and ResNet3D \cite{hara2017learning}. 3D CNN features were also
demonstrated to generalize well to other vision tasks \cite{tang2020asynchronous,duarte2018videocapsulenet,aafaq2019spatio,yang2018exploring,chao2018rethinking,yao2016highlight}. Because 3D CNN based approaches lead to higher computational load, recent
works aim to reduce the complexity by decomposing the
3D convolution into 2D and 1D convolutions \cite{qiu2017learning,xie2018rethinking,tran2018closer}, or
incorporating group convolution \cite{luo2019grouped}; or using a combination of 2D and 3D-CNN \cite{chen2021deep}. Furthermore, SlowFast \cite{feichtenhofer2019slowfast}
network employs two pathways to capture short-term and
long-term temporal information by processing a video
at both slow and fast frame rates. 

Recently, transformer based models have shown remarkable success in various vision tasks, such as image classification, after the introduction of Vision Transformer (ViT)~\cite{dosovitskiy2021image}. The impressive performance led to using transformer-based architectures for video domains. Video transformers have led to state-of-the-art performance on Kinetics-400 \cite{carreira2017quo}, SSv2 \cite{goyal2017something} and Charades \cite{sigurdsson2016hollywood}. Specific to video, a temporal attention encoder was added on top of ViT, further improving performance on action recognition \cite{neimark2021video}. More recently, MViT\cite{fan2021multiscale} was proposed; a multi-scale vision transformer for video recognition that achieved top results on SSv2. A factorized spacetime attention based approach was proposed in Timesformer \cite{bertasius2021spacetime} after analysis of various variants of space-time attention based on compute-accuracy tradeoff. Video Swin Transformer~\cite{liu2021video} investigated spatiotemporal locality and showed that an inductive bias of locality is a better speed-accuracy trade-off compared to using global self-attention. 
We use both CNN-based and recent transformer-based architectures to study their robustness for action recognition. 

\subsection{Robustness}
Many recent works on robustness in the vision community are focused on adversarial attacks, where a computed perturbation is deliberately added to the input sample \cite{akhtar2018threat,zhang2021adversarial}. Different from adversarial attacks, the real-world distribution shifts in data naturally emerge from different scenarios. Some of the recent works are focused towards understanding the robustness of existing methods in the image domain against these distribution shifts \cite{hendrycks2018benchmarking,bhojanapalli2021understanding,hendrycks2021many,sakaridis2021acdc}. In \cite{hendrycks2018benchmarking}, the authors analyzed different image classification models for different corruptions in ImageNet. Similarly, in \cite{recht2019imagenet} the authors presented a new benchmark of naturally occurring distribution shifts using ImageNet and studied the robustness of different image models. 
In a recent study \cite{naseer2021intriguing}, the image based transformer models were found to be more robust towards different kinds of perturbations. The benchmark in \cite{taori2019robustness} analyzed natural robustness and demonstrated that data augmentation is not sufficient to improve model robustness. 

Some recent works have further explored the use of data augmentation to improve the robustness of image models \cite{geirhos2018imagenet,hendrycks2019augmix,yin2019fourier}. Data augmentations such as various noise types \cite{madry2018towards,rusak2020increasing,lopes2019improving}, transformations \cite{geirhos2018imagenet,yun2019cutmix}, and compositions of these simple transformations \cite{cubuk2019autoaugment,hendrycks2019augmix} are shown to be helpful in improving the robustness of deep networks. These robustness studies are mainly focused on images. There are a few works addressing the issue of adversarial robustness in videos \cite{wu2020robustness} and analyzing importance of temporal aspect in videos \cite{sevilla2021only,tclr}. Different from these existing works, this work provides a large-scale benchmark on robustness of video action recognition models against real-world perturbations. 

In a recent effort \cite{yi2021benchmarking}, an initial analysis on robustness against natural distribution shift was presented for videos extending visual augmentations \cite{hendrycks2018benchmarking}. This work was focused on compression specific perturbations including: bit rate, compression, frame rate, and packet loss. Different from this work, we emphasize on temporal perturbations that are not limited to compression. Moreover, this study a small scale benchmark focusing on subsets of Kinetics \cite{kinetics} and SSv2 \cite{stst}. In comparison, our analysis uses the full Kinetics and SSv2 dataset while additionally analyzing models on UCF101 \cite{ucf101} and HMDB51 \cite{hmdb}, the most common action recognition evaluation datasets. As a result, our findings differ from their initial findings, e.g. model capacity and robustness or generalization when trained on perturbations.

\section{Distribution shifts}
\label{sec:corruption}

Existing research in action recognition is mostly focused on training and testing the proposed methods on a benchmark dataset with little to no distribution shift from training to testing samples. In most of the real-world applications, we observe different types of distribution shifts in testing environments before deployment, affecting the performance of the models. To help circumvent this issue, it is important to study robustness of existing deep learning based video action recognition models against real-world perturbations, i.e., they are not artificially created using adversarial attacks and happen naturally for example due to change in environment, different camera settings, etc. Towards this goal, we designed a set of perturbations which are frequently encountered in real-world environments. 
Existing datasets on action recognition do not focus on such distribution shifts and therefore it is important to construct a benchmark that covers a wide range of distribution shifts which will be beneficial for the community. We study five different categories of real-world perturbations which include, \textit{noise}, \textit{blur}, \textit{digital}, \textit{temporal}, and \textit{camera motion}.

%
%




\label{sec_per}



\vspace{-4mm}
\paragraph{\textbf{Noise:}}
We define 4 categories for noise; \textit{Gaussian}, \textit{Shot}, \textit{Impulse}, and \textit{Speckle} noise. \textit{Gaussian noise} can appear due to low-lighting conditions. \textit{Shot noise} tries to capture the electronic noise caused by the discrete nature of light. We use Poisson distribution to approximate it. \textit{Impulse noise} tries to simulate corruptions caused by bit errors and is analogous to salt-and-pepper noise. \textit{Speckle noise} is additive noise where noise added is proportional to the pixel intensity. 





\vspace{-4mm}
\paragraph{\textbf{Blur:}}
We define three kinds of perturbations for blur effect; \textit{Zoom}, \textit{Motion}, and \textit{Defocus}. \textit{Zoom blur} occurs when the camera moves toward an object rapidly. \textit{Motion blur} appears due to the destabilizing motion of camera. Finally, \textit{Defocus blur} may happen when the camera is out of focus.

\vspace{-3mm}
\paragraph{\textbf{Digital:}}
Recent years have seen a sharp increase in video traffic. In fact, video
content consumption increased so much during the initial
months of the pandemic that content providers like Netflix
and Youtube were forced to throttle video-streaming quality
to cope with the surge. Hence efficient video compression to reduce bandwidth consumption without compromising on quality is more critical than ever. We evaluate the models on JPEG and two other video encoding codecs and analyse the drop in accuracy due to these compression methods. \textit{JPEG} is a lossy image compression format which introduces compression artifacts. \textit{MPEG1} is designed to compress raw digital video without excessive quality loss and is used in a large number of products and technologies. \textit{MPEG2} is an enhanced version of MPEG1 and is also a lossy compression for videos which is used in transmission and various other applications.



\begin{figure}[t!]
\begin{center}
    \includegraphics[width=.7\textwidth]{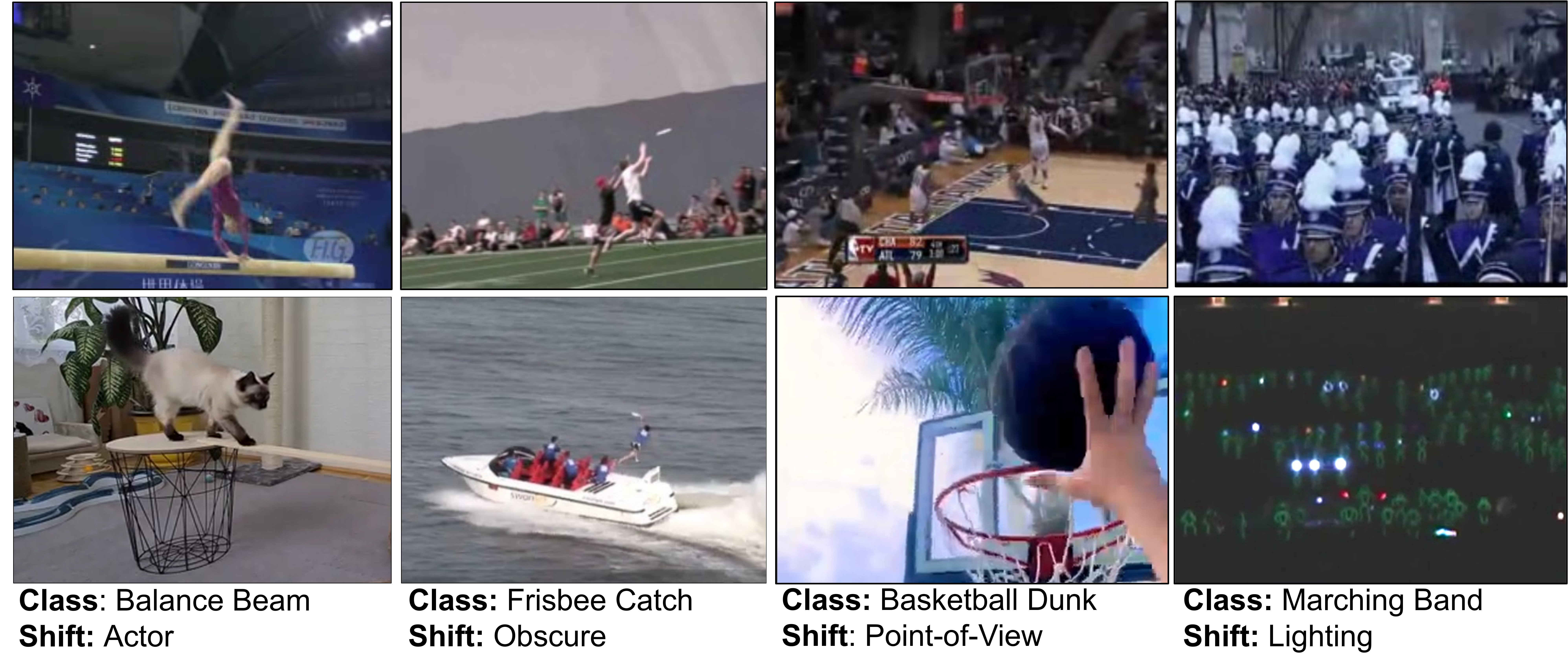}
\end{center}
  \caption{Sample video frames from proposed UCF101-DS dataset (bottom) compared to UCF101 (top) for 5 classes and 5 variations.}
\label{fig:teaser_dataset}
\end{figure}

\vspace{-3mm}
\paragraph{\textbf{Temporal:}}
Although CNN-based approaches have made impressive
progress in action recognition, one of the major questions that still remain unanswered is whether these approaches enable more effective temporal modeling, the crux of the matter for action recognition? How different are the recent Transformers from 3D-CNN based approaches as far as temporal modelling of video data is concerned?
To compare the approaches on effective temporal modelling, we define five different temporal perturbations; \textit{Sampling rate}, \textit{Reversal}, 
\textit{Jumbling}, \textit{Box jumbling}, and \textit{Freezing}. \textit{Sampling rate} evaluates the models against varying skip-frame rates. \textit{Reversal} perturbation reverse the video frames with varying skip-frame rates. 
\textit{Jumbling} shuffles the frames in a segment-wise fashion. We utilize frame index permutation for 5 different segment sizes (4,8,16,32,64). In \textit{Box jumbling}, we shuffle the segments instead of frames inside those segments. \textit{Freezing} perturbation freeze video frames randomly and tries to capture the issues with video buffering.

\vspace{-10pt}
\paragraph{\textbf{Camera motion:}}
To compare the approaches for robustness in the presence of irregularities due to camera motion, we define three  perturbations; \textit{Static rotation}, \textit{Dynamic rotation}, and \textit{Translation}. \textit{Static rotation} uses a constant rotation angle for all the video frames. It captures effects due to tilted camera orientation.
\textit{Random rotation} rotates each frame by a varying random angle. It captures effects due to changing camera angle.
\textit{Translation} randomly crops a video frame with varying crop location across time. This is introduced to capture the random shaking motion of camera.

\vspace{-10pt}
\paragraph{\textbf{Severity level}}
The natural perturbations may occur in videos at different severity levels depending on the environment/situation. Therefore, it is important to study the effect of these perturbations at different severity levels. We generate 5 levels from 1-5 where 1 refers to minimal distribution shift and 5 refers to a large distribution shift.
We apply the proposed perturbations at every severity level on all the testing videos of the benchmark and save it for a consistent evaluation. 
More details about the implementation of these perturbations are provided in the supplementary. 

\section{Model variants}
\label{sec:arch}
    
\begin{wraptable}{r}{.5\linewidth}
\vspace{-20pt}
\setlength\tabcolsep{1.5pt}
\centering
\caption{Details of action recognition models used in this study. 
}
\label{table_model}
\resizebox{\linewidth}{!}{
\begin{tabular}{l|r|r|r|r|r|r}
\hline
\multicolumn{1}{l|}{Model}  & \multicolumn{1}{c|}{R3D \cite{hara2017learning}} & \multicolumn{1}{c|}{I3D \cite{carreira2017quo}} & \multicolumn{1}{c|}{SF \cite{feichtenhofer2019slowfast}} & \multicolumn{1}{c|}{X3D \cite{feichtenhofer2020x3d}} & \multicolumn{1}{c|}{MViT \cite{fan2021multiscale}} & \multicolumn{1}{c}{TF \cite{bertasius2021spacetime}} \\ \hline \hline
Params&  32.5M  & 28.0M &  34.6M &   3.8M     &  36.6M   &     122M \\
FLOPs &  55.1G&  75.1G &  66.6G &        5.15G&    70.7G     &196G      \\
\begin{tabular}[c]{@{}l@{}}\# of frames\end{tabular}   &8        & 8   &        32     &      16  &16     &8      \\
Frame rate &    8    & 8   &2&        5&    4 & 32    \\
\hline
\end{tabular}
}
\end{wraptable} 

We perform our experiments on six different action recognition models which are based on CNN and Transformer architectures. The goal is to benchmark multiple backbones and simultaneously study the behavior of CNN and Transformer based models for robustness in video action recognition. We evaluate three most popular CNN-based action recognition models which are known to perform well, not only in action recognition, but also serve as fundamental building blocks for many other problems in the video domain. These include I3D \cite{carreira2017quo}, ResNet3D (R3D) \cite{hara2017learning}, and SlowFast (SF) \cite{feichtenhofer2019slowfast}. Among these, I3D and R3D are based on 3D convolutions but differ in the backbones, where I3D uses Inception-V1 and R3D uses a ResNet backbone. Slowfast is one of the best action recognition models and is based on a 3D-CNN, which can use any backbone in its two stream approach. We use a R3D 
backbone for both slow as well as the fast branch. We also evaluate X3D, an efficient CNN model \cite{feichtenhofer2020x3d} that attempts to optimize the network size and its complexity.
Recently, Transformer based models have shown a great success in various vision-based tasks \cite{khan2021transformers}. Several models have been proposed for video representation learning \cite{neimark2021video,bertasius2021spacetime,fan2021multiscale,liu2021video}. We use the top two Transformer based models in this study, including Timesformer (TF) \cite{bertasius2021spacetime} and MViT \cite{fan2021multiscale}. Timesformer utilizes a factorized space-time attention whereas MViT uses pooling attention for efficient computation. More details are shown in Table \ref{table_model}.

\section{Robustness benchmarks and evaluation}
\label{sec:benchmark}

\paragraph{\textbf{Datasets}}
We use four action recognition benchmark datasets for our experiments including UCF101 \cite{soomro2012ucf101}, HMDB51 \cite{Kuehne11}, Kinetics-400 \cite{kay2017kinetics}, and SSv2 \cite{mahdisoltani2018effectiveness}. \textbf{UCF101} is an action recognition dataset with 101 action classes. There are a total of 13K videos, with around 100 videos per class. The length of videos in this dataset ranges from 4-10 seconds. \textbf{HMDB51} has 7K videos with 51 classes. For each action, at least 70 videos are for training and 30 videos are for testing. \textbf{SSv2} is a large collection of videos with focus on humans performing basic actions with everyday objects. There are 174 classes and it contains 220,847 videos, with 168,913 in the training set, 24,777 in the validation set and 27,157 in the test set. 
\textbf{Kinetics-400} is another large-scale action recognition benchmark dataset with 400 classes. Each action category has at least 400 videos and each video clip last around 10 seconds. It covers a broad range of action classes including human-object interactions and human-human interactions.

We apply the proposed 90 perturbations to the test set of these datasets to create robustness benchmarks which we refer to as \textbf{HMDB51-P}, \textbf{UCF101-P}, \textbf{Kinetics400-P}, and \textbf{SSv2-P}. HMDB51-P consists of 137,610 videos, UCF101-P consists of 340,380 videos, Kinetics400-P consists of 1,616,670 videos, and SSv2-P consists of 2,229,930 videos. These benchmarks are not used for training.

We additionally propose a new dataset that focuses on real-world distribution shifts, UCF101 Distribution Shift (UCF101-DS). For classes in the UCF101 dataset, we collected videos of uncommon or isolated variations for a number of distribution shifts that are categorized into higher-level groups such as: ``style'', ``lighting'', ``scenery'', ``actor'', ``occlusion''. More details about this dataset can be found in the supplementary. Some examples of these variations are in Fig. \ref{fig:teaser_dataset}. We have a total of 63 distribution shifts organized into 15 categories for 47 classes for a total of 4,708 clips.

\begin{figure*}[t!]
\begin{center}
\includegraphics[width=\linewidth]{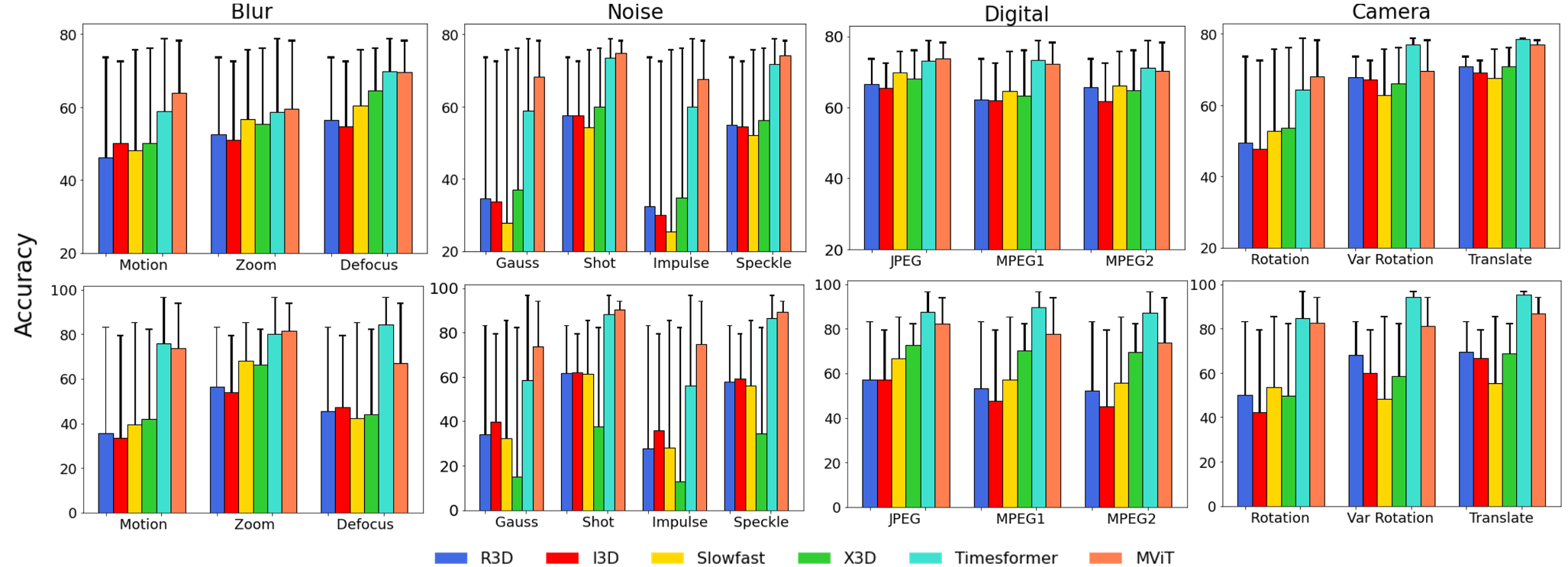}
\end{center}
  \caption{
  Robustness analysis of different action recognition models on the Kinetics-P (top row) and UCF101-P (bottom row)
  benchmark for various perturbations. Each bar plot corresponds to one category of perturbations showing performance drop of each model. The bar shows accuracy on perturbed dataset and the extension indicates performance drop from accuracy on clean data.}
  \vspace{-2mm}
\label{fig:all_perturbation_kinetics_ucf}
\end{figure*}

\vspace{-10pt}
\paragraph{\textbf{Implementation details}}

We train R3D, I3D, SlowFast, X3D, and MViT models for HMDB51 and UCF101 with and without pre-trained weights. The pre-trained weights from Kinetics-400 are used to initialize for the first variation. Furthermore, we consider I3D, Slowfast, X3D and Timesformer models for evaluation on SSv2 dataset since pretrained weights for these four models are publicly available. 
We use the official implementations available with pre-trained weights with the same experimental setup as described in these works. More details in Table \ref{table_model}.



\vspace{-10pt}
\paragraph{\textbf{Evaluation protocol}}
To ensure fair comparison and facilitate reproduciblity,
we evaluate all the models under similar protocol. 
We use clips with a resolution of 224×224 for all the datasets. For evaluation, in Kinetics dataset, we follow the protocol of taking 10 uniform temporal crops for each video 
and applying center crop for each of these 10 crops. The videos in UCF101 and HMDB51 are shorter in comparison to Kinetics-400, so we take 5 uniform temporal crops for each video and apply center crop for each clip.
For UCF101 and HMDB51, we also evaluated models when they are pre-trained on a large-scale dataset, such as Kinetics-400, before finetuning on these smaller datasets. For SSv2 we used a single spatial crop and uniformly sampled the number of frames as used in the original model implementation. 

\vspace{-2mm}
\paragraph{\textbf{Evaluation metrics}}

To measure robustness, we use two metrics; one for absolute accuracy drop and the other for relative accuracy drop. If we have a trained model $f$, we first compute the accuracy $A^f_c$ on the clean test set. Next, we test this classifier on a perturbation $p$ at each of the severity levels $s$, and obtain accuracy $A^f_{p,s}$ for perturbation $p$ and severity $s$. The absolute robustness $\gamma^a$ is computed for each severity level $s$ and perturbation $p$ as $\gamma^a_{p,s} = 1- (A^f_c - A^f_{p,s})/100$. 
%
The aggregated performance of a model can be obtained by averaging all severity levels to get $\gamma^a_{p}$ and over all perturbations to get $\gamma^a$.
Different models provide varying performance on the same test videos and therefore absolute drop in performance will also depend on the models performance on clean videos. To take this into account, we compute relative performance drop to measure models robustness. The relative robustness $\gamma^r$ is computed for each severity level $s$ and perturbation $p$ as $\gamma^r_{p,s} = 1 - (A^f_c - A^f_{p,s})/A^f_c$ which is the difference normalized to the accuracy of the model on the test set without perturbation. 

\section{Experiments}
\label{sec:experiments}

\begin{figure*}
    \centering
    \includegraphics[width=\textwidth]{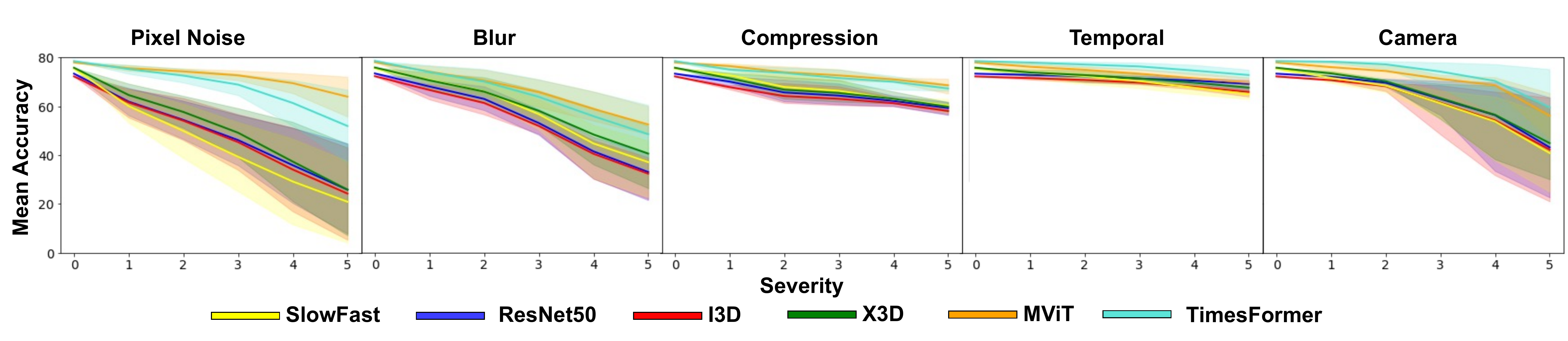}
    \caption{The mean performance on Kinetics-400P across perturbation types and severity for all models. 
    }
    \label{fig:all_severities}
\end{figure*}

\begin{table}[t!]
\setlength{\tabcolsep}{2pt}
\footnotesize
\centering
\caption{$\gamma^a$ and $\gamma^r$ robustness scores of the models on Kinetic-400P benchmark dataset. For both, higher is better. 
}
\resizebox{.6\linewidth}{!}{
\begin{tabular}{l|l|l|l|l|l|l|l|l|l|l|l|l}
\hline
     & \multicolumn{2}{c|}{\textbf{Noise}}& \multicolumn{2}{c|}{\textbf{Blur}}     & \multicolumn{2}{c|}{\textbf{Temporal}} & \multicolumn{2}{c|}{\textbf{Digital}} & \multicolumn{2}{c}{\textbf{Camera}}  & \multicolumn{2}{c}{\textbf{Mean}} \\ \hline
\textbf{Network} & $\gamma^a$ & $\gamma^r$ & $\gamma^a$ & $\gamma^r$ & $\gamma^a$ & $\gamma^r$ & $\gamma^a$ & $\gamma^r$ & $\gamma^a$ & $\gamma^r$& $\gamma^a$ & $\gamma^r$ \\
\hline \hline
R3D         &  .71&.61     & .78&.70  &    \textbf{.98}&\textbf{.97}      &  .91&.88       &  .89&.85     & .85 & .80\\
\hline
I3D         &  .72&.61  & .80&.72 &     .97&.96     &  .91&.87       &    .89&.85   &.86 & .80\\
\hline
SF    &   .64&.53    & .80&.73  &    .95&.93      &  .91&.89       &    .86&.81   &.83 &.78 \\
\hline
X3D         &  .71&.62     & .81&.75 &    .96&.94    &  .90&.86       &    .88&.84  & .85&  .80\\
\hline
TF &  .87&.84     & .84&.79 &      .97&.94    &  \textbf{.94}&.92       &   \textbf{.95}&\textbf{.93} &.91 &.88
\\
\hline
MViT        &   \textbf{.93} &\textbf{.91}    & \textbf{.86}&\textbf{.82}  &    .96&.95      & \textbf{.94}&\textbf{.93}        &   .94&.92  & \textbf{.93}&  \textbf{.91} \\
\hline
\end{tabular}
}
\label{table:summary1}
\end{table}

\vspace{-5pt}
We analyze robustness of models against 5 different kinds of perturbations and what that means for model behavior on the UCF101-P, Kinetics-P, HMDB51-P and SSv2-P. A summary of model robustness across severities and perturbation categories is shown in Figures \ref{fig:all_severities} and \ref{fig:all_perturbation_kinetics_ucf} and Tables \ref{table:summary1} and \ref{table:summary2}.

\begin{figure*}
    \centering
    \includegraphics[width=\textwidth]{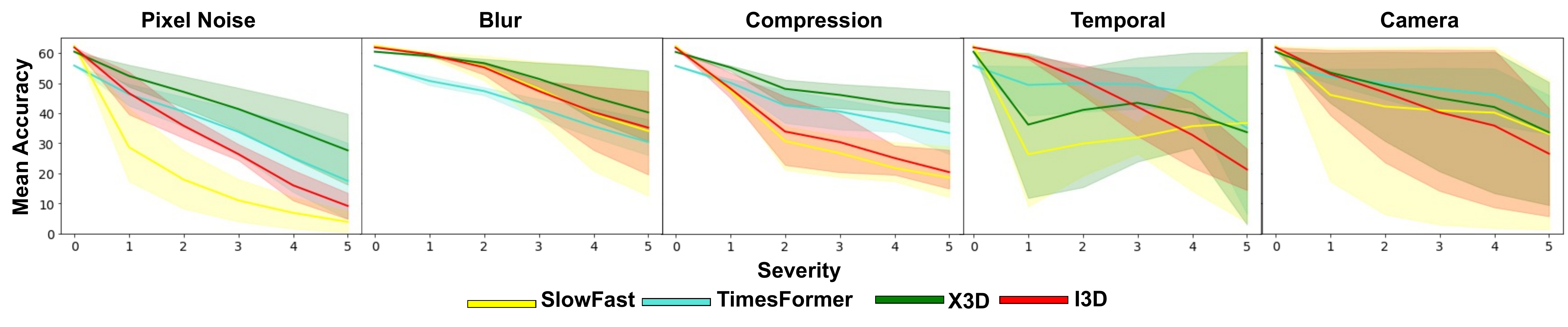}
    \caption{The mean performance on SSV2-P across perturbation types and severity for all models.
    }
    \label{fig:acc_sev_ssv2}
\end{figure*}

\begin{table}[t!]
\footnotesize
\setlength{\tabcolsep}{2pt}
\caption{$\gamma^a$ and $\gamma^r$ robustness scores of the models on SSv2P benchmark dataset. For both, higher is better.}
\centering
\resizebox{.6\linewidth}{!}{
\begin{tabular}{l|l|l|l|l|l|l|l|l|l|l|l|l}

\hline
     & \multicolumn{2}{c|}{\textbf{Noise}}& \multicolumn{2}{c|}{\textbf{Blur}}     & \multicolumn{2}{c|}{\textbf{Temporal}} & \multicolumn{2}{c|}{\textbf{Digital}} & \multicolumn{2}{c}{\textbf{Camera}}  & \multicolumn{2}{c}{\textbf{Mean}} \\ \hline
\textbf{Network} & $\gamma^a$ & $\gamma^r$ & $\gamma^a$ & $\gamma^r$ & $\gamma^a$ & $\gamma^r$ & $\gamma^a$ & $\gamma^r$ & $\gamma^a$ & $\gamma^r$ & $\gamma^a$ & $\gamma^r$ \\
\hline \hline
I3D        &  .63&.40  & .85&.76 &     .69&.50     &  .69&.51       &    .78&.65 & .78&.64\\
\hline
SF    &   .51&.22    & .85&.76  &    .68&.48      &  .67&.48       &    .74&.59  & .75&.58  \\
\hline
X3d         &  \textbf{.80}&\textbf{.67}     & \textbf{.90}&\textbf{.67} &    .77&.61    &  \textbf{.86}&\textbf{.78}       &    .80&.67   &    .85&.76 \\
\hline
TF &  .78&.59     & .85&.74 &      \textbf{.89}&\textbf{.78}    &  .85&.73       &   \textbf{.88}&\textbf{.78}
&   \textbf{.87}&\textbf{.77}\\
\hline
\end{tabular}
}
\label{table:summary2}
\end{table}

\begin{figure*}[t!]
    \centering
    \includegraphics[width=0.95\textwidth]{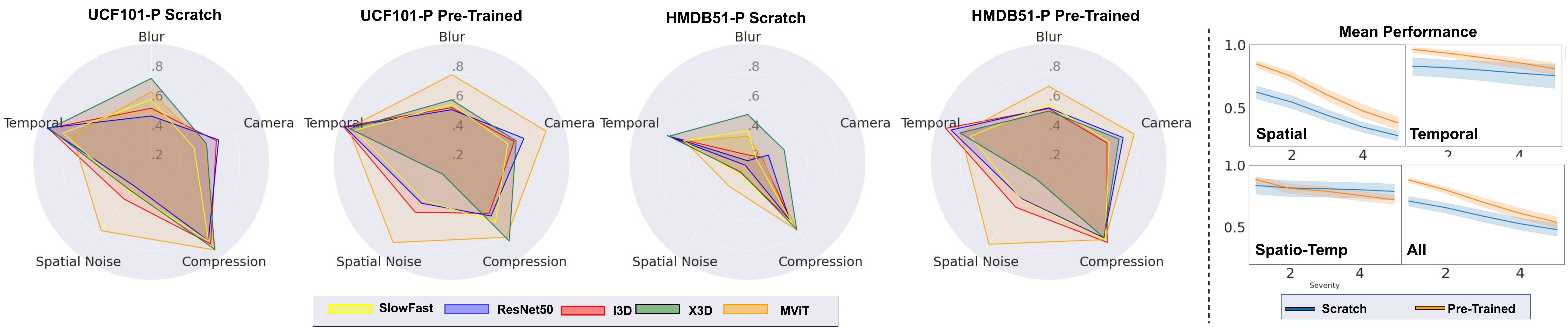}
    \caption{Left: Mean accuracy for each perturbation comparing pre-trained models to models trained from scratch on UCF101-P and HMDB51-P.
    Right: Mean accuracy across models for temporal, spatial and spatio-temporal perturbations on HMDB51-P and UCF101-P. 
    }
    \label{fig:pretrain_vs_scratch}
    \vspace{-10pt}
\end{figure*}

\subsection{Robustness analysis}
\paragraph{Spatial} 
Here we focus on \textit{Noise}, \textit{Camera} and \textit{Blur} perturbations. In Figure \ref{fig:all_severities} we observe for Kinetics-P that spatial perturbations have the largest drop in performance as severity increases. For all three categories, we see that the transformer-based Timesformer and MViT models are typically more robust than CNN-based models. For example, performance of Timesformer and ResNet based R3D drops by $\sim$5\% and $\sim$30\% respectively. In Figure \ref{fig:all_perturbation_kinetics_ucf}, surprisingly models are more robust to variable rotation compared to a static rotation. This may be because randomly rotating may provide some frames closer to the expected but if the static rotation is far from the expected, performance drops. Behavior on SSv2 data is similar, however, in Figure \ref{fig:acc_sev_ssv2} we observe that MViT and Timesformer models are typically less robust than X3D. This may indicate that with a more temporal-specific dataset, the CNN-based models are more robust. In summary, \textit{all models struggle with spatial-based perturbations and the Transformer-based architectures are typically more robust than CNN-based architectures.} 

\paragraph{Temporal}
\begin{wrapfigure}[14]{r}{0.5\textwidth}
\vspace{-6pt}
\includegraphics[width=\linewidth]{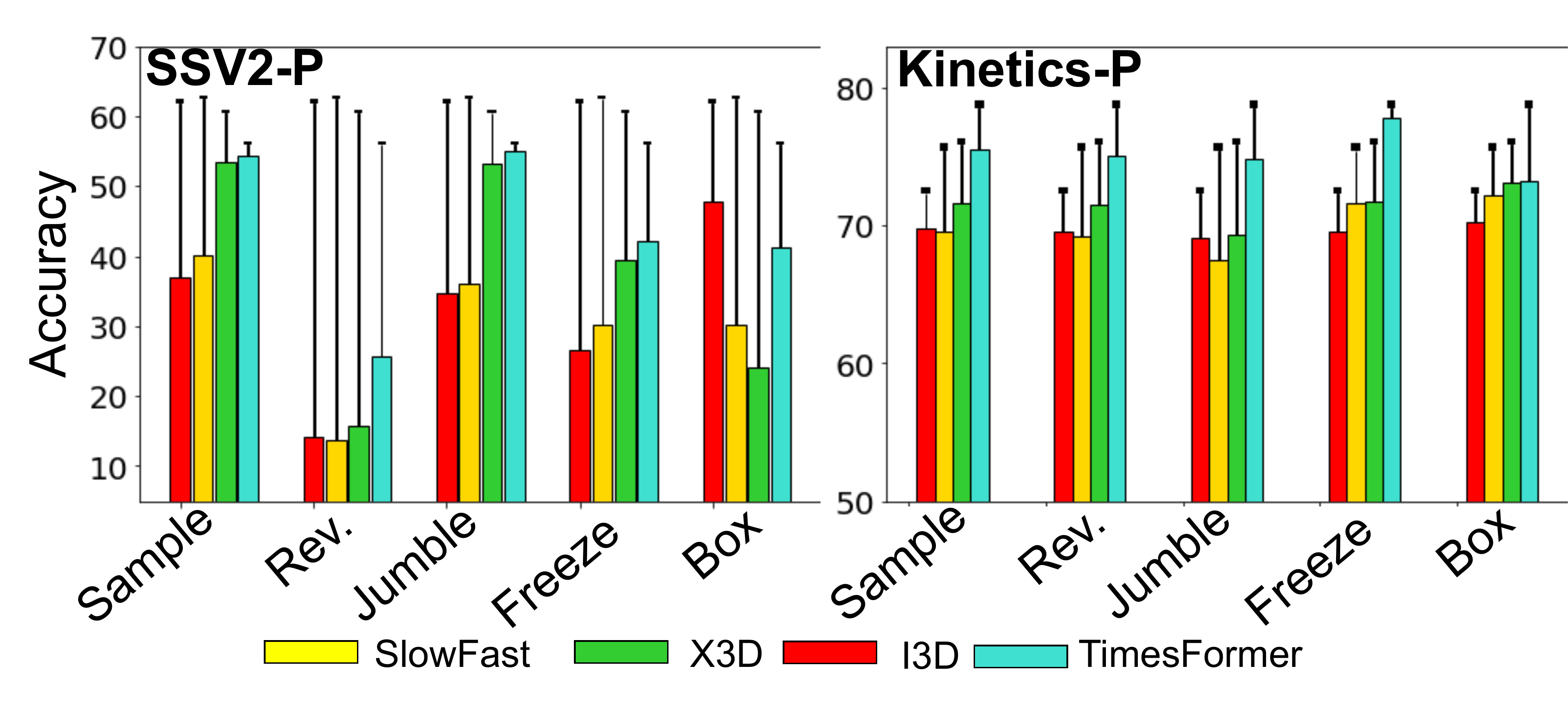} 
  \caption{
  Robustness analysis for temporal perturbations. Each bar shows accuracy on clean data and drop on perturbed benchmarks.}
\label{fig:utd_generated}
\end{wrapfigure}

To study the effect of temporal perturbation on videos, we perform experiments after applying the different types of perturbations: \textit{jumbling}, \textit{box jumbling}, \textit{jumbling}, \textit{sampling}, \textit{reverse sampling} and \textit{freezing} of frames. The results are presented in Fig. \ref{fig:all_severities}, \ref{fig:acc_sev_ssv2}, and \ref{fig:utd_generated}. To our surprise, we observed different behaviors on different datasets. Models are typically robust on the UCF101-P, Kinetics-P, and HMDB51-P datasets while not robust on the SSv2 dataset.

\begin{figure*}[t!]
    \centering
    \includegraphics[width=\linewidth]{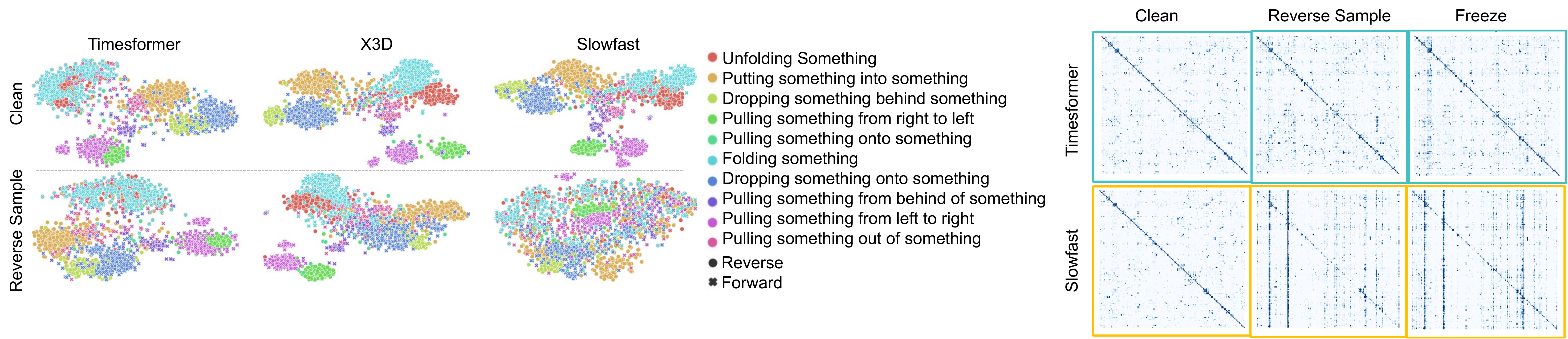}
    \caption{Comparing clean to temporally perturbed videos. Left: we visualize the feature space of a subset of classes that have a reverse activity class for SSv2. 
    Right: The confusion matrices of SSv2 classes for different perturbations at severity 4. 
    }
    \label{fig:ssv2_opposing_classes}
\end{figure*}

In order to gain further insights in their behaviors, we visualize features of CNN and transformer models using t-SNE \cite{vanDerMaaten2008} features. In Fig. \ref{fig:ssv2_opposing_classes} we visualize t-SNE features of Timesformer, X3D and Slowfast models under reverse temporal perturbation for 5 action classes and their respective opposite classes in the arrow of time from SSv2. We observe that CNN-based, X3D and Slowfast, models confuse between classes, but Timesformer model clusters different classes properly even at high severity levels. To understand class confusion further, Fig. \ref{fig:ssv2_opposing_classes}  also visualizes a confusion matrix of SSv2 classes for freeze and reverse sampling between Timesformer and Slowfast. We see a noticeable different between transformer-based model and CNN-based model on over-predictions, which are visible by the dark vertical lines. This again indicates transformers may be more robust to temporal perturbations. 

These observations provide an interesting phenomena about the action recognition datasets. Firstly, \textit{temporal information is more important for action recognition on the SSv2}, where activities can often be reversed and become a different activity. Secondly, \textit{temporal learning may not be required for shorter clips} that do not have any potential of a reversal of activities. We believe such findings will open up many interesting research directions in video action recognition.

\vspace{-3mm}
\paragraph{{Spatio-Temporal.}} Here we focus on \textit{Compression} perturbations which affect both spatial and temporal signals in a video. In Figure \ref{fig:all_perturbation_kinetics_ucf} and \ref{fig:all_severities}, we observe that models are typically robust to these perturbations but struggle more on UCF101-P. For UCF101-P and HMDB51-P, we do see that the transformer-based models are typically more robust. On SSv2-P, we observe that models struggle more with compression than compared to the other datasets (Figure \ref{fig:acc_sev_ssv2}). This further indicates that SSv2-P requires more temporal learning compared to the other datasets, and therefore models struggle when temporal perturbations are present.  


\subsection{Effect of pretraining on robustness}

We conducted experiments on the UCF101-P and HMDB51-P benchmarks, where models are pretrained on the Kinetics-400 dataset. 
The mean relative robustness scores across perturbation categories are shown in Fig. \ref{fig:pretrain_vs_scratch} where the closer to the center, the less robust. A breakdown of the results for each perturbation type is shown in the Supplementary for both datasets. Overall,\textit{ we observe that pretraining models results in higher robustness}. We also observe that \textit{the relative benefit of pretraining is more evident in Transformer models compared to CNN models} (Figure \ref{fig:model_capacity}). 
%
%



\subsection{Model capacity vs. robustness}
To understand how model capacity might impact robustness, we compare accuracy and relative robustness $\gamma^r$ in Figure \ref{fig:model_capacity}. Models trained from scratch are solid colors while those pre-trained are dashed. The size of the dot for each model is based on the model capacity as shown in Table \ref{table_model}. While most of the models have around 35M parameters, the X3D \cite{feichtenhofer2020x3d} is significantly smaller and is still comparatively robust when compared to other models. When comparing the pre-trained MViT and Timesformer, we again see that a model with significantly less parameters is just as accurate and robust as one with significantly more parameters. In contrast to the findings in \cite{yi2021benchmarking}, our analysis \textit{indicates that high model capacity does not necessarily mean more robustness}.


\subsection{Augmentations for robustness}
In this experiment we study the role of augmentations on model robustness. We explore the use of perturbations as augmentation and analyze both CNN and Transformer based models. We also experiment with PixMix \cite{hendrycks2022pixmix}, which is one of the recent approach for robust model learning. We use some perturbations for training and keep others for evaluation. Similarly, we use severity of 1,2, and 3 for training and  4 or 5 for testing. For PixMix \cite{hendrycks2022pixmix}, we apply the augmentation at severity 3 for each frame individually, in which a different fractal image is chosen for each. 

\begin{figure*}[t!]
    \centering
   \raisebox{-0.5\height}{\includegraphics[width=.65\linewidth]{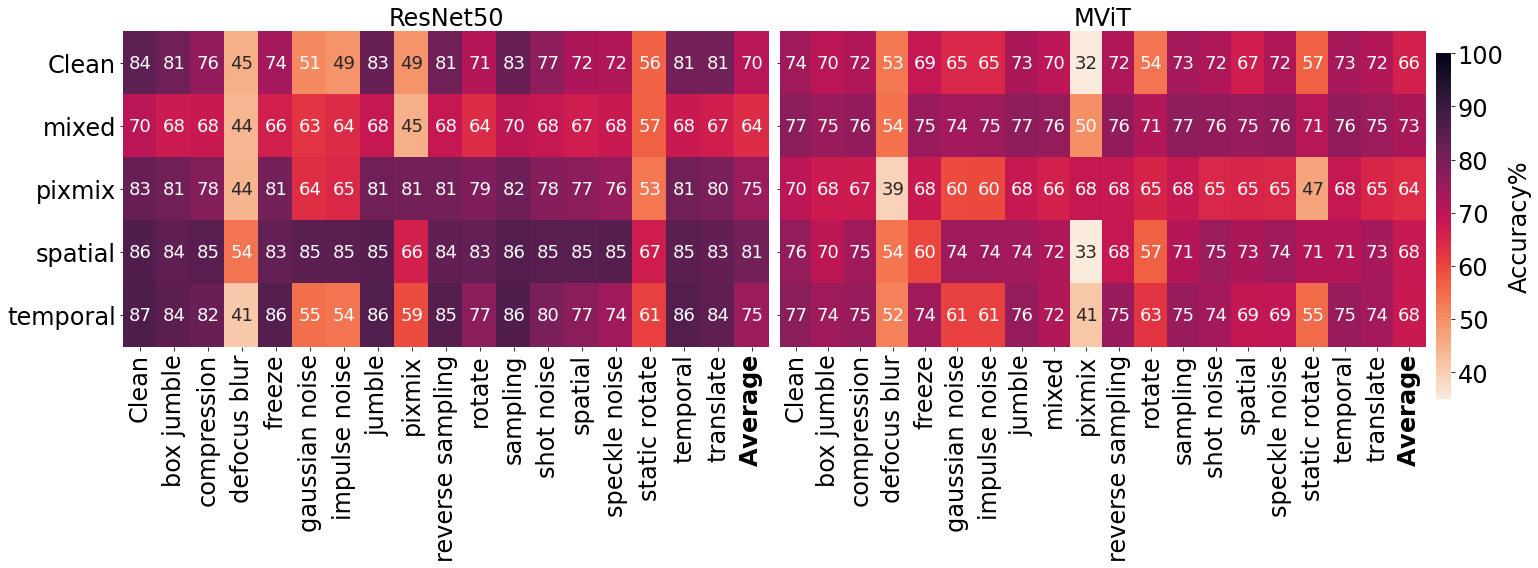}}
    \raisebox{-0.5\height}{\includegraphics[width=.33\linewidth]{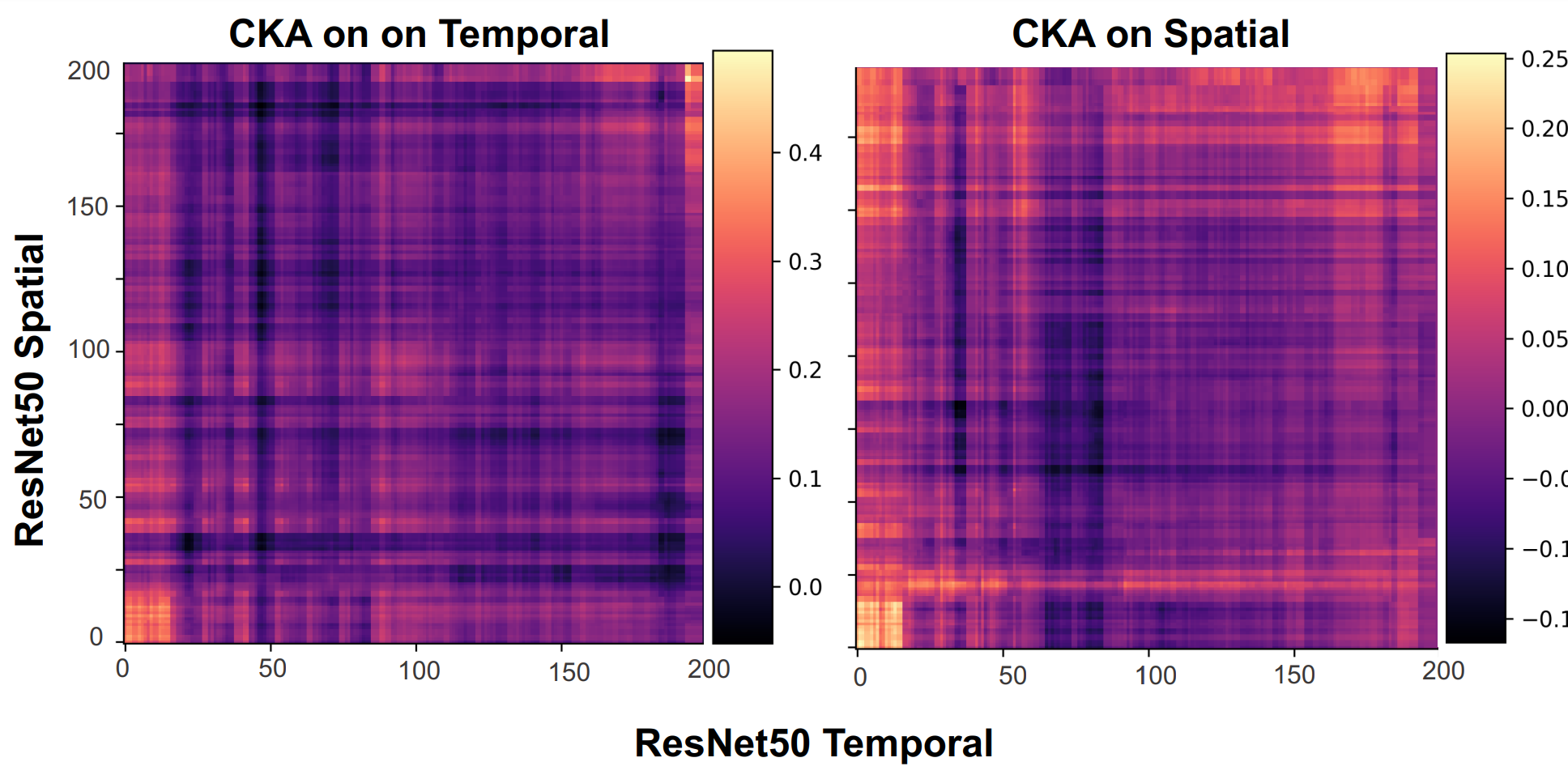}}
    \vspace{-10pt}
    \caption{Left: Accuracy for each perturbation (x-axis) compared to what data models were trained on (y-axis). We compare a CNN to a transformer model when trained on clean data or different combinations of perturbations. 
    Right: A heatmap of CKA values \cite{nguyen2020wide} for ResNet50 when trained and tested on either spatial or temporal perturbations.}
    \label{fig:robustness_training}
    \vspace{-10pt}
\end{figure*}
The overall results for these experiments are shown in Figure \ref{fig:robustness_training}. We observe that certain perturbations may be more beneficial for different architectures. ResNet50 becomes less robust when trained on a mix of perturbations but is more robust when trained on Spatial and PixMix. To understand changes to the networks when trained on perturbed data, we use CKA \cite{nguyen2020wide} to compare layer activations for the ResNet50 model on different perturbed data. Fig \ref{fig:robustness_training} shows a comparison between a model trained on temporal versus spatial perturbations when evaluated on UCF101-P for temporal or spatial perturbations. We find both variations of the ResNet50 are more similar for temporal perturbations compared to spatial based on the resulting scales. Both variations are also more similar at the initial layers, where most changes are in the middle or final layers. Our results indicate that \textit{the CNN-based models may benefit more from spatial perturbations during training than transformer-based models}. 

\subsection{Robustness analysis on real-world videos}

\begin{wrapfigure}[14]{r}{0.6\textwidth}
    \centering
    \includegraphics[width=\linewidth]{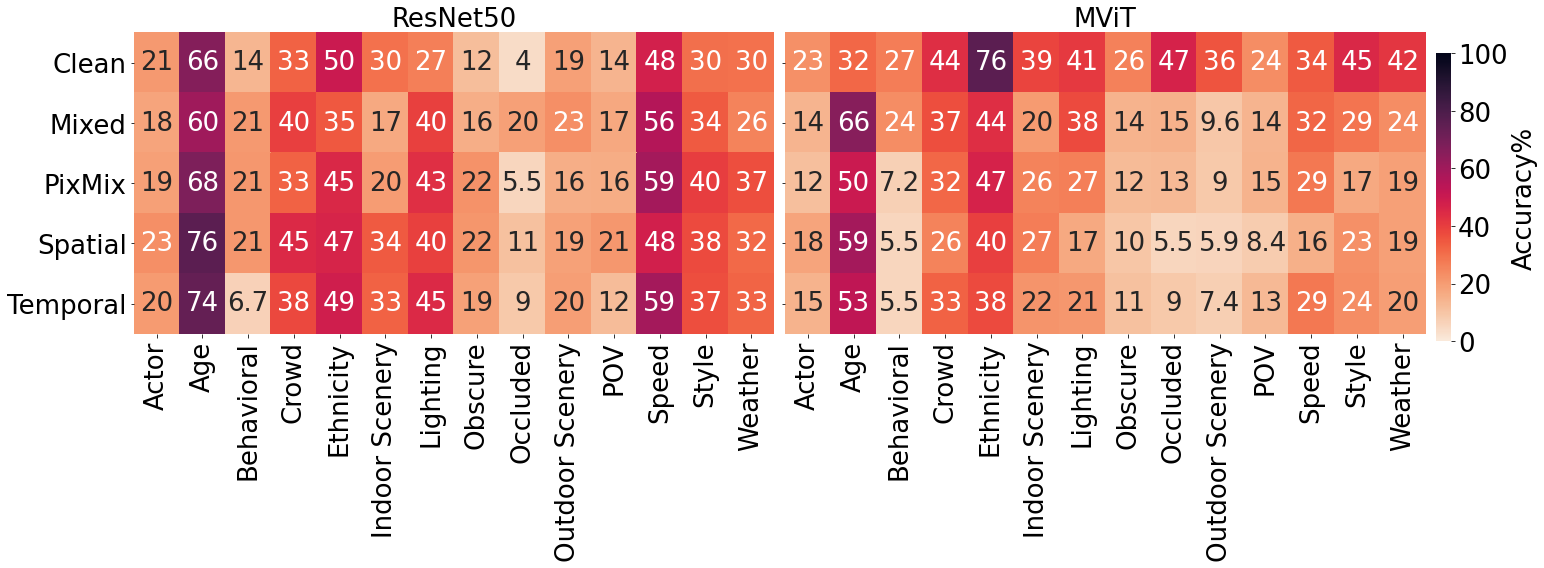}
    \caption{Overall results on our proposed UCF101-DS dataset.}
    \label{fig:ucf101DS_results}
\end{wrapfigure}

To better understand model behavior under natural distribution shifts, we evaluate the CNN-based model ResNet50 and the Transformer-based model MViT on UCF101-DS. The results are shown in Figure \ref{fig:ucf101DS_results}. When trained on UCF101, The MViT model is typically more robust to UCF101-DS compared to the ResNet50. MViT is more robust to ethnicity variations, occlusion, and changes in scene while ResNet50 is more robust to natural variations in play speed and age variations. Similar to our findings on UCF101-P in Figure \ref{fig:robustness_training}, we find that training on perturbed data is less beneficial for MVIT compared to ResNet50. When not trained on UCF101-P, we find the MViT model is more robust to natural distribution shifts. However, when trained on UCF101-P, ResNet50 becomes more robust than MViT. \textit{This further supports that training transformer-based models on perturbed data may not benefit robustness while it does on CNN-based approaches}. The results also indicate that these models are not typically robust to natural distribution shifts.

\section{Discussion and conclusion}
\label{sec:analysis}
We have conducted a large-scale robustnesss analysis on standard CNN and Transformer based action recognition models. We created benchmark datasets based on Kinetics400, SSv2, UCF101 and HMDB51. 
We proposed a new dataset, UCF101-DS, that captures real-world distribution shifts in areas like scenery, point-of-view and more. 
Our study provides the following initial insights:
\begin{itemize}
\setlength\itemsep{-.2em}
\item Transformer models 
are generally more robust 
than CNN. 
\item Pre-training improves robustness for transformer-based models more than CNN-based models.
\item Training on perturbed data benefits CNN-based models more than Transformer-based models.
\item More parameters do not necessarily mean robustness.
\item 
Unlike the other datasets studied in this benchmark, SSv2, with its reversible actions, requires temporal learning.
\item Like what is seen with images, models are not robust to spatial noise but unlike with images, they are \textit{sometimes} robust to temporal noise.

\end{itemize}

  
This study presented a benchmark for robustness of video models against real-world distribution shifts. The findings and the benchmark in this work can potentially open up  interesting questions about robustness of video action recognition models. The benchmark introduced in this study will be released publicly at \url{bit.ly/3TJLMUF}.

\section{Acknowledgement}
This research is based upon work supported by the Office of the Director of National Intelligence (IARPA) via 2022-21102100001. The views and conclusions contained herein are those of the authors and should not be interpreted as necessarily representing the official policies, either expressed or implied, of ODNI, IARPA, or the US Government. The US Government is authorized to reproduce and distribute reprints for governmental purposes notwithstanding any copyright annotation therein.

{\small
\bibliographystyle{plain}
\bibliography{egbib}

\begin{thebibliography}{10}

\bibitem{aafaq2019spatio}
Nayyer Aafaq, Naveed Akhtar, Wei Liu, Syed~Zulqarnain Gilani, and Ajmal Mian.
\newblock Spatio-temporal dynamics and semantic attribute enriched visual
  encoding for video captioning.
\newblock In {\em Proceedings of the IEEE/CVF Conference on Computer Vision and
  Pattern Recognition}, pages 12487--12496, 2019.

\bibitem{abu2016youtube}
Sami Abu-El-Haija, Nisarg Kothari, Joonseok Lee, Paul Natsev, George Toderici,
  Balakrishnan Varadarajan, and Sudheendra Vijayanarasimhan.
\newblock Youtube-8m: A large-scale video classification benchmark.
\newblock {\em arXiv preprint arXiv:1609.08675}, 2016.

\bibitem{akhtar2018threat}
Naveed Akhtar and Ajmal Mian.
\newblock Threat of adversarial attacks on deep learning in computer vision: A
  survey.
\newblock {\em Ieee Access}, 6:14410--14430, 2018.

\bibitem{ardulov2021robust}
Victor Ardulov, Victor~R Martinez, Krishna Somandepalli, Shuting Zheng, Emma
  Salzman, Catherine Lord, Somer Bishop, and Shrikanth Narayanan.
\newblock Robust diagnostic classification via q-learning.
\newblock {\em Scientific reports}, 11(1):1--9, 2021.

\bibitem{asan2020artificial}
Onur Asan, Alparslan~Emrah Bayrak, Avishek Choudhury, et~al.
\newblock Artificial intelligence and human trust in healthcare: focus on
  clinicians.
\newblock {\em Journal of medical Internet research}, 22(6):e15154, 2020.

\bibitem{bertasius2021spacetime}
Gedas Bertasius, Heng Wang, and Lorenzo Torresani.
\newblock Is space-time attention all you need for video understanding?, 2021.

\bibitem{bhojanapalli2021understanding}
Srinadh Bhojanapalli, Ayan Chakrabarti, Daniel Glasner, Daliang Li, Thomas
  Unterthiner, and Andreas Veit.
\newblock Understanding robustness of transformers for image classification.
\newblock {\em arXiv preprint arXiv:2103.14586}, 2021.

\bibitem{carreira2017quo}
Joao Carreira and Andrew Zisserman.
\newblock Quo vadis, action recognition? a new model and the kinetics dataset.
\newblock In {\em proceedings of the IEEE Conference on Computer Vision and
  Pattern Recognition}, pages 6299--6308, 2017.

\bibitem{kinetics}
Joao Carreira and Andrew Zisserman.
\newblock Quo vadis, action recognition? a new model and the kinetics dataset.
\newblock In {\em proceedings of the IEEE Conference on Computer Vision and
  Pattern Recognition}, pages 6299--6308, 2017.

\bibitem{carreira2018quo}
Joao Carreira and Andrew Zisserman.
\newblock Quo vadis, action recognition? a new model and the kinetics dataset,
  2018.

\bibitem{chao2018rethinking}
Yu-Wei Chao, Sudheendra Vijayanarasimhan, Bryan Seybold, David~A Ross, Jia
  Deng, and Rahul Sukthankar.
\newblock Rethinking the faster r-cnn architecture for temporal action
  localization.
\newblock In {\em Proceedings of the IEEE Conference on Computer Vision and
  Pattern Recognition}, pages 1130--1139, 2018.

\bibitem{chen2021deep}
Chun-Fu~Richard Chen, Rameswar Panda, Kandan Ramakrishnan, Rogerio Feris, John
  Cohn, Aude Oliva, and Quanfu Fan.
\newblock Deep analysis of cnn-based spatio-temporal representations for action
  recognition.
\newblock In {\em Proceedings of the IEEE/CVF Conference on Computer Vision and
  Pattern Recognition}, pages 6165--6175, 2021.

\bibitem{soccer_fail}
Cohen Coberly.
\newblock Ai failure in real-world application.
\newblock {\em Techspot}, 2020.
\newblock
  \url{https://www.techspot.com/news/87431-ai-powered-camera-zooms-bald-head-instead-soccer.html}.
  [Accessed: Oct 18, 2021].

\bibitem{cubuk2019autoaugment}
Ekin~Dogus Cubuk, Barret Zoph, Dandelion Mane, Vijay Vasudevan, and Quoc~V Le.
\newblock Autoaugment: Learning augmentation policies from data.
\newblock 2019.

\bibitem{tclr}
Ishan Dave, Rohit Gupta, Mamshad~Nayeem Rizve, and Mubarak Shah.
\newblock Tclr: Temporal contrastive learning for video representation.
\newblock {\em Computer Vision and Image Understanding}, page 103406, 2022.

\bibitem{dosovitskiy2021image}
Alexey Dosovitskiy, Lucas Beyer, Alexander Kolesnikov, Dirk Weissenborn,
  Xiaohua Zhai, Thomas Unterthiner, Mostafa Dehghani, Matthias Minderer, Georg
  Heigold, Sylvain Gelly, Jakob Uszkoreit, and Neil Houlsby.
\newblock An image is worth 16x16 words: Transformers for image recognition at
  scale, 2021.

\bibitem{duarte2018videocapsulenet}
Kevin Duarte, Yogesh Rawat, and Mubarak Shah.
\newblock Videocapsulenet: A simplified network for action detection.
\newblock {\em Advances in Neural Information Processing Systems},
  31:7610--7619, 2018.

\bibitem{fan2021multiscale}
Haoqi Fan, Bo~Xiong, Karttikeya Mangalam, Yanghao Li, Zhicheng Yan, Jitendra
  Malik, and Christoph Feichtenhofer.
\newblock Multiscale vision transformers, 2021.

\bibitem{feichtenhofer2020x3d}
Christoph Feichtenhofer.
\newblock X3d: Expanding architectures for efficient video recognition.
\newblock In {\em Proceedings of the IEEE/CVF Conference on Computer Vision and
  Pattern Recognition}, pages 203--213, 2020.

\bibitem{feichtenhofer2019slowfast}
Christoph Feichtenhofer, Haoqi Fan, Jitendra Malik, and Kaiming He.
\newblock Slowfast networks for video recognition, 2019.

\bibitem{geirhos2018imagenet}
Robert Geirhos, Patricia Rubisch, Claudio Michaelis, Matthias Bethge, Felix~A
  Wichmann, and Wieland Brendel.
\newblock Imagenet-trained cnns are biased towards texture; increasing shape
  bias improves accuracy and robustness.
\newblock In {\em International Conference on Learning Representations}, 2018.

\bibitem{stst}
Raghav Goyal, Samira Ebrahimi~Kahou, Vincent Michalski, Joanna Materzynska,
  Susanne Westphal, Heuna Kim, Valentin Haenel, Ingo Fruend, Peter Yianilos,
  Moritz Mueller-Freitag, et~al.
\newblock The" something something" video database for learning and evaluating
  visual common sense.
\newblock In {\em Proceedings of the IEEE international conference on computer
  vision}, pages 5842--5850, 2017.

\bibitem{goyal2017something}
Raghav Goyal, Samira~Ebrahimi Kahou, Vincent Michalski, Joanna Materzyńska,
  Susanne Westphal, Heuna Kim, Valentin Haenel, Ingo Fruend, Peter Yianilos,
  Moritz Mueller-Freitag, Florian Hoppe, Christian Thurau, Ingo Bax, and Roland
  Memisevic.
\newblock The "something something" video database for learning and evaluating
  visual common sense, 2017.

\bibitem{hara2017learning}
Kensho Hara, Hirokatsu Kataoka, and Yutaka Satoh.
\newblock Learning spatio-temporal features with 3d residual networks for
  action recognition, 2017.

\bibitem{hendrycks2021many}
Dan Hendrycks, Steven Basart, Norman Mu, Saurav Kadavath, Frank Wang, Evan
  Dorundo, Rahul Desai, Tyler Zhu, Samyak Parajuli, Mike Guo, et~al.
\newblock The many faces of robustness: A critical analysis of
  out-of-distribution generalization.
\newblock In {\em Proceedings of the IEEE/CVF International Conference on
  Computer Vision}, pages 8340--8349, 2021.

\bibitem{hendrycks2018benchmarking}
Dan Hendrycks and Thomas Dietterich.
\newblock Benchmarking neural network robustness to common corruptions and
  perturbations.
\newblock In {\em International Conference on Learning Representations}, 2018.

\bibitem{hendrycks2019augmix}
Dan Hendrycks, Norman Mu, Ekin~Dogus Cubuk, Barret Zoph, Justin Gilmer, and
  Balaji Lakshminarayanan.
\newblock Augmix: A simple data processing method to improve robustness and
  uncertainty.
\newblock In {\em International Conference on Learning Representations}, 2019.

\bibitem{hendrycks2022pixmix}
Dan Hendrycks, Andy Zou, Mantas Mazeika, Leonard Tang, Bo~Li, Dawn Song, and
  Jacob Steinhardt.
\newblock Pixmix: Dreamlike pictures comprehensively improve safety measures.
\newblock In {\em Proceedings of the IEEE/CVF Conference on Computer Vision and
  Pattern Recognition}, pages 16783--16792, 2022.

\bibitem{huang2018makes}
De-An Huang, Vignesh Ramanathan, Dhruv Mahajan, Lorenzo Torresani, Manohar
  Paluri, Li~Fei-Fei, and Juan~Carlos Niebles.
\newblock What makes a video a video: Analyzing temporal information in video
  understanding models and datasets.
\newblock In {\em Proceedings of the IEEE Conference on Computer Vision and
  Pattern Recognition}, pages 7366--7375, 2018.

\bibitem{KarpathyCVPR14}
Andrej Karpathy, George Toderici, Sanketh Shetty, Thomas Leung, Rahul
  Sukthankar, and Li~Fei-Fei.
\newblock Large-scale video classification with convolutional neural networks.
\newblock In {\em CVPR}, 2014.

\bibitem{kay2017kinetics}
Will Kay, Joao Carreira, Karen Simonyan, Brian Zhang, Chloe Hillier, Sudheendra
  Vijayanarasimhan, Fabio Viola, Tim Green, Trevor Back, Paul Natsev, Mustafa
  Suleyman, and Andrew Zisserman.
\newblock The kinetics human action video dataset, 2017.

\bibitem{khan2021transformers}
Salman Khan, Muzammal Naseer, Munawar Hayat, Syed~Waqas Zamir, Fahad~Shahbaz
  Khan, and Mubarak Shah.
\newblock Transformers in vision: A survey.
\newblock {\em arXiv preprint arXiv:2101.01169}, 2021.

\bibitem{krizhevsky2012imagenet}
Alex Krizhevsky, Ilya Sutskever, and Geoffrey~E Hinton.
\newblock Imagenet classification with deep convolutional neural networks.
\newblock {\em Advances in neural information processing systems},
  25:1097--1105, 2012.

\bibitem{Kuehne11}
H.~Kuehne, H.~Jhuang, E.~Garrote, T.~Poggio, and T.~Serre.
\newblock {HMDB}: a large video database for human motion recognition.
\newblock In {\em Proceedings of the International Conference on Computer
  Vision (ICCV)}, 2011.

\bibitem{hmdb}
H.~Kuehne, H.~Jhuang, E.~Garrote, T.~Poggio, and T.~Serre.
\newblock {HMDB}: a large video database for human motion recognition.
\newblock In {\em Proceedings of the International Conference on Computer
  Vision (ICCV)}, 2011.

\bibitem{li2021neural}
Han Li, Sunghyun Park, Aswarth Dara, Jinseok Nam, Sungjin Lee, Young-Bum Kim,
  Spyros Matsoukas, and Ruhi Sarikaya.
\newblock Neural model robustness for skill routing in large-scale
  conversational ai systems: A design choice exploration.
\newblock {\em arXiv preprint arXiv:2103.03373}, 2021.

\bibitem{lin2019tsm}
Ji~Lin, Chuang Gan, and Song Han.
\newblock Tsm: Temporal shift module for efficient video understanding.
\newblock In {\em Proceedings of the IEEE/CVF International Conference on
  Computer Vision}, pages 7083--7093, 2019.

\bibitem{liu2021video}
Ze~Liu, Jia Ning, Yue Cao, Yixuan Wei, Zheng Zhang, Stephen Lin, and Han Hu.
\newblock Video swin transformer, 2021.

\bibitem{lopes2019improving}
Raphael~Gontijo Lopes, Dong Yin, Ben Poole, Justin Gilmer, and Ekin~D Cubuk.
\newblock Improving robustness without sacrificing accuracy with patch gaussian
  augmentation.
\newblock 2019.

\bibitem{luo2019grouped}
Chenxu Luo and Alan~L Yuille.
\newblock Grouped spatial-temporal aggregation for efficient action
  recognition.
\newblock In {\em Proceedings of the IEEE/CVF International Conference on
  Computer Vision}, pages 5512--5521, 2019.

\bibitem{ma2018improved}
Xiaobai Ma, Katherine Driggs-Campbell, and Mykel~J Kochenderfer.
\newblock Improved robustness and safety for autonomous vehicle control with
  adversarial reinforcement learning.
\newblock In {\em 2018 IEEE Intelligent Vehicles Symposium (IV)}, pages
  1665--1671. IEEE, 2018.

\bibitem{madry2018towards}
Aleksander Madry, Aleksandar Makelov, Ludwig Schmidt, Dimitris Tsipras, and
  Adrian Vladu.
\newblock Towards deep learning models resistant to adversarial attacks.
\newblock In {\em International Conference on Learning Representations}, 2018.

\bibitem{mahdisoltani2018effectiveness}
Farzaneh Mahdisoltani, Guillaume Berger, Waseem Gharbieh, David Fleet, and
  Roland Memisevic.
\newblock On the effectiveness of task granularity for transfer learning.
\newblock {\em arXiv preprint arXiv:1804.09235}, 2018.

\bibitem{monfort2019moments}
Mathew Monfort, Alex Andonian, Bolei Zhou, Kandan Ramakrishnan, Sarah~Adel
  Bargal, Tom Yan, Lisa Brown, Quanfu Fan, Dan Gutfreund, Carl Vondrick, et~al.
\newblock Moments in time dataset: one million videos for event understanding.
\newblock {\em IEEE transactions on pattern analysis and machine intelligence},
  42(2):502--508, 2019.

\bibitem{naseer2021intriguing}
Muhammad~Muzammal Naseer, Kanchana Ranasinghe, Salman~H Khan, Munawar Hayat,
  Fahad Shahbaz~Khan, and Ming-Hsuan Yang.
\newblock Intriguing properties of vision transformers.
\newblock {\em Advances in Neural Information Processing Systems}, 34, 2021.

\bibitem{neimark2021video}
Daniel Neimark, Omri Bar, Maya Zohar, and Dotan Asselmann.
\newblock Video transformer network, 2021.

\bibitem{nguyen2020wide}
Thao Nguyen, Maithra Raghu, and Simon Kornblith.
\newblock Do wide and deep networks learn the same things? uncovering how
  neural network representations vary with width and depth.
\newblock In {\em International Conference on Learning Representations}, 2020.

\bibitem{qiu2017learning}
Zhaofan Qiu, Ting Yao, and Tao Mei.
\newblock Learning spatio-temporal representation with pseudo-3d residual
  networks.
\newblock In {\em proceedings of the IEEE International Conference on Computer
  Vision}, pages 5533--5541, 2017.

\bibitem{recht2019imagenet}
Benjamin Recht, Rebecca Roelofs, Ludwig Schmidt, and Vaishaal Shankar.
\newblock Do imagenet classifiers generalize to imagenet?
\newblock In {\em ICML}, 2019.

\bibitem{rusak2020increasing}
Evgenia Rusak, Lukas Schott, Roland~S Zimmermann, Julian Bitterwolf, Oliver
  Bringmann, Matthias Bethge, and Wieland Brendel.
\newblock Increasing the robustness of dnns against image corruptions by
  playing the game of noise.
\newblock 2020.

\bibitem{sakaridis2021acdc}
Christos Sakaridis, Dengxin Dai, and Luc Van~Gool.
\newblock {ACDC}: The adverse conditions dataset with correspondences for
  semantic driving scene understanding.
\newblock In {\em Proceedings of the IEEE/CVF International Conference on
  Computer Vision (ICCV)}, October 2021.

\bibitem{sevilla2021only}
Laura Sevilla-Lara, Shengxin Zha, Zhicheng Yan, Vedanuj Goswami, Matt Feiszli,
  and Lorenzo Torresani.
\newblock Only time can tell: Discovering temporal data for temporal modeling.
\newblock In {\em Proceedings of the IEEE/CVF Winter Conference on Applications
  of Computer Vision}, pages 535--544, 2021.

\bibitem{sigurdsson2016hollywood}
Gunnar~A. Sigurdsson, Gül Varol, Xiaolong Wang, Ali Farhadi, Ivan Laptev, and
  Abhinav Gupta.
\newblock Hollywood in homes: Crowdsourcing data collection for activity
  understanding, 2016.

\bibitem{soomro2012ucf101}
Khurram Soomro, Amir~Roshan Zamir, and Mubarak Shah.
\newblock Ucf101: A dataset of 101 human actions classes from videos in the
  wild.
\newblock {\em arXiv preprint arXiv:1212.0402}, 2012.

\bibitem{ucf101}
Khurram Soomro, Amir~Roshan Zamir, and Mubarak Shah.
\newblock Ucf101: A dataset of 101 human actions classes from videos in the
  wild.
\newblock {\em arXiv preprint arXiv:1212.0402}, 2012.

\bibitem{tang2020asynchronous}
Jiajun Tang, Jin Xia, Xinzhi Mu, Bo~Pang, and Cewu Lu.
\newblock Asynchronous interaction aggregation for action detection.
\newblock In {\em European Conference on Computer Vision}, pages 71--87.
  Springer, 2020.

\bibitem{taori2019robustness}
Rohan Taori, Achal Dave, Vaishaal Shankar, Nicholas Carlini, Benjamin Recht,
  and Ludwig Schmidt.
\newblock When robustness doesn’t promote robustness: Synthetic vs. natural
  distribution shifts on imagenet.
\newblock 2019.

\bibitem{tran2015learning}
Du~Tran, Lubomir Bourdev, Rob Fergus, Lorenzo Torresani, and Manohar Paluri.
\newblock Learning spatiotemporal features with 3d convolutional networks.
\newblock In {\em Proceedings of the IEEE international conference on computer
  vision}, pages 4489--4497, 2015.

\bibitem{tran2018closer}
Du~Tran, Heng Wang, Lorenzo Torresani, Jamie Ray, Yann LeCun, and Manohar
  Paluri.
\newblock A closer look at spatiotemporal convolutions for action recognition.
\newblock In {\em Proceedings of the IEEE conference on Computer Vision and
  Pattern Recognition}, pages 6450--6459, 2018.

\bibitem{ullo2020advances}
Silvia~Liberata Ullo and GR~Sinha.
\newblock Advances in smart environment monitoring systems using iot and
  sensors.
\newblock {\em Sensors}, 20(11):3113, 2020.

\bibitem{vanDerMaaten2008}
Laurens van~der Maaten and Geoffrey Hinton.
\newblock Visualizing data using {t-SNE}.
\newblock {\em Journal of Machine Learning Research}, 9:2579--2605, 2008.

\bibitem{wang2016temporal}
Limin Wang, Yuanjun Xiong, Zhe Wang, Yu~Qiao, Dahua Lin, Xiaoou Tang, and Luc
  Van~Gool.
\newblock Temporal segment networks: Towards good practices for deep action
  recognition.
\newblock In {\em European conference on computer vision}, pages 20--36.
  Springer, 2016.

\bibitem{wu2020robustness}
Min Wu and Marta Kwiatkowska.
\newblock Robustness guarantees for deep neural networks on videos.
\newblock In {\em Proceedings of the IEEE/CVF Conference on Computer Vision and
  Pattern Recognition}, pages 311--320, 2020.

\bibitem{xie2018rethinking}
Saining Xie, Chen Sun, Jonathan Huang, Zhuowen Tu, and Kevin Murphy.
\newblock Rethinking spatiotemporal feature learning: Speed-accuracy trade-offs
  in video classification.
\newblock In {\em Proceedings of the European conference on computer vision
  (ECCV)}, pages 305--321, 2018.

\bibitem{yang2018exploring}
Ke~Yang, Peng Qiao, Dongsheng Li, Shaohe Lv, and Yong Dou.
\newblock Exploring temporal preservation networks for precise temporal action
  localization.
\newblock In {\em Proceedings of the AAAI Conference on Artificial
  Intelligence}, volume~32, 2018.

\bibitem{yao2016highlight}
Ting Yao, Tao Mei, and Yong Rui.
\newblock Highlight detection with pairwise deep ranking for first-person video
  summarization.
\newblock In {\em Proceedings of the IEEE conference on computer vision and
  pattern recognition}, pages 982--990, 2016.

\bibitem{yi2021benchmarking}
Chenyu Yi, SIYUAN YANG, Haoliang Li, Yap peng Tan, and Alex Kot.
\newblock Benchmarking the robustness of spatial-temporal models against
  corruptions.
\newblock In {\em Thirty-fifth Conference on Neural Information Processing
  Systems Datasets and Benchmarks Track (Round 2)}, 2021.

\bibitem{yim2007modular}
Mark Yim, Wei-Min Shen, Behnam Salemi, Daniela Rus, Mark Moll, Hod Lipson, Eric
  Klavins, and Gregory~S Chirikjian.
\newblock Modular self-reconfigurable robot systems [grand challenges of
  robotics].
\newblock {\em IEEE Robotics \& Automation Magazine}, 14(1):43--52, 2007.

\bibitem{yin2019fourier}
Dong Yin, Raphael~Gontijo Lopes, Jonathon Shlens, Ekin~D Cubuk, and Justin
  Gilmer.
\newblock A fourier perspective on model robustness in computer vision.
\newblock In {\em Proceedings of the 33rd International Conference on Neural
  Information Processing Systems}, pages 13276--13286, 2019.

\bibitem{yun2019cutmix}
Sangdoo Yun, Dongyoon Han, Seong~Joon Oh, Sanghyuk Chun, Junsuk Choe, and
  Youngjoon Yoo.
\newblock Cutmix: Regularization strategy to train strong classifiers with
  localizable features.
\newblock In {\em Proceedings of the IEEE/CVF International Conference on
  Computer Vision}, pages 6023--6032, 2019.

\bibitem{zhang2021adversarial}
Xingwei Zhang, Xiaolong Zheng, and Wenji Mao.
\newblock Adversarial perturbation defense on deep neural networks.
\newblock {\em ACM Computing Surveys (CSUR)}, 54(8):1--36, 2021.

\bibitem{zhou2018temporal}
Bolei Zhou, Alex Andonian, Aude Oliva, and Antonio Torralba.
\newblock Temporal relational reasoning in videos.
\newblock In {\em Proceedings of the European Conference on Computer Vision
  (ECCV)}, pages 803--818, 2018.

\end{thebibliography}
}

\appendix
\clearpage

\section{Overview}
This supplementary includes additional results that were not available in the main paper.  Section \ref{additional_results} includes additional results on UCF-101P, HMDB-51P, Kinetcs-400P, and SSv2P. More specifically:

\begin{itemize}
  \item Section \ref{sec:variation_in_severity} goes over the results for perturbations of varying severity in the datasets UCF-101P, HMDB-51P, Kinetics-400P, and provides more detail on SSv2P.
  \item Section \ref{sec:absolute_relative_robustness} shows more details on the results for absolute and relative robustness scores on all three datasets.
  \item  Section \ref{sec:pretraining_vs_scratch} provides further analysis on pre-training versus models from scratch on UCF-101P and HMDB-51P.
  \item Section \ref{sec:ssv2_class_analysis} provides a more in-depth analysis on the class confusions that result from SSv2P. 
\end{itemize}
Section \ref{sec:implementation_details} goes into more detail on the perturbations applied to generate UCF-101P, HMDB-51P, Kinetics-400P and SSv2P. Next, we will provide details on model training using perturbations as augmentations followed by details of UCF-101-DS dataset.

\begin{wrapfigure}[24]{r}{0.5\textwidth}
    \centering
    \includegraphics[width=\linewidth]{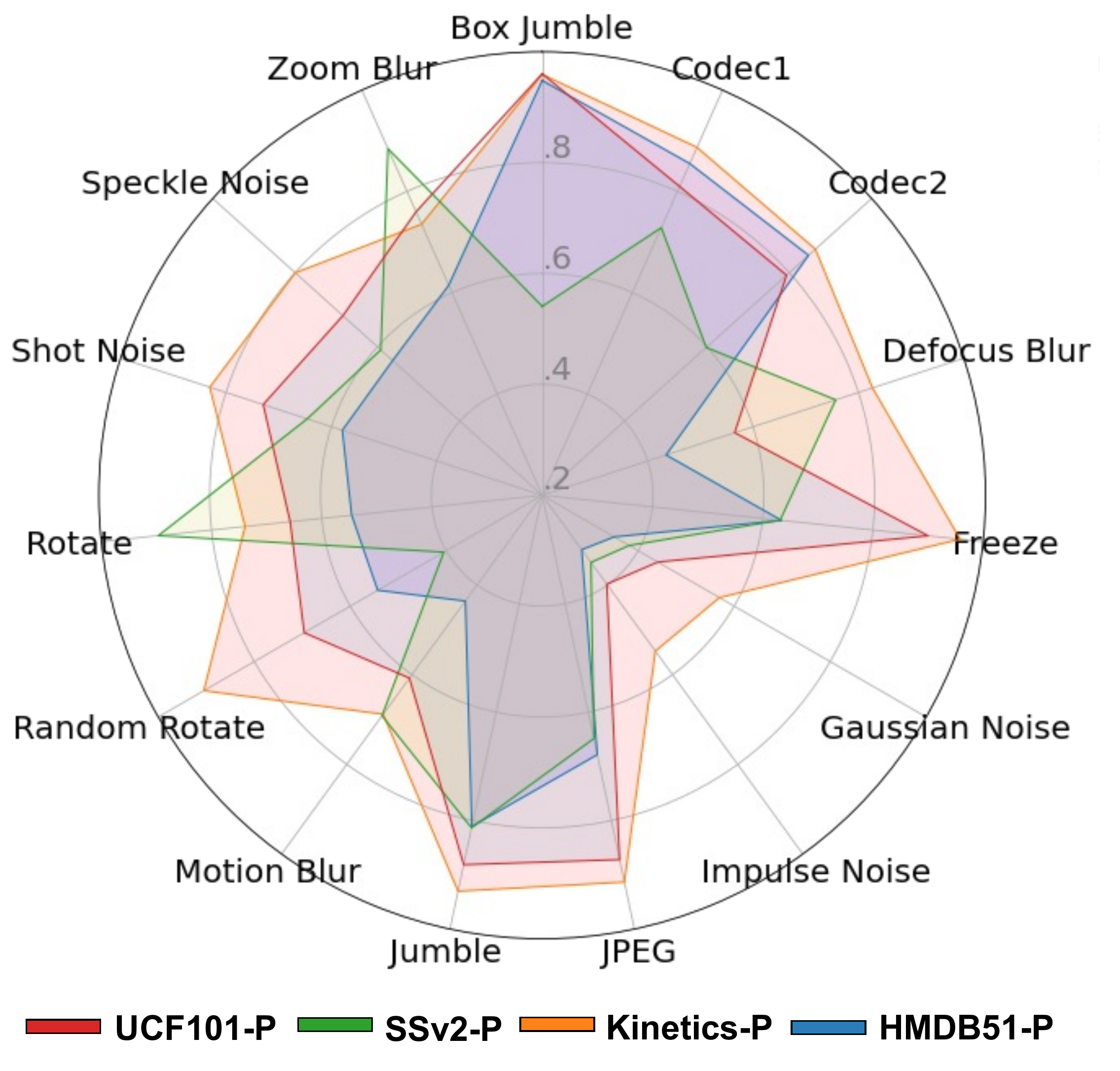}
    \caption{Mean performance on the different perturbed datasets.}
    \label{fig:dataset_differences}
\end{wrapfigure}

\section{Additional results}
\label{additional_results}
Here we provide detailed results on all the datasets. Figure \ref{fig:dataset_differences} summarizes the performance of different models on various datasets for different perturbations. Here we can observe the differences in the behavior of these datasets how they differ for different type of perturbations. More specifically, SSv2-P differs from other datasets such as Kinetics-400P and UCF-101P.

\subsection{Real-World Distribution Shifts}
\begin{figure*}
    \centering
    \includegraphics[width=\linewidth]{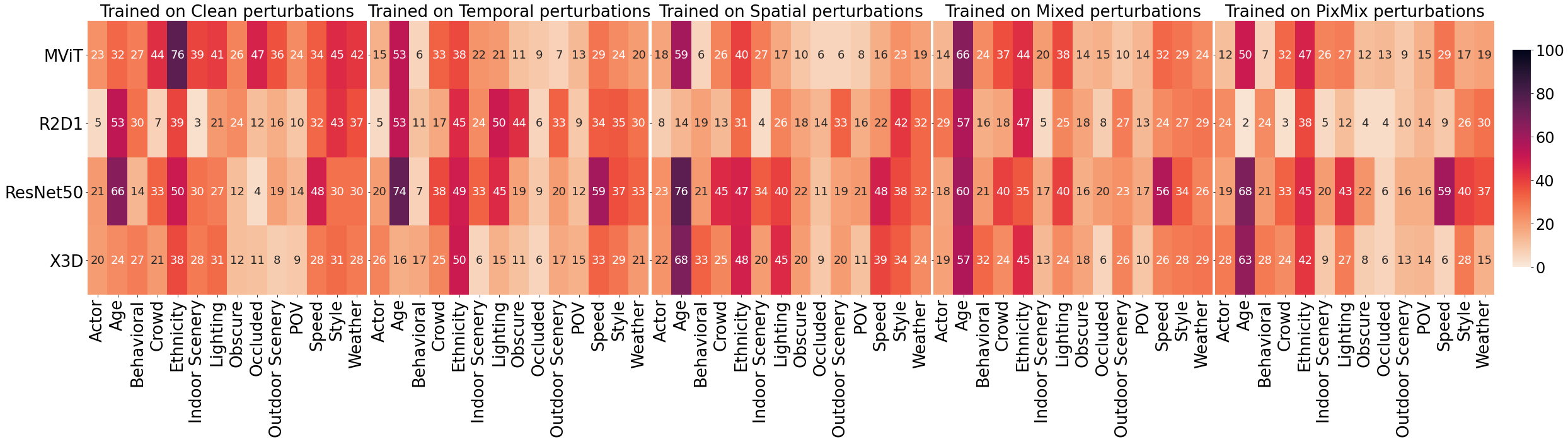}
    \caption{Model evaluation on the UCF101-DS dataset when models are trained on clean versus different combinations of perturbed datasets. Mixed perturbations are both spatial and temporal perturbations and PixMix is from \cite{hendrycks2022pixmix}.}
    \label{fig:all_models_ucf101DS}
\end{figure*}

Results for models when trained on different data and evaluated on USF101-DS is shown in Figure \ref{fig:all_models_ucf101DS}. Each column is a different version of the UCF101 dataset. Mixed perturbations are a combination of all perturbations and PixMix are perturbations from \cite{hendrycks2022pixmix} extended for video. For more details on the training implementation, see Section \ref{training_perturbations}.  Table \ref{tab:rel_robustness_ucf101DS} shows the relative robustness ($\gamma^r$) scores across the different distribution shift categories comparing models trained on the clean UCF101 dataset versus perturbed. In this case, we treat the models trained on perturbations as the original score, where $\gamma^r_p = 1 - (A^{f}_p - A^{f_p}_p)/ A^{f_p}_p$. We can observe that MViT performs much worse when trained on perturbations while CNN based models like ResNet and X3D improve scores when using spatial perturbations. 

\begin{wrapfigure}[10]{r}{0.6\textwidth}
\vspace{-6pt}
    \centering
    \resizebox{\linewidth}{!}{\begin{tabular}{lllll}
        \hline
         Model &  Mixed $(\gamma^r)$ & PixMix $(\gamma^r)$ & Spatial $(\gamma^r)$ & Temporal $(\gamma^r)$ \\
        \hline
       MViT     &              $0.59$ &              $0.29$ &               $0.08$ &                $0.21$ \\
        R2D1     &              $1.02$ &              $0.37$ &               $0.86$ &       $\mathbf{1.16}$ \\
        ResNet50 &  $\underline{1.06}$ &     $\mathbf{1.11}$ &   $\underline{1.16}$ &    $\underline{1.12}$ \\
        X3D      &     $\mathbf{1.12}$ &  $\underline{0.98}$ &      $\mathbf{1.25}$ &                $0.90$ \\
        \hline
        \end{tabular}}

    \caption{Relative robustness scores comparing models trained on perturbed UCF101 to models trained on clean UCF101, evaluated and averaged over UCF101-DS categories. Here, $\gamma^r_p = 1 - (A^{f}_p - A^{f_p}_p)/ A^{f_p}_p$. }
    \label{tab:rel_robustness_ucf101DS}
\end{wrapfigure}

\subsection{Variation in severity}
\label{sec:variation_in_severity}
\begin{figure*}[t!]
\begin{center}
\includegraphics[width=.24\linewidth]{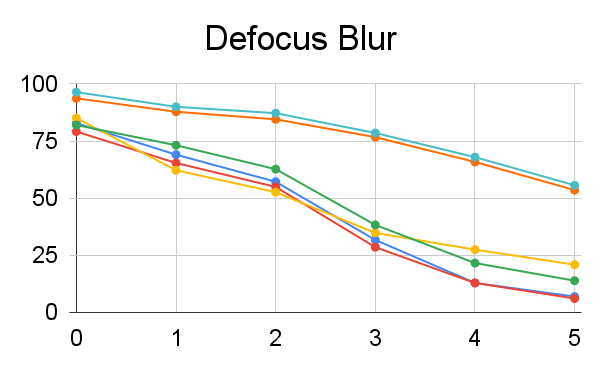}
\includegraphics[width=.24\linewidth]{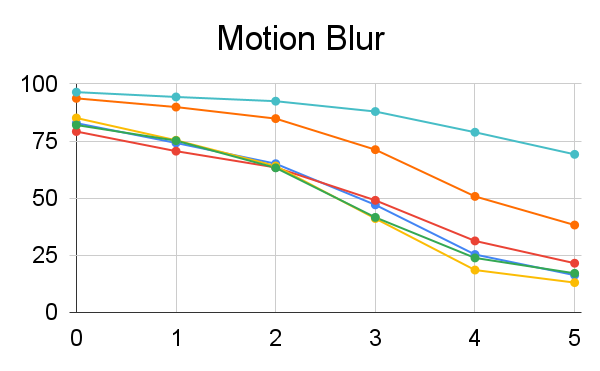}
\includegraphics[width=.24\linewidth]{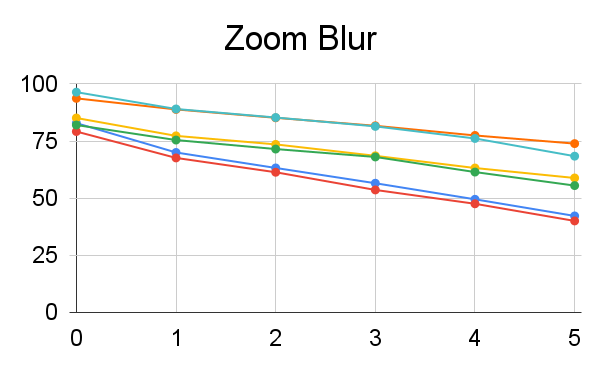}
\includegraphics[width=.24\linewidth]{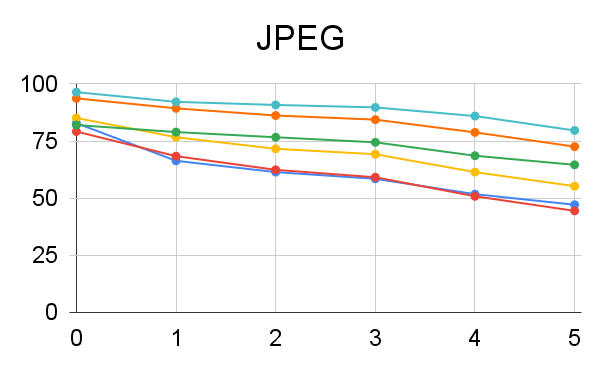}
\includegraphics[width=.24\linewidth]{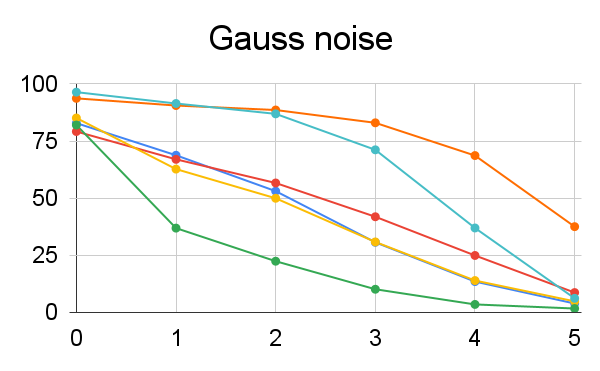}
\includegraphics[width=.24\linewidth]{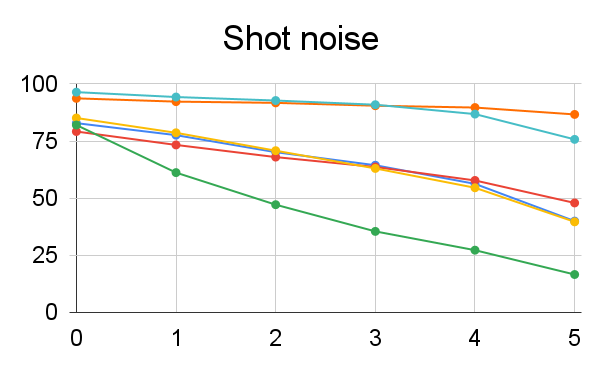}
\includegraphics[width=.24\linewidth]{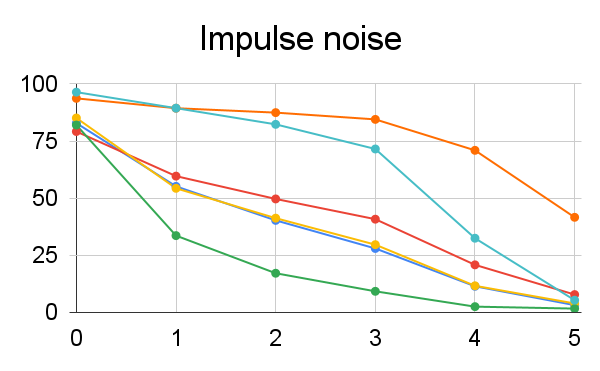}
\includegraphics[width=.24\linewidth]{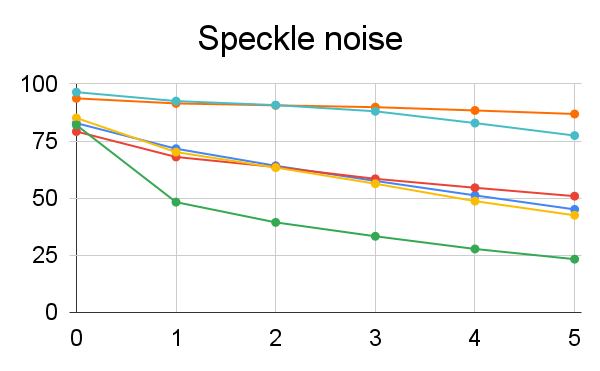}\\
\includegraphics[width=.24\linewidth]{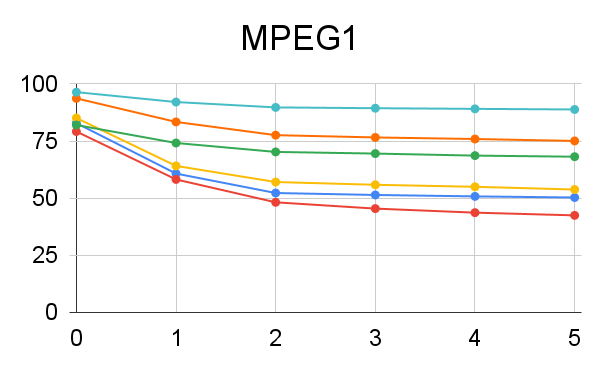}
\includegraphics[width=.24\linewidth]{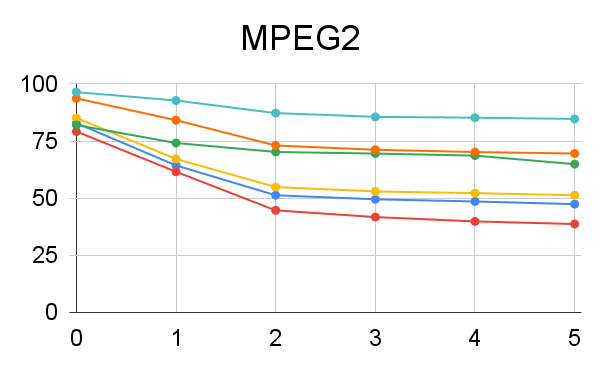}
\includegraphics[width=.24\linewidth]{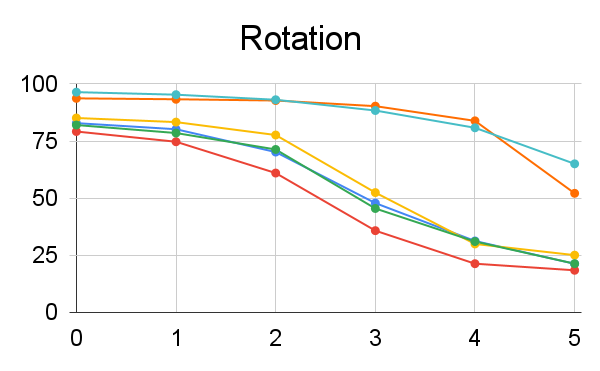}
\includegraphics[width=.24\linewidth]{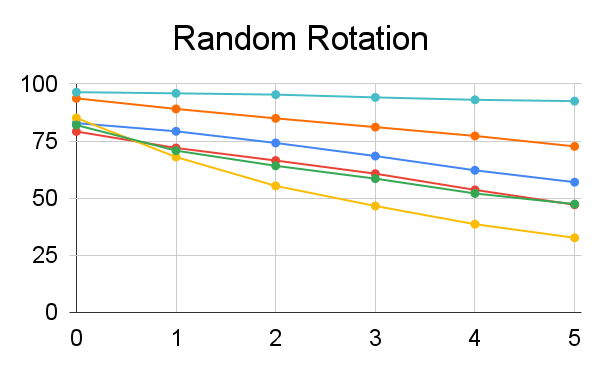}\\
\includegraphics[width=.24\linewidth]{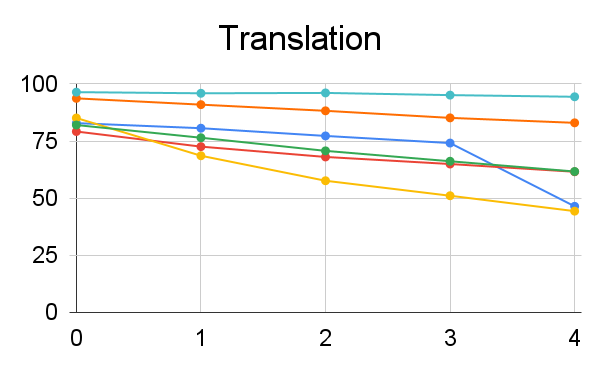}
\includegraphics[width=.24\linewidth]{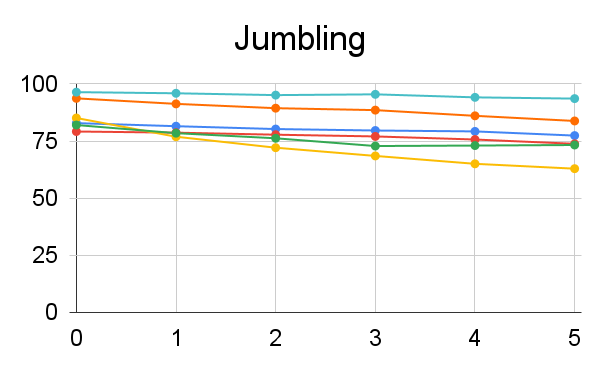}
\includegraphics[width=.24\linewidth]{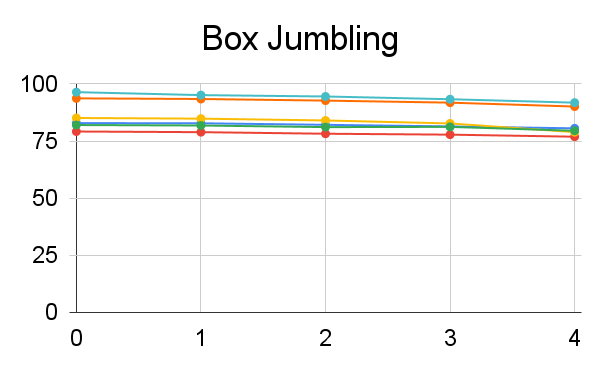}
\includegraphics[width=.24\linewidth]{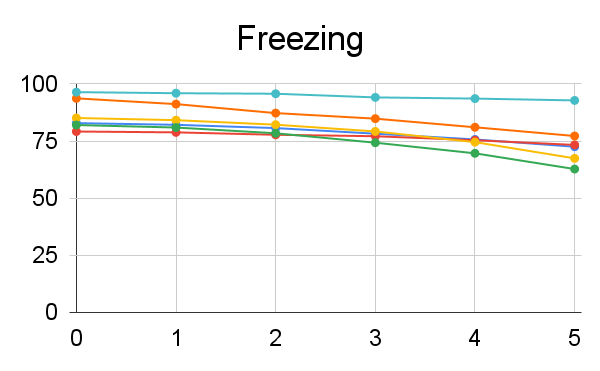}\\
\includegraphics[width=.45\linewidth]{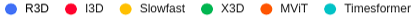}  
\end{center}
  \caption{Robustness analysis of models with varying severity levels on UCF-101P benchmark. The y-axis shows accuracy and the x-axis represents severity level where 0 indicates performance on clean videos. 
  }
\label{fig:acc_sev_ucf}
\end{figure*}

\begin{figure*}[t!]
\begin{center}
\includegraphics[width=0.24\linewidth]{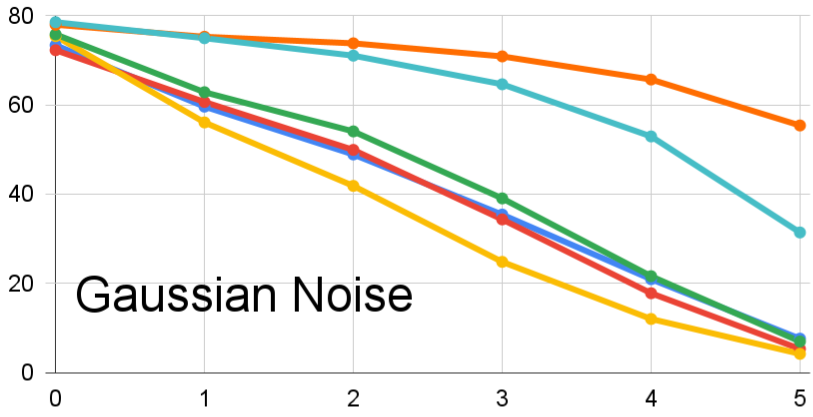}
\includegraphics[width=0.24\linewidth]{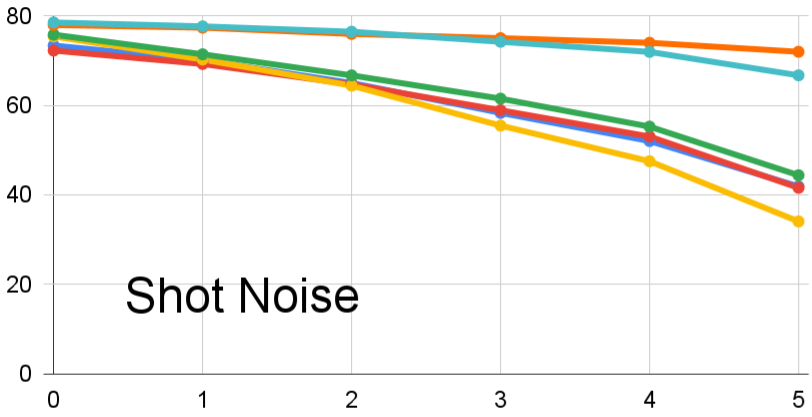}
\includegraphics[width=0.24\linewidth]{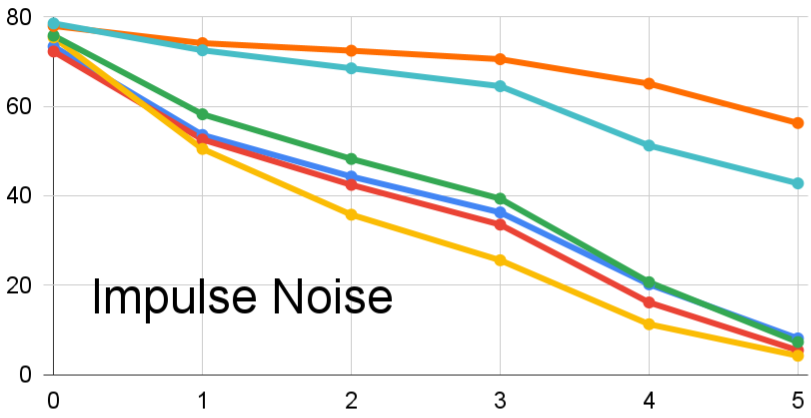}
\includegraphics[width=0.24\linewidth]{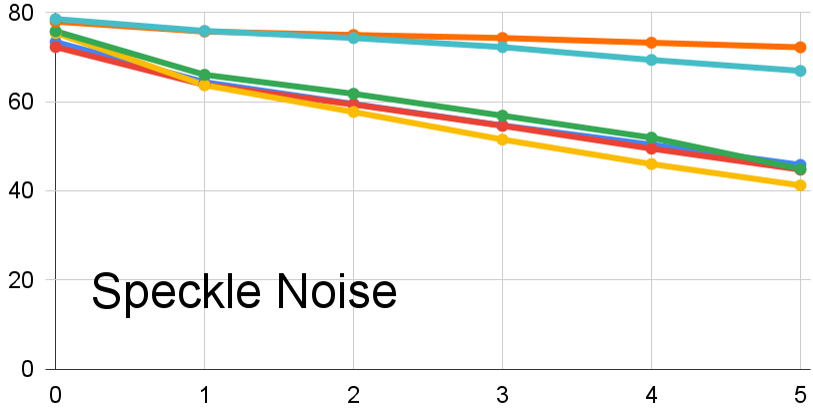}
\\
\includegraphics[width=0.24\linewidth]{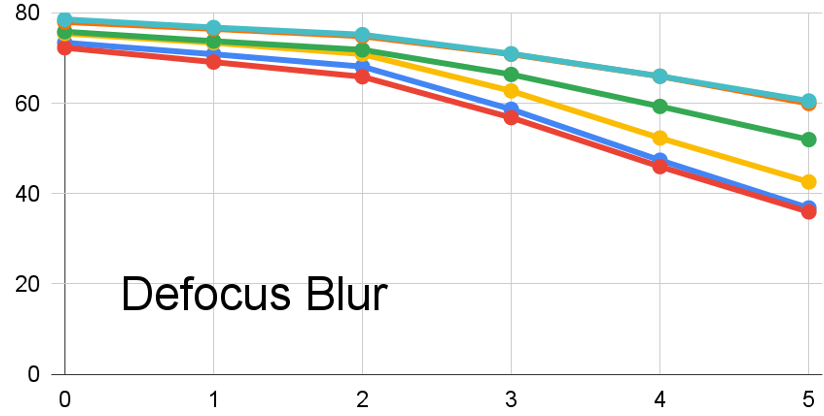}
\includegraphics[width=0.24\linewidth]{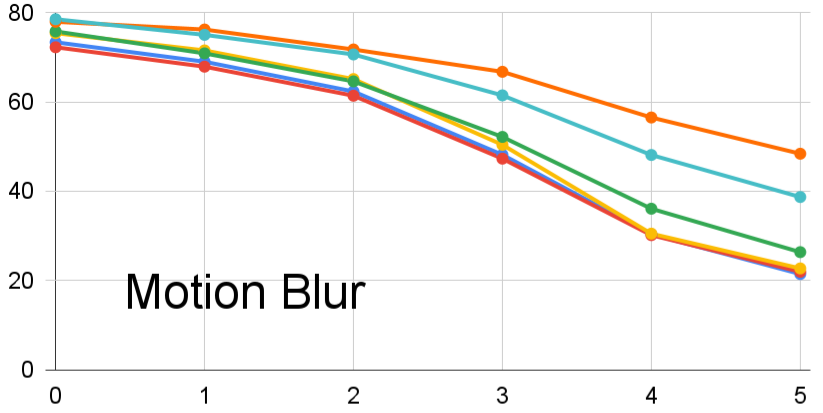}
\includegraphics[width=0.24\linewidth]{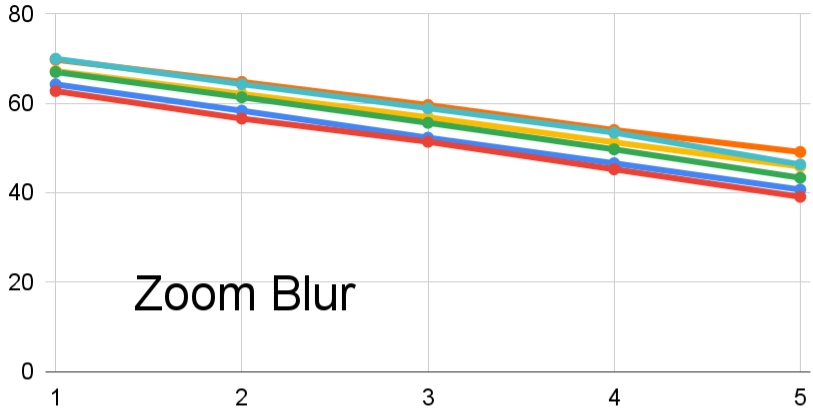}
\includegraphics[width=0.24\linewidth]{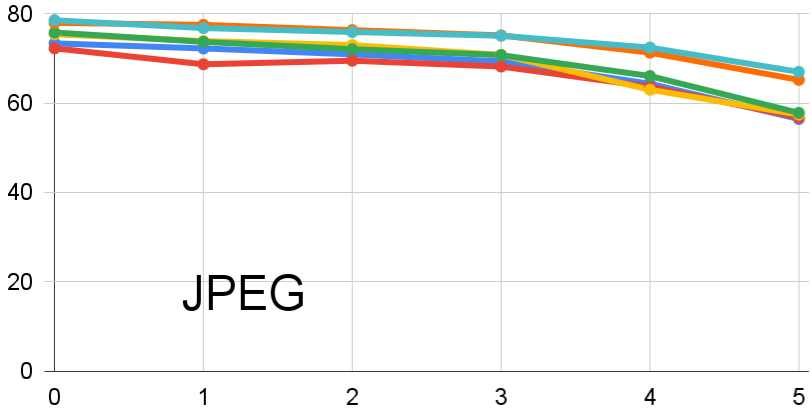}
\\
\includegraphics[width=0.24\linewidth]{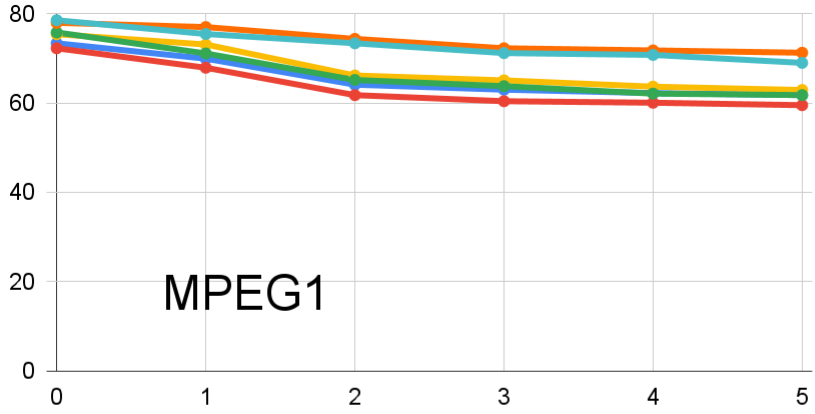}
\includegraphics[width=0.24\linewidth]{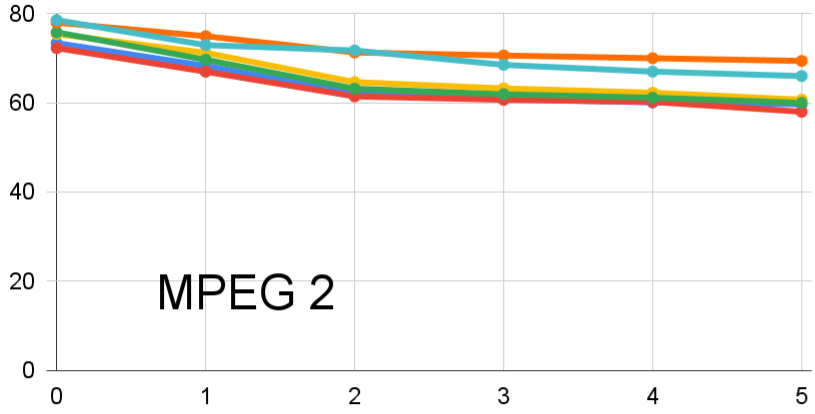}
\includegraphics[width=0.24\linewidth]{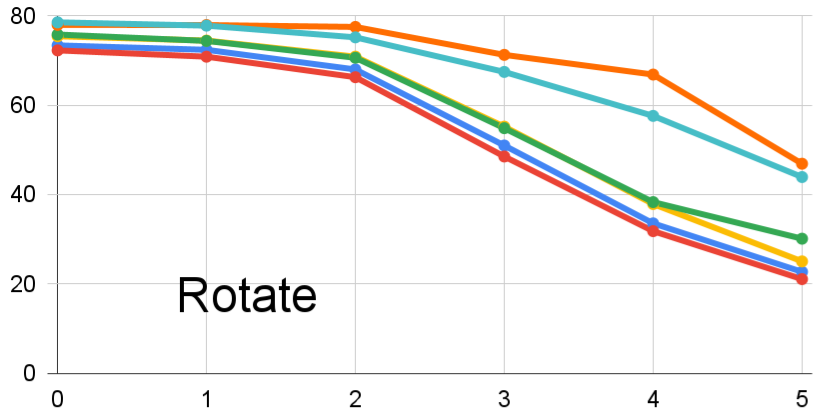}
\includegraphics[width=0.24\linewidth]{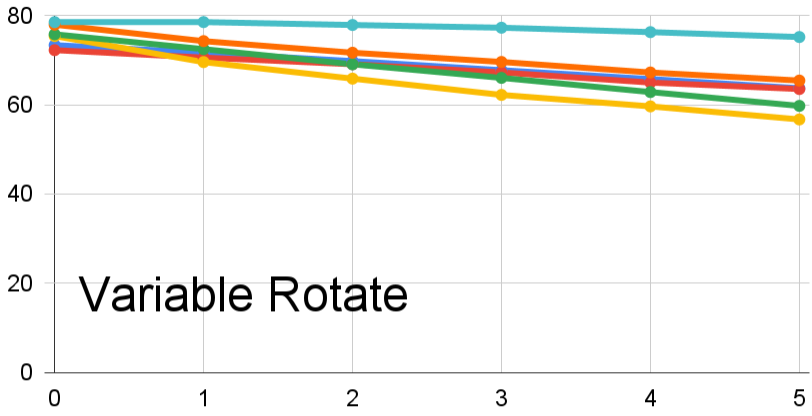}
\\
\includegraphics[width=0.24\linewidth]{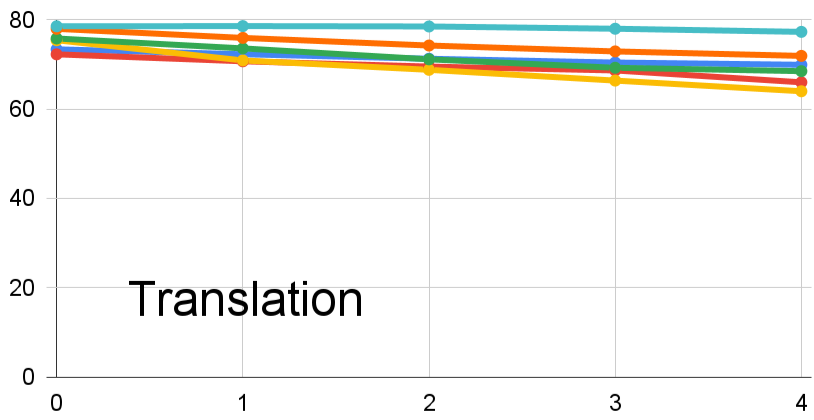}
\includegraphics[width=0.24\linewidth]{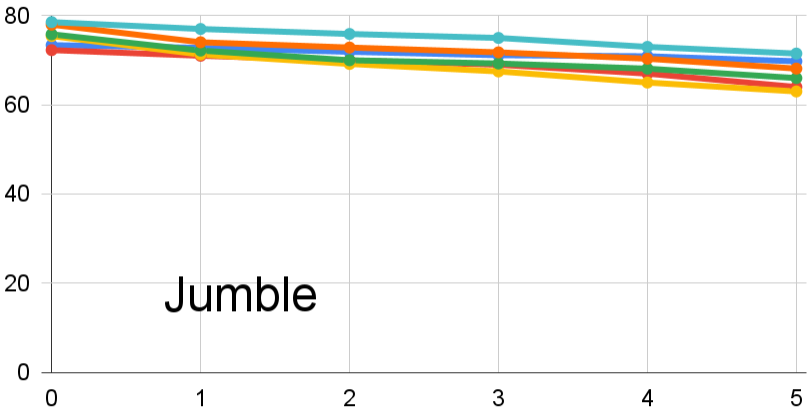}
\includegraphics[width=0.24\linewidth]{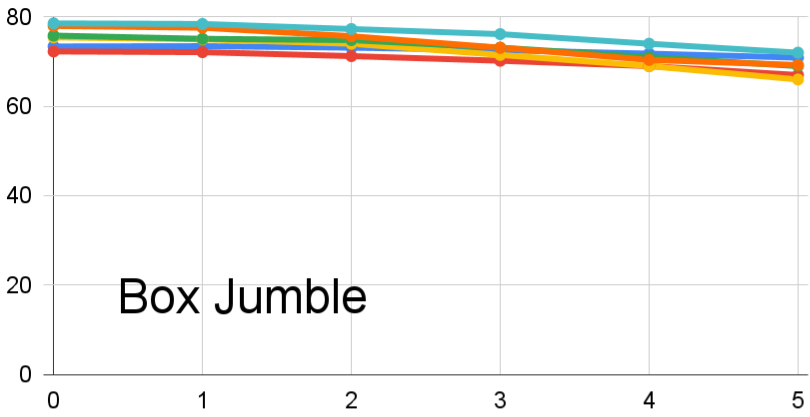}
\includegraphics[width=0.24\linewidth]{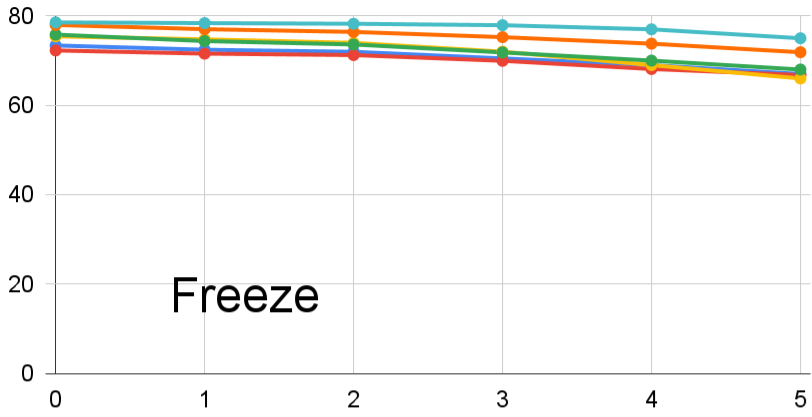}
\\
\includegraphics[width=0.30\linewidth]{images/Fig41/leg.png}  
\end{center}
  \caption{Robustness analysis of models with varying severity levels on Kinetics-400P benchmark. The y-axis shows accuracy and the x-axis represents severity level. When the severity is 0, this means no perturbation was applied and is the model's performance on clean videos. }
\label{fig:acc_sev_kinetics}
\end{figure*}

\begin{figure*}[t!]
\begin{center}
\includegraphics[width=.24\linewidth]{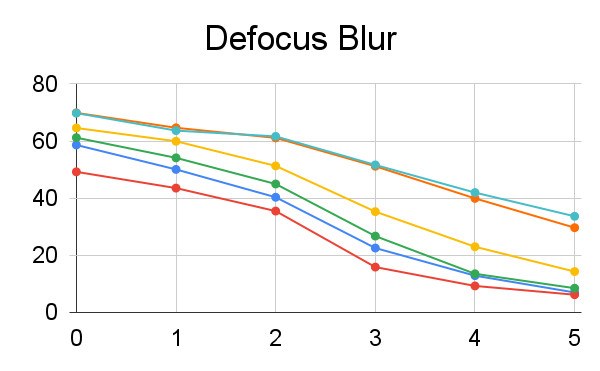}
\includegraphics[width=.24\linewidth]{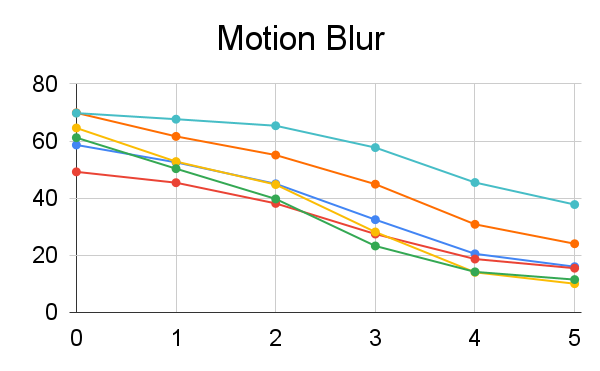}
\includegraphics[width=.24\linewidth]{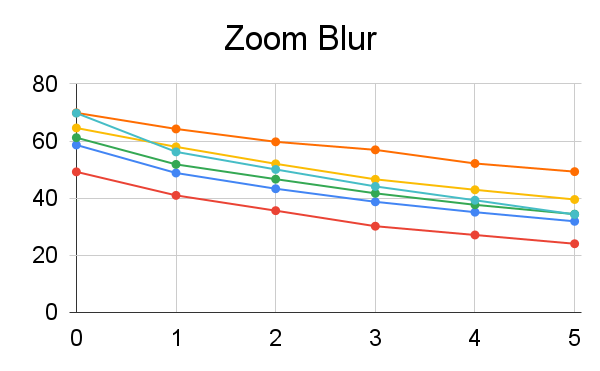}
\includegraphics[width=.24\linewidth]{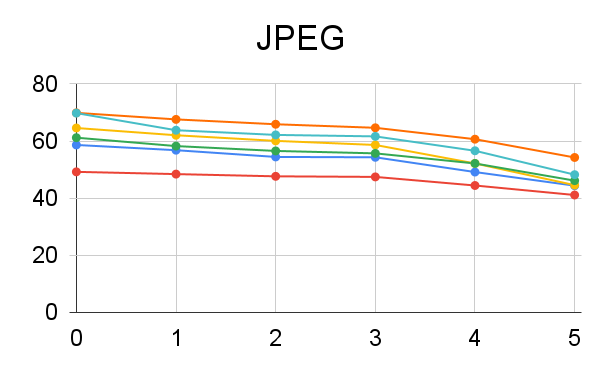}
\includegraphics[width=.24\linewidth]{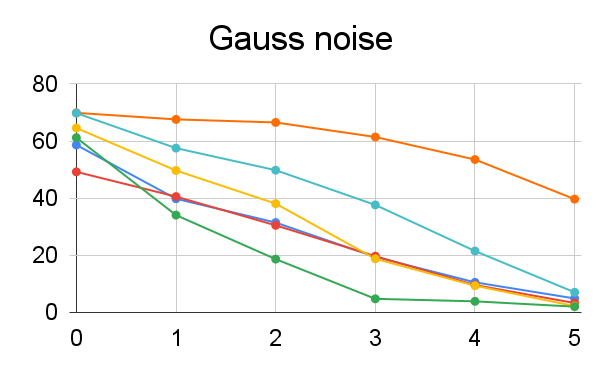}
\includegraphics[width=.24\linewidth]{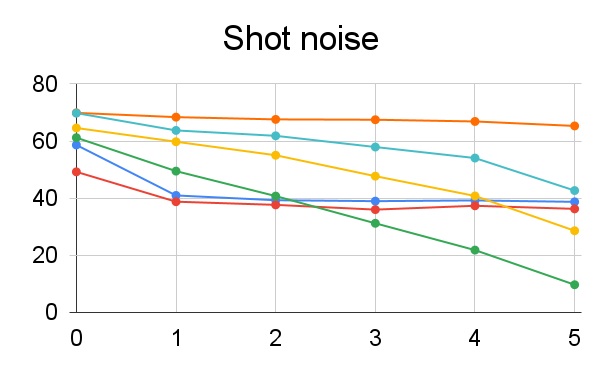}
\includegraphics[width=.24\linewidth]{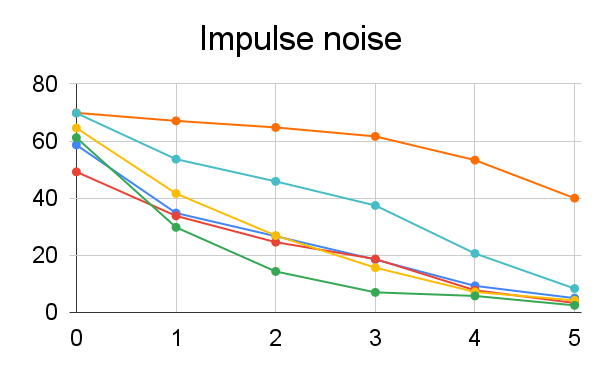}
\includegraphics[width=.24\linewidth]{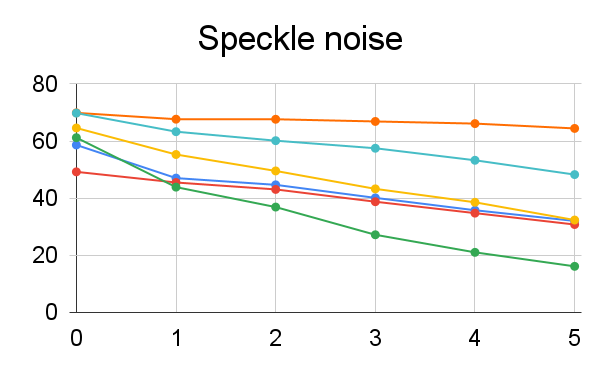}\\
\includegraphics[width=.24\linewidth]{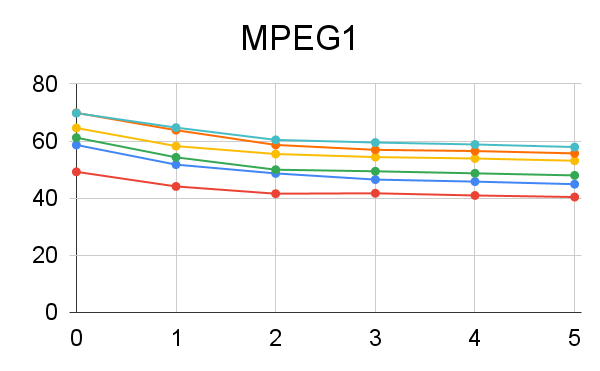}
\includegraphics[width=.24\linewidth]{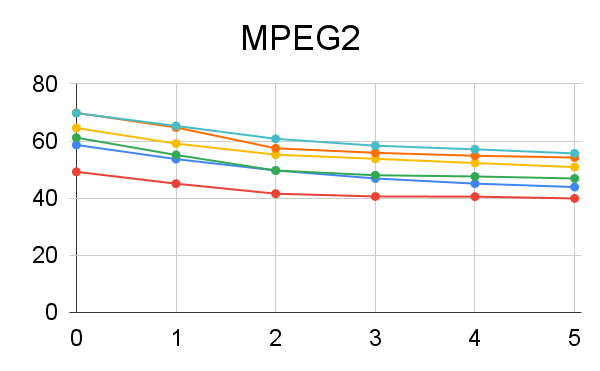}
\includegraphics[width=.24\linewidth]{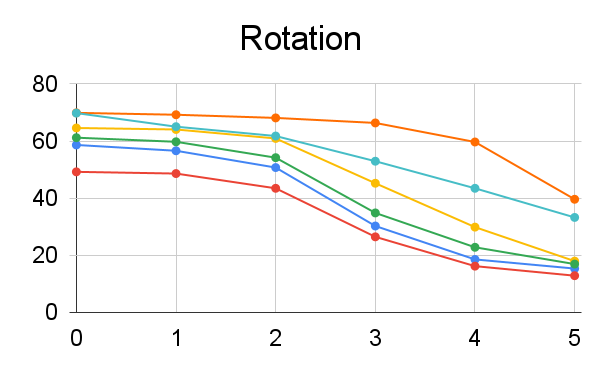}
\includegraphics[width=.24\linewidth]{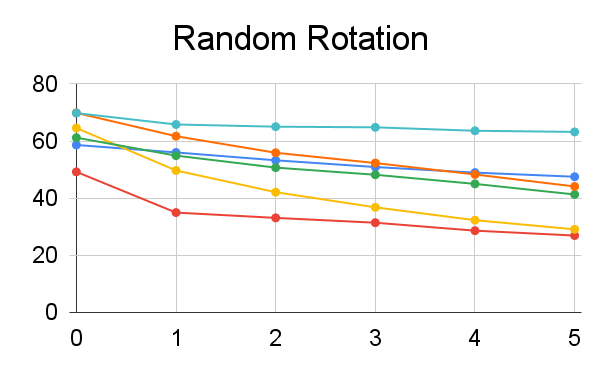}\\
\includegraphics[width=.24\linewidth]{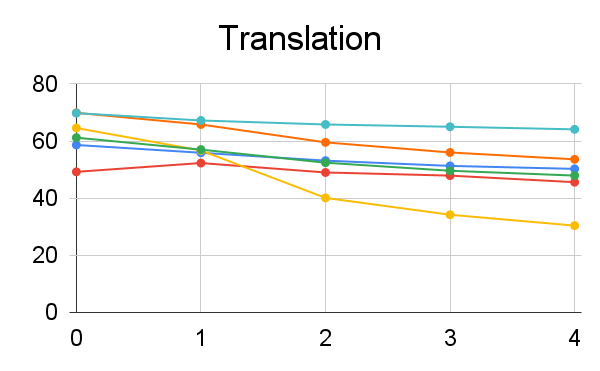}
\includegraphics[width=.24\linewidth]{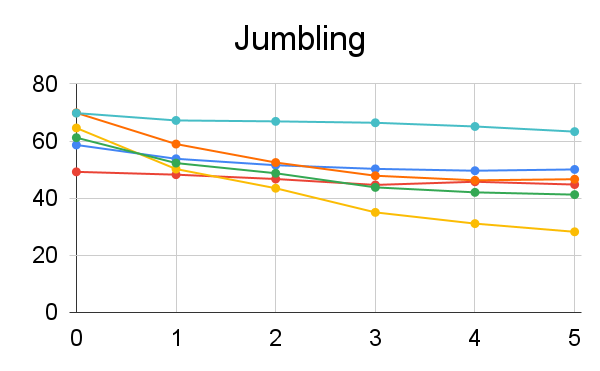}
\includegraphics[width=.24\linewidth]{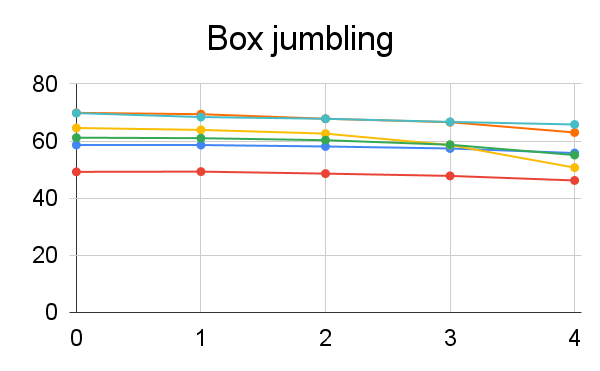}
\includegraphics[width=.24\linewidth]{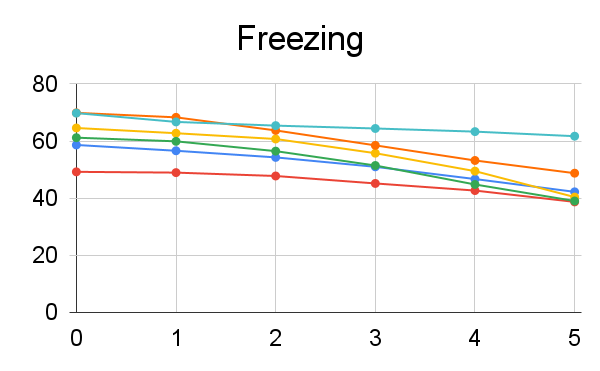}\\
\includegraphics[width=.45\linewidth]{images/Fig41/leg.png}  
\end{center}
  \caption{Robustness analysis of models with varying severity levels on HMDB-51P benchmark. The y-axis shows accuracy and the x-axis represents severity level. When the severity is 0, this means no perturbation was applied and is the model's performance on clean videos.
  }
\label{fig:acc_sev_hmdb}
\end{figure*}

Figure \ref{fig:acc_sev_ucf}, Figure \ref{fig:acc_sev_kinetics}, and Figure \ref{fig:acc_sev_hmdb} shows the performance of all the models with different perturbations at varying severity levels for UCF-101P, Kinetics-400P, and HMDB-51P respectively. The severity level varies from 0 to 5, where 0 refers to clean videos and 5 refers to heavy perturbation. We observe that the transformer based models, MViT and Timesformer, are generally more robust as severity level increases in most of the perturbations, For some of the perturbations, such as defocus blur and motion blur, the performance of all the models drops significantly. However, there are some perturbations, such as short noise and speckle noise, where the transformer based models are robust across all the severity levels. Moreover, there are some perturbations, where all the models are found to be robust against all severity levels, such as box jumbling. We observe similar behavior in HMDB-51P dataset as well. In addition, Timesformer model is found to be robust against Translation and Random rotation for all the severity levels.

\begin{figure*}[t!]
\begin{center}
\includegraphics[width=.24\linewidth]{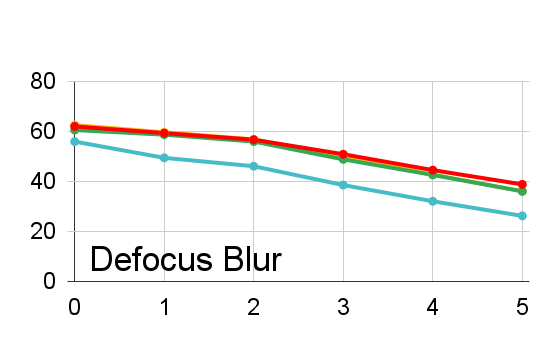}
\includegraphics[width=.24\linewidth]{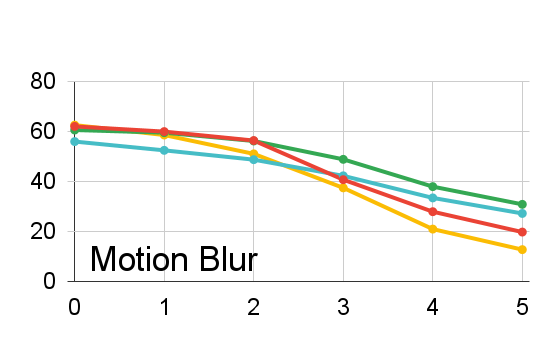}
\includegraphics[width=.24\linewidth]{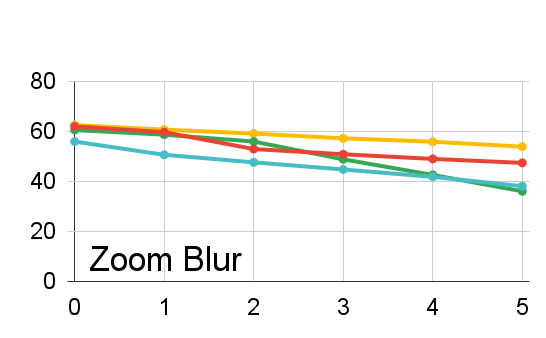}
\includegraphics[width=.24\linewidth]{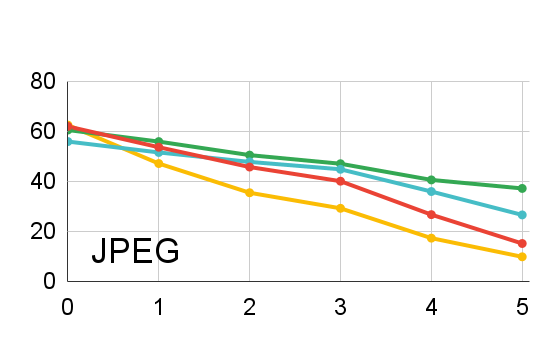}
\includegraphics[width=.24\linewidth]{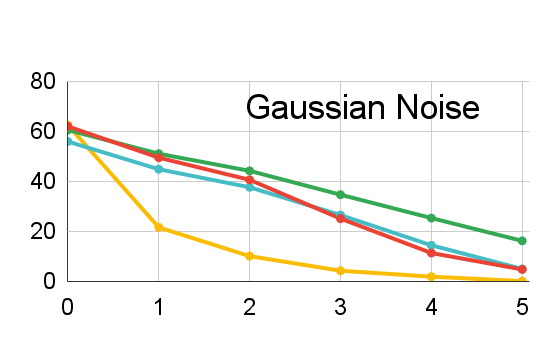}
\includegraphics[width=.24\linewidth]{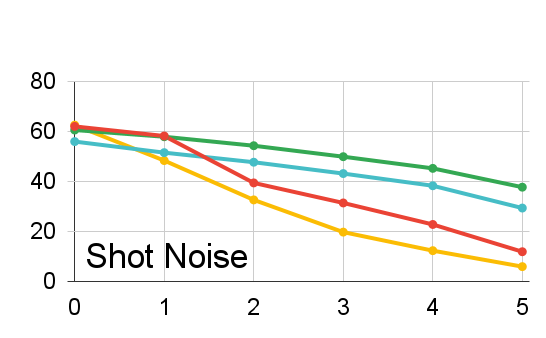}
\includegraphics[width=.24\linewidth]{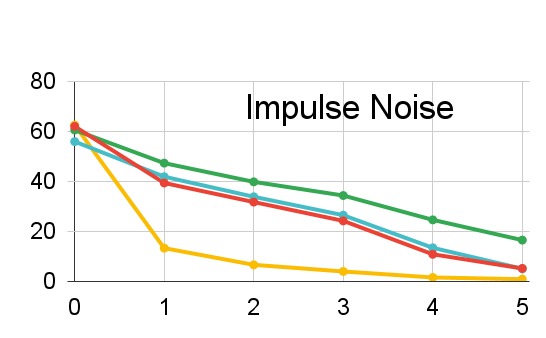}
\includegraphics[width=.24\linewidth]{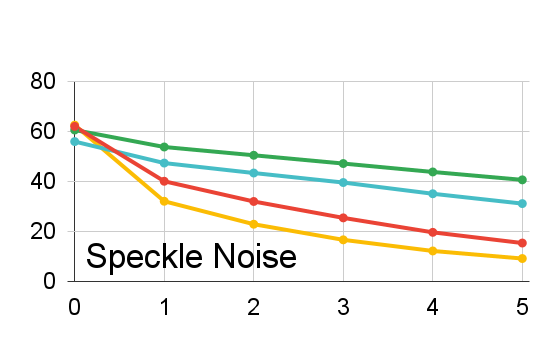}\\
\includegraphics[width=.24\linewidth]{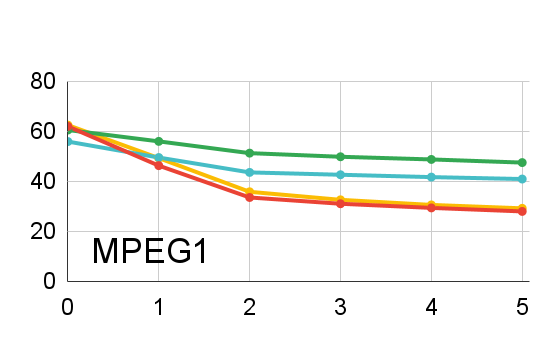}
\includegraphics[width=.24\linewidth]{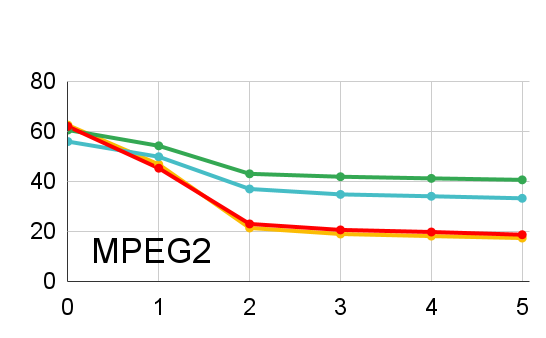}
\includegraphics[width=.24\linewidth]{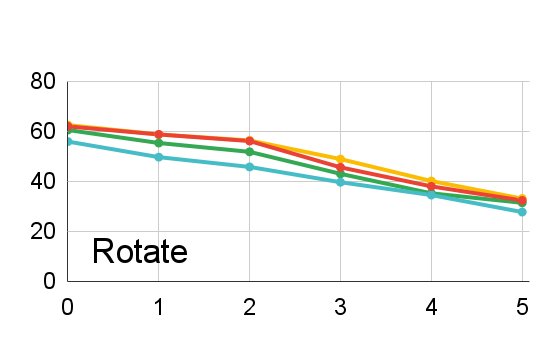}
\includegraphics[width=.24\linewidth]{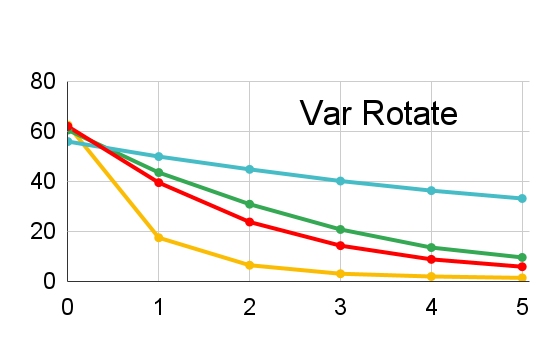}\\
\includegraphics[width=.24\linewidth]{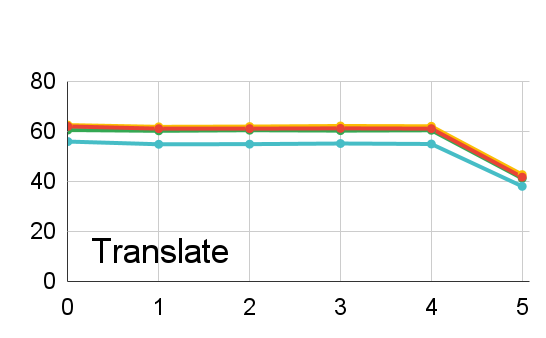}
\includegraphics[width=.24\linewidth]{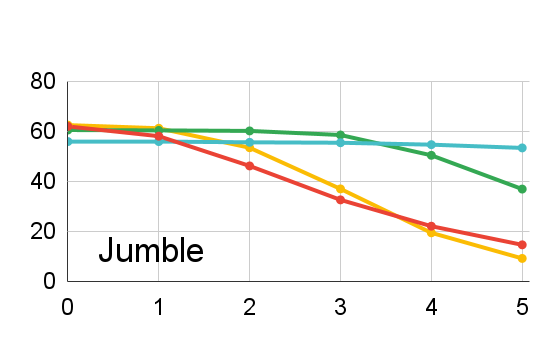}
\includegraphics[width=.24\linewidth]{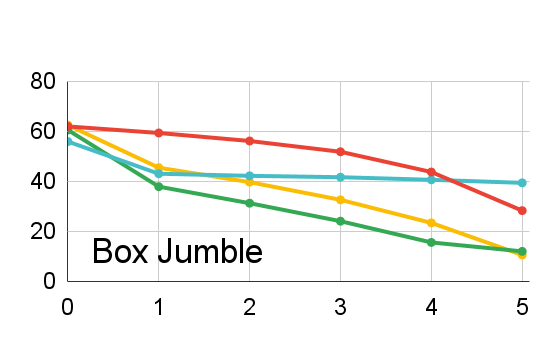}
\includegraphics[width=.24\linewidth]{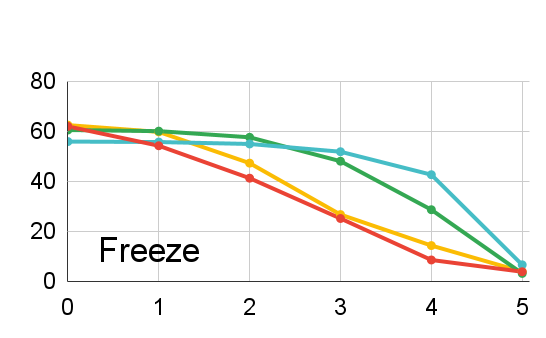}\\
\includegraphics[width=.45\linewidth]{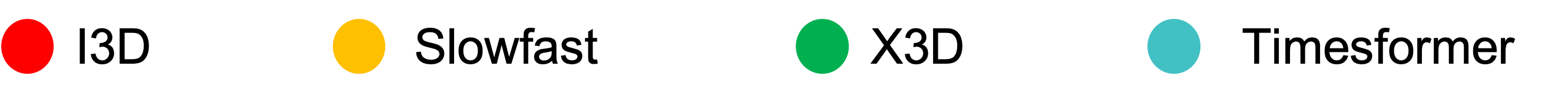}  
\end{center}
  \caption{Robustness analysis of models with varying severity levels on SSv2P benchmark. The y-axis shows accuracy and the x-axis represents severity level. When the severity is 0, this means no perturbation was applied and is the model's performance on clean videos.
  }
\label{supp:fig:acc_sev_ssv2}
\end{figure*}

\begin{table*}
\centering
\begin{center}
\resizebox{.95\textwidth}{!}{
\begin{tabular}{|l|p{.75cm}|p{.75cm}|p{.75cm}|p{.75cm}|p{.75cm}|p{.75cm}|p{.75cm}|p{.75cm}|p{.75cm}|p{.75cm}|p{.75cm}|p{.75cm}|}
\hline
             & \multicolumn{2}{l|}{R3D} & \multicolumn{2}{l|}{I3D} & \multicolumn{2}{l|}{SF} & \multicolumn{2}{l|}{X3D} & \multicolumn{2}{l|}{Timesformer} & \multicolumn{2}{l|}{MViT} \\
            \hline
Defocus Blur &     .83       &      .77      & .82           &       .75     &    .85        &       .80    &         .89    &      .85     & .84             &     .80        &.87             &        .83    \\
Motion Blur  &      .73      &       .63     & .78           &      .70      &    .73        &       .64    &.74            &          .66  &           .80   &       .75      &.86             &     .82       \\
Zoom Blur    &  .79          &      .71      &.81            &        .74    &.85            &       .80    &.89            &        .85    & .91             &          .89   &   .92          &   .90         \\
\hline
Blur         &   .78&.70       &      .80&.72     &        .80&.73    &      .81&.75            &      .84&.79            &    \textbf{.86}&\textbf{.82}            \\
\hline
Gaussian     &.61            &       .47     &.61            &  .46          &.52            &        .36   &        .61    &        .49    &   .80  &           .75 &.90             &           .87 \\
Shot         &.84            &  .78          &.85            &     .79       &.79           &      .82     &.84            &           .79 &  .95            &    .94         & .96            &            .95\\
Impulse      &      .59      &     .44       &.58            &      .42      &    .50        &        .34   &.59            &        .46    &    .81          &        .76     &    .90         &            .87\\
Speckle      & .82           &        .75    & .82           &       .75     &    .77        &  .70         &.81           &           .75 &  .93        &           .91  &  .96           &            .95\\
\hline
Noise        &     .71       &      .61      &      .72      &      .61      &            .64&.53            &           .71&.62             &              .87&.84            &              \textbf{.93} &\textbf{.91}            \\
\hline
JPEG         &      .93      &  91          & .93           &   .90         &    .94        &      .92     &.92            &           .89 &   .95           &           .94  &   .95          &            .94\\
MPEG1        &  .92          &    .89        &.90           &     .86     &.91           &       .88   &          .90  &          .87  & .94             &            .92 &     .95        &            .94\\
MPEG2        &      .89     &       .85     &.90           &        .86    &.89           &        .85   &     .87       &        .83    &            .91  &         .89    &        .93     &          .91 \\
\hline
Digital      &       .91&.88      &      .91&.87       &     .91&.89   &   .90&.86         &         \textbf{.94}&.92     &     \textbf{.94}&\textbf{.93}    \\
\hline
Rotation        &   .76        &   .67         &.75            &        .65    &  .77         & .70          &.78           &           .71 & .86            &            .82 &   .90          &            .87\\
Variable Rotation       &  .94          & .92           &.95            &  .93          &.87            &      .83     &          .90  &   .87         & .98             &.97         &     .92        &      .90      \\
Translate       &      .98     &    .97        &.97            &   .96         &.92           &        .89   &     .95      &           .93 &            .99  &           .99  &        .96   &            .95\\
\hline
Camera motion    &      .89&.85     &       .89&.85     &    .86&.81       &   .88&.84          &     \textbf{.95}&\textbf{.93}       &        .94&.92 \\
\hline
Sampling     &.97            &      .96      & .97           &      .96     &.94            &      .92     &    .96        &           .95 &  .97            &        .96     &.96             &            .95\\
Reversing    &    .97        &         .96   & .97           &        .96    &.94            &         .92  &    .96        &           .95 &.97              &          .96   &.95             &           .94 \\
Jumbling     & .98           &          .97  &.97            &      .96     &.92           &      .89     &    .93        &           .91 &       .96       &        .95     & .93            &          .91  \\
Box Jumbling    &          .99  &        .99    &.98            &      .97      &.97            &      .96     &.97            &          .96  & .98             &        .97     &.95             &    .94       \\
Freezing     &   .97         &          .96  & .97           &            .96 &.96            &         .95  &    .96        &           .95 & .99             &            .99 &.97             &    .96        \\
\hline
Temporal     &    \textbf{.98}&\textbf{.97}    &      .97&.96    &       .95&.93     &       .96&.94      &         .97&.96      &      .96&.95  \\
\hline
\end{tabular}}
\end{center}
\caption{Absolute and relative robustness scores averaged across all severity levels for all categories of perturbations for Kinetics-400P dataset. For each category an average is also shown at the end of all sub-categories. The best models are marked as BOLD for both relative and absolute robustness for each perturbation and their categories. }
\label{table_kinetics}
\end{table*}

\begin{table*}
\centering
\begin{center}
\begin{tabular}{|l|p{1cm}|p{1cm}|p{1cm}|p{1cm}|p{1cm}|p{1cm}|p{1cm}|p{1cm}|p{1cm}|p{1cm}|}
\hline
      &  \multicolumn{2}{l|}{I3D} &  \multicolumn{2}{l|}{SF} &  \multicolumn{2}{l|}{X3D} &  \multicolumn{2}{l|}{Timesformer} \\
     \hline
        Defocus Blur &               .88 &               .81 &                    .87 &                    .78 &               .88 &               .80 &                       .83 &                       .69 \\
      Motion Blur &               .79 &               .66 &                    .74 &                    .58 &               .86 &               .77 &                       .85 &                       .73 \\
        Zoom Blur &               .90 &               .84 &                    .95 &                    .92 &               .96 &               .94 &                       .89 &                       .80 \\
    \hline
    Blur    &   .85&.76     &    .85&.76    &    \textbf{.90}&.67    &    .85&\textbf{.74}  \\
    \hline
   
   Gaussian Noise &               .64 &               .42 &                    .45 &                    .12 &               .74 &               .57 &                       .70 &                       .46 \\
       Shot Noise &               .71 &               .53 &                    .61 &                    .38 &               .88 &               .81 &                       .86 &                       .75 \\
    Impulse Noise &               .60 &               .36 &                    .43 &                    .08 &               .72 &               .54 &                       .68 &                       .43 \\
    Speckle Noise &               .64 &               .43 &                    .56 &                    .30 &               .87 &               .78 &                       .83 &                       .70 \\
    \hline
    Noise &   .63&.40 & .51&.22 & \textbf{.80}&\textbf{.67} & .78&.59 \\
    \hline
             JPEG &               .74 &               .58 &                    .65 &                    .44 &               .86 &               .76 &                       .85 &                       .74 \\
            MPEG1 &               .72 &               .54 &                    .73 &                    .57 &               .90 &               .84 &                       .88 &                       .78 \\
            MPEG2 &               .63 &               .41 &                    .62 &                    .39 &               .84 &               .73 &                       .82 &                       .67 \\
    \hline
    Digital  & .69&.51 & .67&.48 & \textbf{.86}&\textbf{.78} & .85&.73 \\
    \hline
           Rotate &               .84 &               .74 &                    .85 &                    .76 &               .83 &               .72 &                       .84 &                       .71 \\
       Var Rotate &               .56 &               .30 &                    .44 &                    .10 &               .63 &               .39 &                       .85 &                       .73 \\

        Translate &               .95 &               .92 &                    .96 &                    .93 &               .96 &               .93 &                       .96 &                       .92 \\
      
      \hline
      Camera & .78&.65    &      .74&.59    & .80&.67  &     \textbf{.88}&\textbf{.78} \\ 
      \hline
      
         Sampling &               .75 &               .60 &                    .78 &                    .64 &               .93 &               .88 &                       .99 &                       .98 \\
 Reverse Sampling &               .52 &               .23 &                    .51 &                    .22 &               .55 &               .26 &                       .70 &                       .46 \\
           Jumble &               .73 &               .56 &                    .74 &                    .58 &               .93 &               .88 &                       .99 &                       .98 \\
       Box Jumble &               .86 &               .77 &                    .68 &                    .48 &               .64 &               .40 &                       .85 &                       .74 \\
           Freeze &               .65 &               .43 &                    .68 &                    .49 &               .79 &               .65 &                       .86 &                       .76 \\
       \hline
       Temporal & .69&.50 & .68&.48 & .77 & .61 & \textbf{.89}&\textbf{.78} \\     
\hline
\end{tabular}
\end{center}
\caption{Absolute and relative robustness scores averaged across all severity levels for all categories of perturbations for SSv2P dataset. For each category an average is also shown at the end of all sub-categories. The best models are marked as BOLD for both relative and absolute robustness for each perturbation and their categories. The TimeSformer architecture shows significantly higher robustness scores as compared to the CNN-based architectures while the more appearance based perturbations there is variation between the X3D and Timesformer architecture.}
\label{table_ssv2}
\end{table*}

\begin{table*}[]
\centering
\resizebox{\textwidth}{!}{
\begin{tabular}{|l|l|l|l|l|l|l|l|l|l|l|l|l|l|l|l|l|l|l|l|l|l|l|}
\hline
             & \multicolumn{2}{l|}{R3D-S} & \multicolumn{2}{l|}{R3D-P} & \multicolumn{2}{l|}{I3D-S} & \multicolumn{2}{l|}{I3D-P} & \multicolumn{2}{l|}{SF-S} & \multicolumn{2}{l|}{SF-P} & \multicolumn{2}{l|}{X3D-S} & \multicolumn{2}{l|}{X3D-P} & \multicolumn{2}{l|}{MViT-S} & \multicolumn{2}{l|}{MViT-P} & \multicolumn{2}{l|}{Times-P} \\
             \hline
Defocus Blur &   .63 & .37 & .53 & .43 & .74 & .56 & .54 & .42 & .72 & .61 & .55 & .47 & .84 & .73 & .60 & .51 & .74 & .63 & .80 & .79 & .79 & .78\\
Motion Blur  &    .74 & .56 & .63 & .55 & .75 & .57 & .68 & .60 & .65 & .51 & .57 & .50 & .82 & .69 & .62 & .54 & .71 & .58 & .73 & .62 & .88 & .88\\
Zoom Blur   &  .77 & .61 & .63 & .68 & .79 & .64 & .75 & .68 & .82 & .74 & .83 & .80 & .94 & .89 & .84 & .81 & .88 & .82 & .88 & .87 & .84 & .83\\
\hline
Blur         & .71 & .51   & .63 & .55    & .76 & .59    & .66 & .57    & .73 & .62     & .65 & .59     & \abs{\textbf{.86}} & \rel{\underline{.76}}     & .69 & .62    & .78 & .67    & .81 & .79    & \abs{\underline{.84}} & \rel{\textbf{.83 }}    \\

\hline
Gaussian     &     .57 & .28 & .51 & .41 & .64 & .39 & .61 & .50 & .52 & .33 & .47 & .38 & .59 & .30 & .33 & .18 & .78 & .68 & .80 & .79 & .62 & .61\\
Shot         & .75 & .58 & .79 & .74 & .83 & .71 & .83 & .78 & .73 & .63 & .76 & .72 & .78 & .62 & .56 & .46 & .94 & .91 & .96 & .96 & .92 & .92 \\
Impulse      &  .55 & .23 & .45 & .33 & .59 & .31 & .57 & .45 & .57 & .25 & .43 & .33 & .58 & .28 & .31 & .16 & .76 & .65 & .81 & .80 & .60 & .59\\
Speckle      & .72 & .52 & .75 & .70 & .79 & .63 & .80 & .75 & .70 & .57 & .71 & .66 & .74 & .55 & .54 & .42 & .91 & .87 & .96 & .95 & .90 & .90 \\
\hline
Noise       & .65 & .40   & .62 & .55   & .71 & .51   & .70 & .62   & .60 & .45   & .59 & .52   & .67 & .44   & .43 & .31   & \abs{\underline{.85}} & \rel{\underline{.78}}   & \abs{\textbf{.88}} & \rel{\textbf{.87}}   & .76 & .75    \\

\hline
JPEG         &  .97 & .96 & .74 & .69 & .98 & .97 & .78 & .72 & .93 & .90 & .82 & .79 & .97 & .94 & .90 & .88 & .97 & .95 & .90 & .88 & .91 & .91\\
MPEG1        & .89 & .82 & .70 & .64 & .91 & .85 & .69 & .61 & .90 & .85 & .72 & .67 & .96 & .93 & .88 & .86 & .95 & .93 & .84 & .83 & .93 & .93 \\
MPEG2        & .88 & .80 & .70 & .63 & .91 & .84 & .66 & .57 & .88 & .84 & .71 & .65 & .96 & .84 & .87 & .85 & .95 & .93 & .80 & .79 & .91 & .91 \\
\hline
Digital     &   .92 & .86    & .71 & .65   & .93 & .89   & .71 & .63   & .90 & .86   & .75 & .70 &   .96 & .93 &   .89 & .86 &   \abs{\textbf{.96}} & \rel{\textbf{.94}} & .84 & .83 & \abs{\underline{.92}} & \rel{\underline{.91}}   \\

\hline
Rotate       &    .75 & .57 & .67 & .61 & .76 & .59 & .63 & .53 & .63 & .62 & .68 & .63 & .77 & .61 & .68 & .60 & .80 & .71 & .89 & .88 & .88 & .88 \\
Var Rotate     &      .88 & .79 & .85 & .82 & .85 & .74 & .81 & .76 & .57 & .39 & .63 & .57 & .75 & .58 & .76 & .71 & .63 & .45 & .87 & .71 & .97 & .97 \\
Translate     &  .92 & .87 & .87 & .84 & .88 & .8 & .88 & .84 & .67 & .53 & .70 & .65 & .82 & .70 & .87 & .84 & .65 & .48 & .92 & .93 & .98 & .98 \\
\hline
Camera   &  .85 & .74   & .80 & .76   & .83 & .71   & .77 & .71   & .65 & .51   & .67 & .62   & .78 & .63   & .77 & .72   & .69 & .55   & \abs{\underline{.90}} & \rel{\underline{.89}}   & \abs{\textbf{.95}} & \rel{\textbf{.95}}          \\

\hline
Sampling     &   .94 & .90 & .95 & .94 & .94 & .90 & .95 & .94 & .86 & .81 & .87 & .85 & .92 & .86 & .91 & .89 & .80 & .72 & .93 & .93 & .98 & .98 \\
Reversing    & .94 & .90 & .95 & .94 & .94 & .9 & .95 & .94 & .86 & .81 & .87 & .85 & .92 & .86 & .91 & .89 & .80 & .72 & .93 & .93 & .98 & .98 \\
Jumbling    & .96 & .94 & .97 & .96 & .96 & .93 & .97 & .97 & .78 & .70 & .84 & .81 & .93 & .88 & .92 & .91 & .80 & .70 & .94 & .93 & .98 & .98 \\
Box Jumbl    & .98 & .96 & .98 & .98 & .98 & .95 & .98 & .97 & .95 & .92 & .96 & .95 & .98 & .94 & .98 & .97 & .95 & .91 & .98 & .96 & .97 & .97 \\
Freezing     &  .96 & .91 & .95 & .93 & .95 & .91 & .97 & .96 & .91 & .87 & .92 & .96 & .96 & .93 & .91 & .89 & .79 & .69 & .91 & .89 & .98 & .98 \\
\hline
Temporal   &   .95 & .92 &   .96 & .95 &   .95 & .92   & \abs{\underline{.97}} & \rel{\underline{.96}}   & .86 & .81    & .89 & .87    & .94 & .90    & .93 & .91    & .81 & .73    & .94 & .93    & \abs{\textbf{.98}} & \rel{\textbf{.98}}      \\ 
\hline
\end{tabular}}
\caption{Performance of all the models for various perturbation categories when trained from scratch and using pre-trained weights on UCF-101P dataset. S indicates training from scratch and P indicates using pre-trained weights. We use pre-trained weights from Kinetics-400 for all the models. Red values are for the absolute robustness while blue values are for relative robustness. Bold values are the best models while underlined are the second best models.}
\label{table_ucf}
\end{table*}



\begin{figure*}[t!]
    \centering
    \includegraphics[width=.99\linewidth]{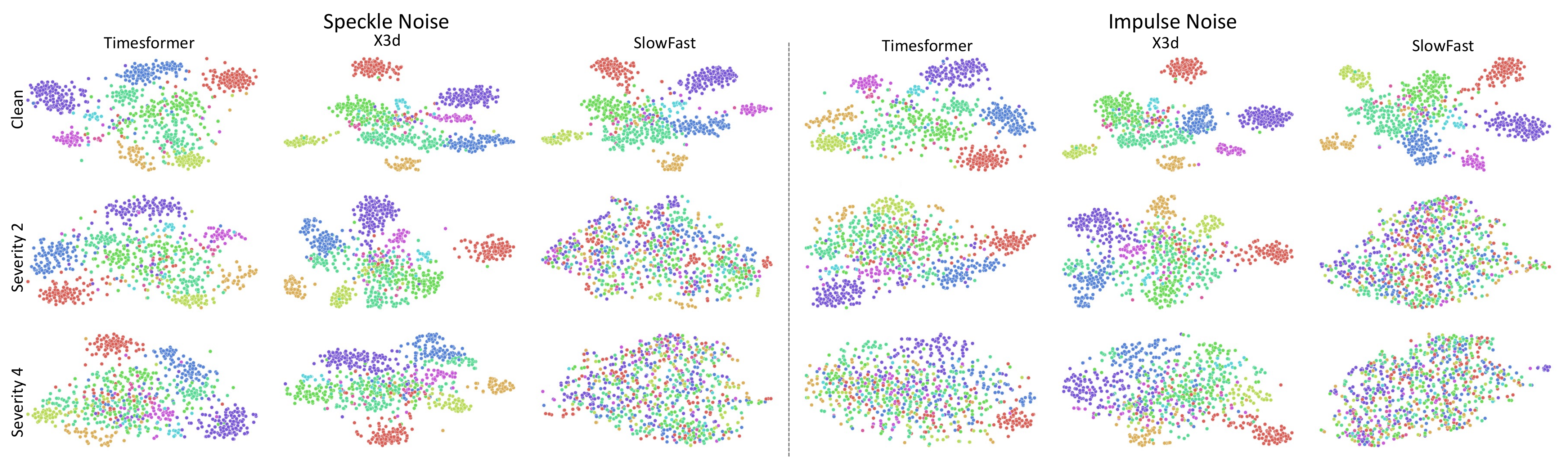}
    \caption{A comparison of embeddings for speckle noise and impulse noise for Timesformer, X3D, and Slowfast on the SSv2 dataset. The first row shows clusters at with no perturbation, the second row shows severity 2 and the third row shows severity 4. As the severity increases, the three models shown consistently decrease in their ability to form distinct clusters.}
    \label{fig:ssv2_tsne_noise}
\end{figure*}

Similarly, Figure \ref{supp:fig:acc_sev_ssv2} shows the performance of four models with different perturbations at varying levels for SSv2P. We observe that the transformer based model, Timesformer, is generally more robust as severity level increases for the temporal perturbations. For the appearance based perturbations that are blur and noise related, all models drop in performance significantly. Figure \ref{fig:ssv2_tsne_noise} further visualizes the decrease in performance for appearance based perturbations. We observe the degradation of distinct clusters for the SSv2 dataset over increased severity for noise perturbations for all three models.

\subsection{Absolute and relative robustness}
\label{sec:absolute_relative_robustness}

\begin{figure*}
    \centering
    \includegraphics[width=.99\linewidth]{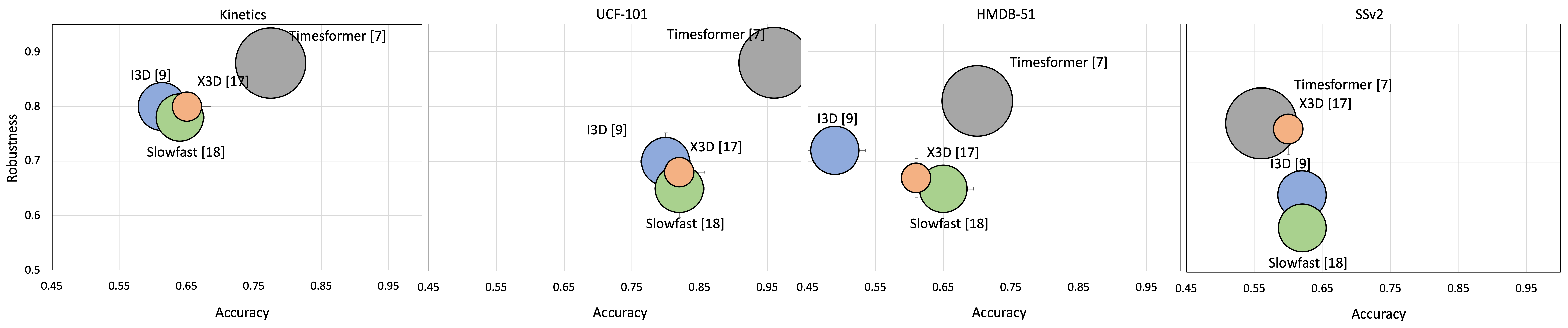}
    \caption{A performance and robustness visualization of pre-trained action recognition models on Kinetics-400P, UCF-101P, HMDB-51P, and SSv2P. The y-axis: relative robustness $\gamma^r$ (higher is better), x-axis: accuracy on clean videos, and the size of circle indicates FLOPs. The transformer based model is consistently the most robust model across datasets but at the expense of a higher number of FLOPs.}
    \label{fig:teaser_figures}
\end{figure*}

Table \ref{table_kinetics}, Table \ref{table_ucf}, Table \ref{table_hmdb}, and Table \ref{table_ssv2} show the robustness scores for all the perturbations for all models on Kinetics-400P, UCF-101P, HMDB-51P and SSv2P respectively. In addition, Table \ref{table_ucf}, and Table \ref{table_hmdb} also show the robustness score for both pre-trained and training from scratch performance where the pre-trained weights are taken from Kinetics-400 pre-training. In Table \ref{table_kinetics}, we observe that the transformer based models are generally more robust than CNN counterparts where MViT performs the best on Blur, Noise, and Digital perturbations, Timesformer performs the best on Camera motion perturbations. We also observe that R3D based model performs best on Temporal perturbations but the margin is very small when compared with other models. 

In Table \ref{table_ucf}, we observe a similar behavior, where transformer based models are the best performers for all the perturbations when pre-training is used. However, we observe that when pre-training is not utilized, CNN models are better performers for Blur, Camera motion, and Temporal perturbations. 

In Table \ref{table_ssv2} we observe that while the transformer based model outperforms in temporal and camera perturbations, it does not in the other appearance based categories. The transformer based model however significantly outperforms the other models in temporal perturbations while it is less significantly outperformed by other models in blur, noise, and digital. This emphasizes how important time is for the SSv2 dataset and how well transformer-based models learn temporally relevant features.

\subsection{Pretraining vs Scratch}
\label{sec:pretraining_vs_scratch}
\begin{figure*}[t!]
\centering
\begin{center}
\centering
\includegraphics[width=.99\linewidth]{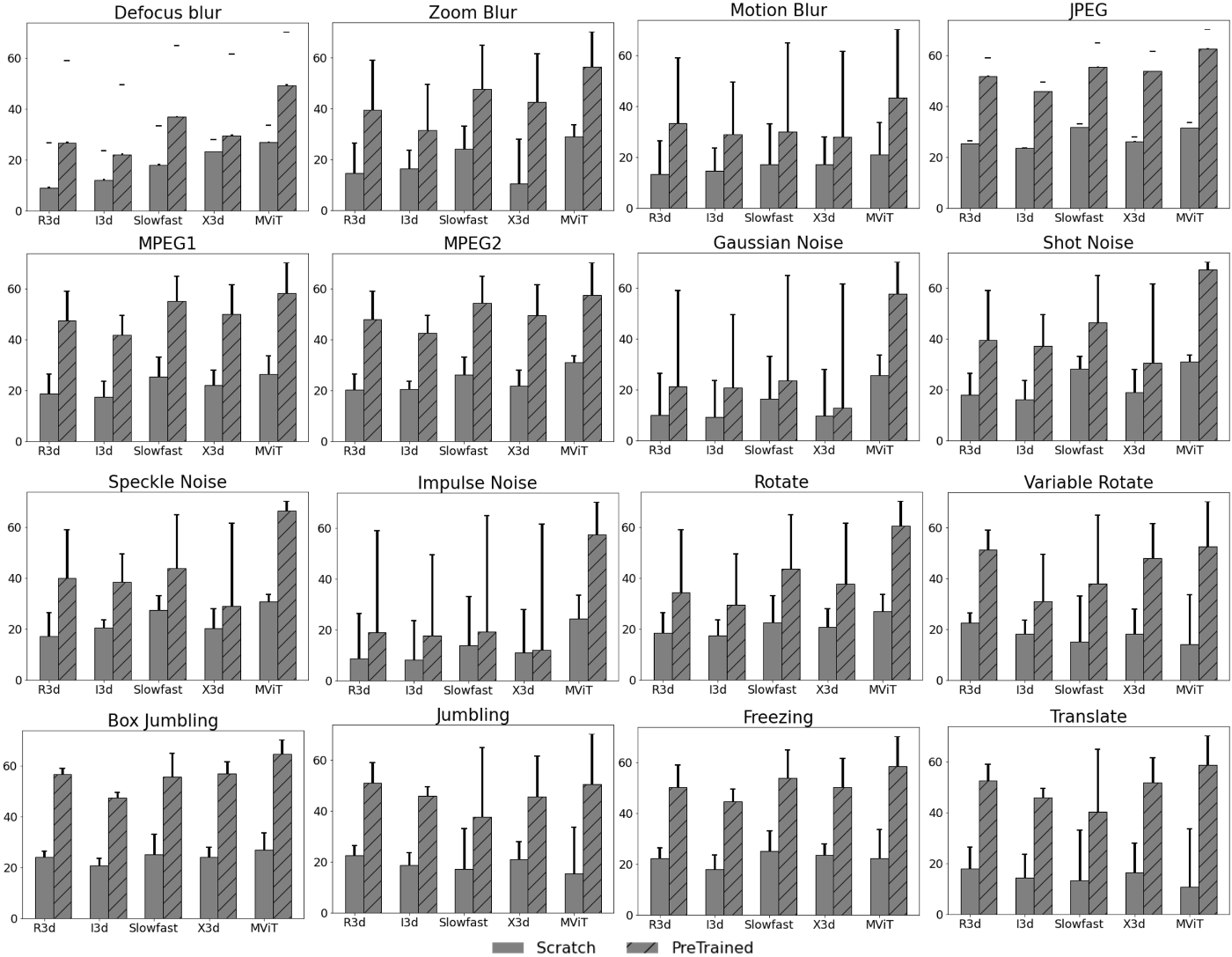}
\end{center}
  \caption{
  A comparison of model robustness against different perturbations with pretrained and scratch training on HMDB-51P benchmark. The plain bar represents performance without pretrained weights and striped bar represents a pretrained model. The top extension indicates drop in performance in comparison with accuracy on clean videos. 
  }
\label{fig:pretrain_scratch_hmdb}
\end{figure*}

Figure \ref{fig:pretrain_scratch_hmdb} shows a comparison between pre-training and scratch performance for all the models across various perturbation categories on HMDB-51P dataset. We observe that although MViT is more robust against various perturbations when pre-training is used, its performance drops significantly when pre-training is not utilized. This behavior is similar to what we observe on UCF-101P dataset. More notably, we observe that for Translate and Variable rotation perturbation, the performance of MViT is worse than all the other CNN counterparts, despite the fact that it outperforms all those models when clean videos are used.

\subsection{SSv2 Class Analysis}
\label{sec:ssv2_class_analysis}

\begin{figure*}[t!]
    \centering
    \includegraphics[width=.99\linewidth]{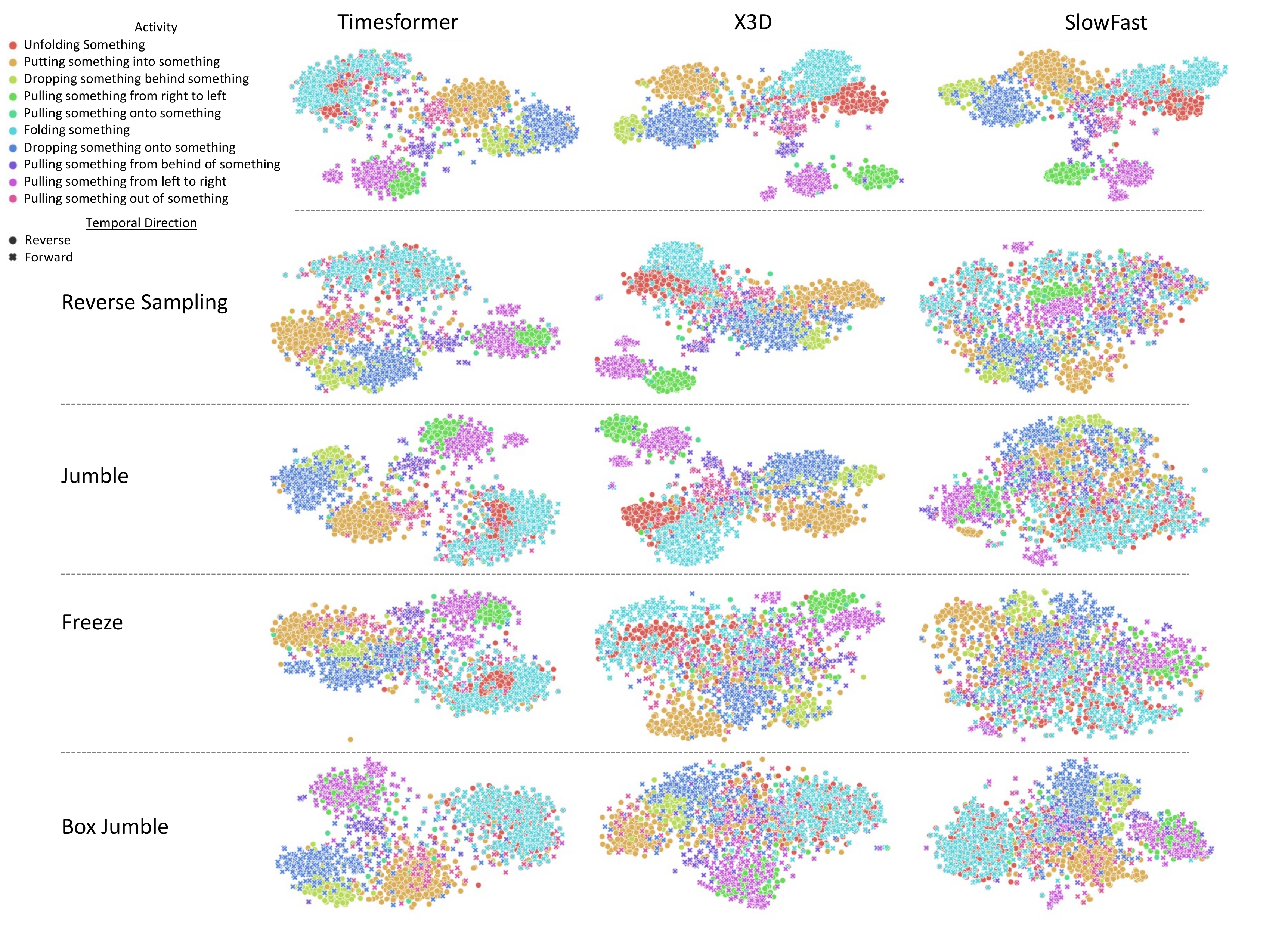}
    \caption{The embedding space for a sample of models on SSv2 for five classes and their respective opposite based on the direction of time. The first row is without any perturbations while the remainder are reverse sampling, jumble and freeze at severity 4. }
    \label{fig:ssv2_tsne_temporal}
\end{figure*}

Figure \ref{fig:ssv2_tsne_temporal} shows additional perturbations at severity 4 with five classes and their respective opposites. While all models struggle to significantly separate classes that are equivalent in all but direction, the Timesformer is noticeable better at maintaining clusters when at higher severities of temporal perturbations. While X3D shows better distinction of clusters from the start, as the severity increases, these clusters overlap more and more. The worse performing CNN example is unable to maintain clusters for all the temporal perturbations at increased severity. 

\begin{figure*}[t!]
    \centering
    \includegraphics[width=.8\linewidth]{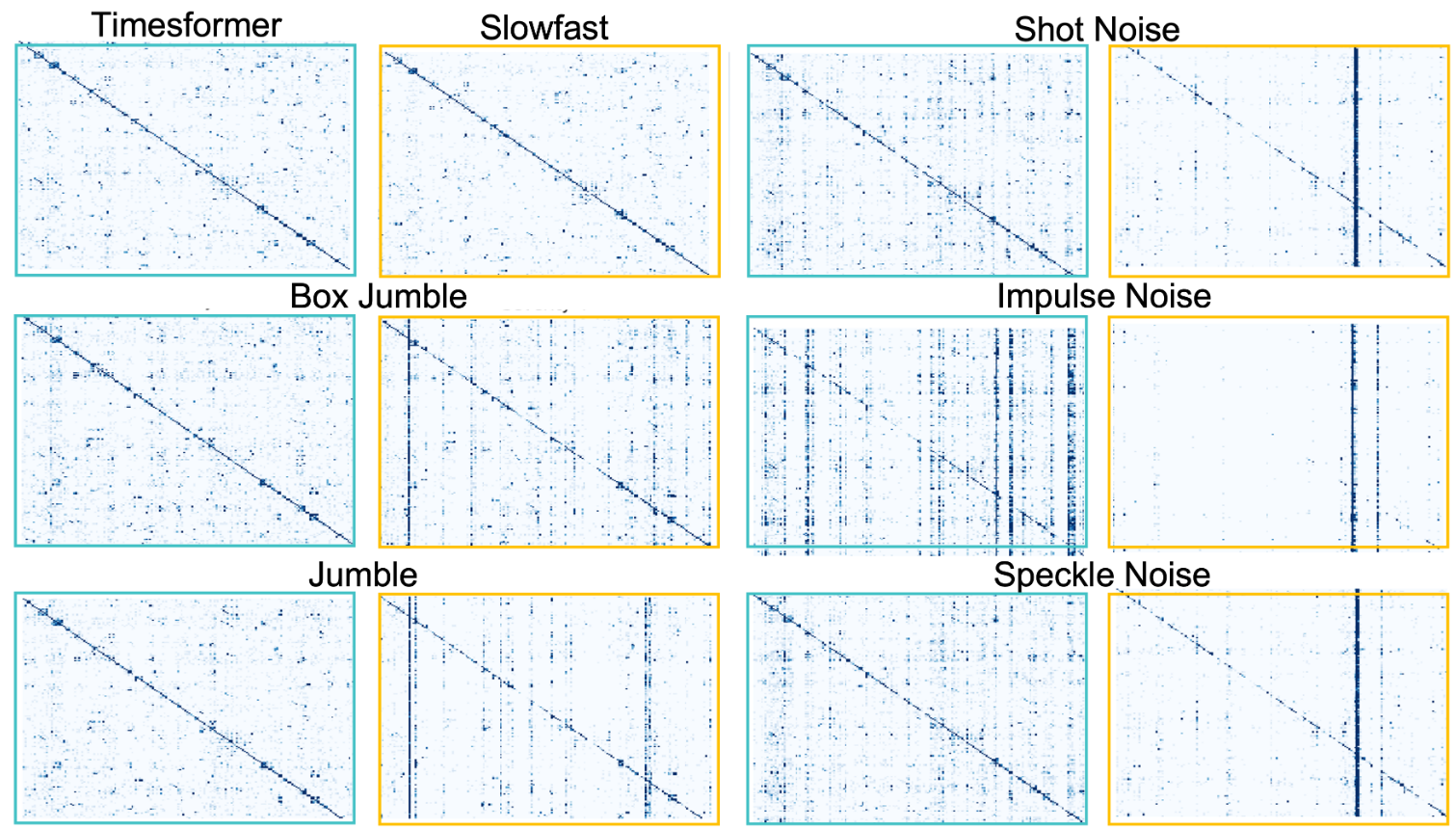}
    \caption{Confusion Matrices for the SSv2 dataset for box jumble, jumble, shot noise, impulse noise, and speckle noise perturbations at severity 4. When models fail, they are often predicting a smaller selection of classes for a majority of the samples. These classes also differ between whether the perturbations are appearance based or temporal based. }
    \label{fig:ssv2_conf}
\end{figure*}

\begin{figure*}
    \centering
    \includegraphics[width=.99\linewidth]{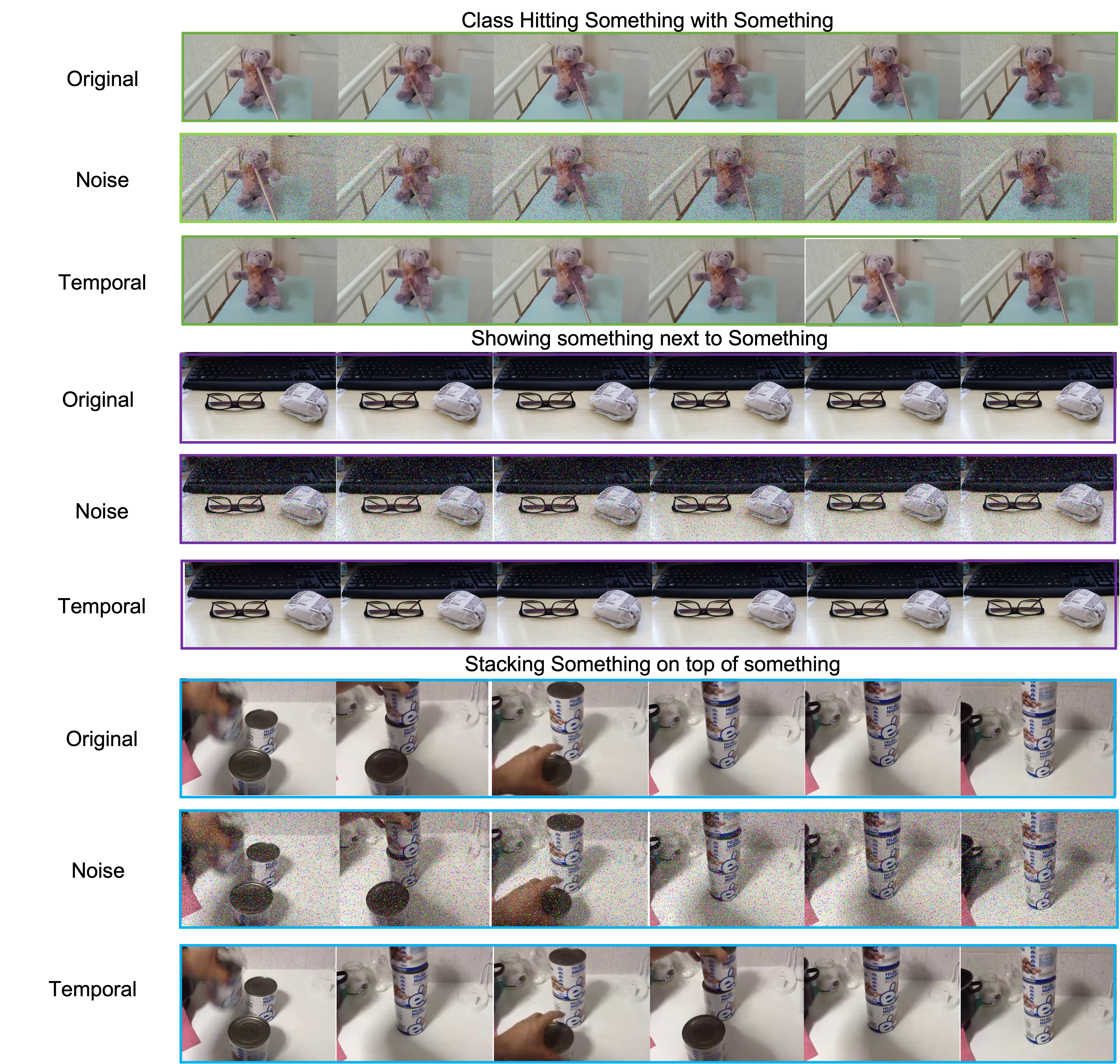}
    \caption{Examples of the most predicted classes for either temporal or noise perturbations. For noise, the CNN based model Slowfast predicts  ``Showing something next to something" for a majority of samples while for temporal perturbations it predicts ``Hitting something with something" and ``Stacking number of something". }
    \label{fig:class_examples}
\end{figure*}

Figure \ref{fig:ssv2_conf} shows additional confusion matrices for Timesformer and Slowfast with box jumble, jumble, shot noise, impulse noise and speckle noise perturbations at severity 4. The Timesformer architecture significantly outperforms Slowfast in class predictions for temporal predictions but suffers similarly with noise. The confusion matrices additionally show that the models are often predicting only a small selection of classes for all samples as shown by the vertical blue bars. This is especially noticeable for Slowfast on the noisy perturbations for the class ``Showing something next to something". The classes most often predicted differ when it is temporal perturbations, in which case for Slowfast the most incorrectly predicted class is ``Hitting something with something" followed by ``Stacking number of something". Sample videos for these three classes are shown in Figure \ref{fig:class_examples} where if there was a noise perturbation, the bottom two videos would be classified as the top video while for noisy pertrubations it would be the reverse. For the temporal confusion, is likely the case because when temporal pertrubations are applied, the actual interaction between the objects is jumbled and could appear to just be one object hitting the other at random points in time. 

\begin{figure*}[t!]
\begin{center}
\includegraphics[width=0.96\linewidth]{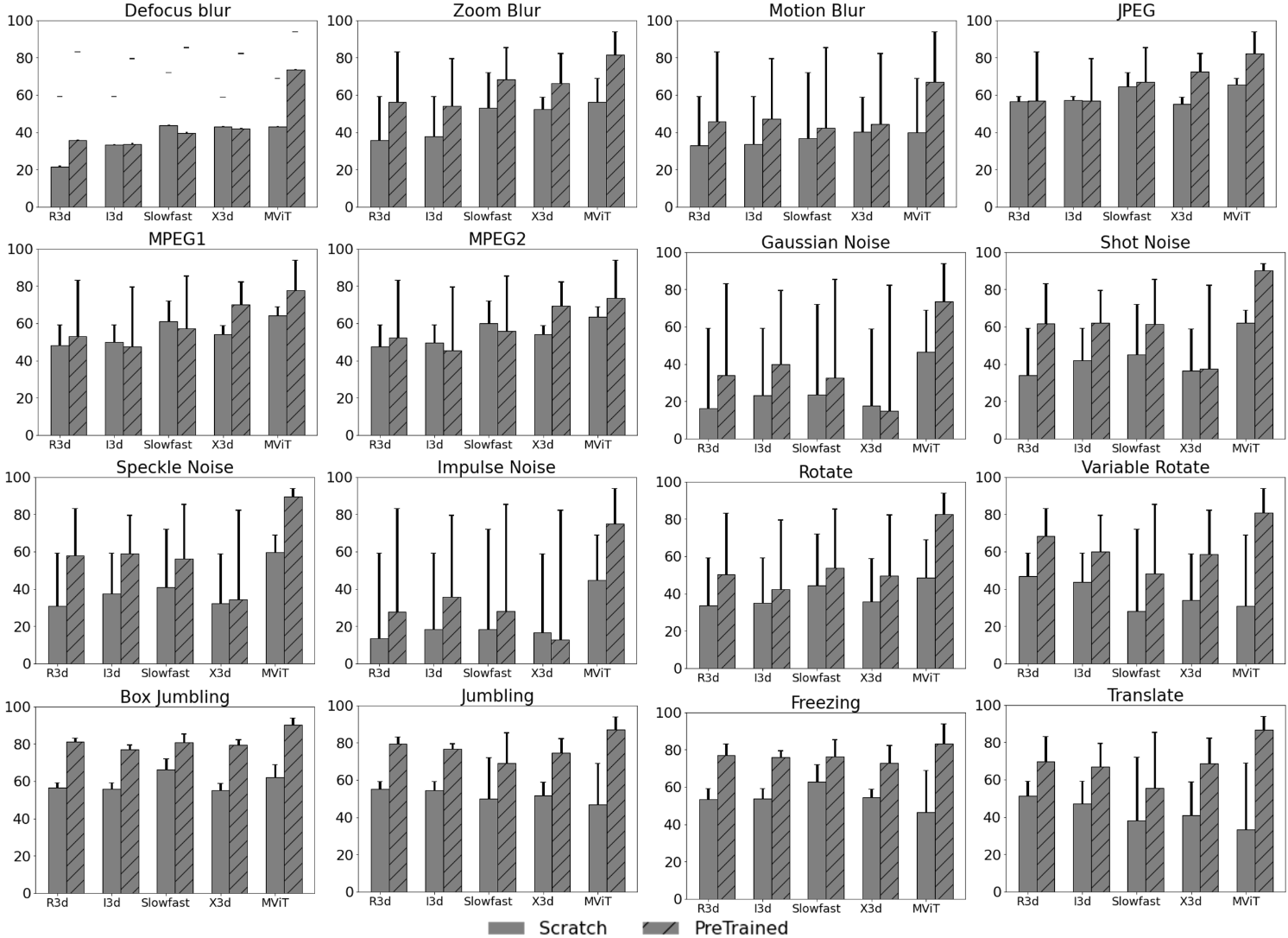}
\end{center}
  \caption{
  A comparison of model robustness against different perturbations with pretrained and scratch training on UCF-101P benchmark. The plain bar represents performance without pretrained weights and striped bar represents a pretrained model. The top extension indicates drop in performance in comparison with accuracy on clean videos. 
  }
\label{fig:pretrain_scratch_ucf}
\end{figure*}

\begin{table*}[]
\centering
\resizebox{\textwidth}{!}{
\begin{tabular}{|l|l|l|l|l|l|l|l|l|l|l|l|l|l|l|l|l|l|l|l|l|l|l|}
\hline
             & \multicolumn{2}{l|}{R3D-S} & \multicolumn{2}{l|}{R3D-P} & \multicolumn{2}{l|}{I3D-S} & \multicolumn{2}{l|}{I3D-P} & \multicolumn{2}{l|}{SF-S} & \multicolumn{2}{l|}{SF-P} & \multicolumn{2}{l|}{X3D-S} & \multicolumn{2}{l|}{X3D-P} & \multicolumn{2}{l|}{MViT-S} & \multicolumn{2}{l|}{MViT-P} & \multicolumn{2}{l|}{Times-P} \\
             \hline
Defocus Blur &    .83         &       .34      &       .95      &       .45      &        .89     &     .52        & .73            &       .45      &  .85           &     .54       &        .82     &    .57        &      .94       &   .82          &  .68           & .48            & .93              &    .81         &   .79           &    .70         &  .81            &       .73      \\

Motion Blur  &     .87        &     .51        &   .75          &  .57           &    .91         & .63            & .80            &        .59     &  .84           &      .53      &.65             &          .46  &       .89      &        .62     &        .67     &          .45   & .87             & .63            &     .73         &    .62         &  .85            &           .79  \\

Zoom Blur    &    .88         &      .56       &   .81          &  .67           &    .93         & .70            &     .82        &        .64     &      .91       &      .73      &    .83         & .74           &      .83       &    .38         &   .81          &  .69           &        .96      &            .87 &    .87          &            .81 &        .75      &         .64    \\

\hline

Blur        &   .86 & .47     &   .84& .56     &       \abs{\textbf{.91}}& .62            &  .78&.56     &       .87&.6  &        .73& .59          &  .89&.61    &             .72&.54   &    \abs{\textbf{.92}}&\rel{\textbf{.77}}        &       .80&.71         &      .80&\rel{\underline{.72}}     \\

\hline
Gaussian     &      .84      &     .38        &       .63      &       .36      &    .86         &     .40        &     .72        &    .42         &  .84           &      .50      &    .59         &           .37 &      .82       &     .35        &   .51          &       .21      &    .92          &        .77     &        .88     &         .83    &    .65          &           .50  \\

Shot         &    .92         &    .68         &   .81          &   .67          &    .93         &         .69    &     .88        &       .76      &      .96       &     .86       &    .82         &    .72        &      .91       &       .68      &       .69      &         .50    &    .98          & .93            &    .97          &        .96     &.86              &           .80  \\

Impulse      &  .83           &     .33        &  .60           &        .32     & .85            &            .35 &          .68   &        .36     &  .81           &  .42          &     .55        &     .30      &   .83          &     .40        &         .51    &        .19     &     .91     & .73            &         .87     &      .82       &             .73 &             .47\\

Speckle      &      .91       &       .66      &       .81      &       .68      &        .97     &         .88    &     .89        &           .78  &  .95           &     .84       &    .79         &        .68    &      .92       &  .73           &   .68          &         .47    &    .97          &.92             &        .97      & .95            &    .85          &            .79 \\

\hline
Noise     &   .90&.51        &        .71&.51          &      .9& .58            &     .79&.58 &           .89&.56      &      .69&.52       &     .87&.54     &           .60&.34    &   \abs{\textbf{.94}}&\rel{\underline{.84}}  &      \abs{\textbf{.92}}&\rel{\textbf{.89}}          &         .75&.64      \\

\hline
JPEG         &     .89        &    .97         &   .93         &       .88      &    .89         &     .97        &     .97        &          .93  &      .99       &     .97      &    .91         &    .86        &  .98           &      .94      &   .93          &       .88      &    .98          &           .94  &        .93      &   .90          &    .89          &             .84\\

MPEG1        &     .93 & .72 & .89 & .81 & .94 & .75 & .93 & .85 & .93 & .78 & .9 & .85 & .94 & .90 & .89 & .82 & .93 & .79 & .88 & .83 & .91 & .87 \\
MPEG2        & .94 & .78 & .89 & .82 & .97 & .89 & .93 & .84 & .93 & .80 & .90 & .84 & .94 & .78 & .88 & .81 & .98 & .93 & .88 & .82 & .9 & .86\\
\hline
Digital      &        .95&.82        &        .90&.84         &      \abs{\textbf{.97}}&\rel{\textbf{.87}}       &         .94&.87     &       .95&.85          &      .90&.85           &      .95&.13        &            .90&.84       &    \abs{\underline{.96}}&\rel{\underline{.89}}       &        .90&.85        &       .90&.86  \\
\hline
Rotate       &  .92 & .71 & .76 & .59 & .94 & .74 & .80 & .60 & .90 & .69 & .79 & .68 & .93 & .75 & .76 & .62 & .93 & .80 & .91 & .87 & .82 & .74 \\
Var Rotate     &    .96 & .87 & .93 & .88 & .95 & .78 & .82 & .63 & .82 & .46 & .73 & .59 & .90 & .65 & .87 & .78 & .81 & .42 & .83 & .75 & .96 & .94 \\
Translate     & .92 & .69 & .94 & .90 & .91 & .61 & .97 & .93 & .81 & .40 & .76 & .62 & .88 & .58 & .91 & .85 & .77 & .32 & .89 & .84 & .96 & .94 \\
\hline
Camera    &       \abs{\textbf{.93}}&.76 & .88&.79 & \abs{\underline{.93}}&.72 & .86&.72 & .84&.52 & .76&.63 & .91&.66 & .85&.82 & .84&.52 & .88&\rel{\underline{.82}} & .91&\rel{\textbf{.87}}       \\

\hline

Sampling     &     .93 & .73 & .89 & .82 & .93 & .70 & .92 & .84 & .84 & .52 & .76 & .63 & .90 & .64 & .85 & .75 & .85 & .55 & .81 & .53 & .97 & .96 \\
Reversing    &    .93 & .73 & .89 & .82 & .93 & .70 & .92 & .84 & .84 & .52 & .76 & .63 & .90 & .64 & .85 & .75 & .85 & .55 & .81 & .53 & .97 & .96\\ 
Jumbling     &   .96 & .87 & .92 & .87 & .96 & .81 & .96 & .93 & .85 & .52 & .74 & .58 & .93 & .76 & .84 & .75 & .82 & .46 & .81 & .72 & .96 & .94 \\
Box Jumbl    &      .98 & .93 & .98 & .97 & .98 & .90 & .98 & .96 & .92 & .76 & .91 & .86 & .96 & .86 & .96 & .93 & .94 & .81 & .95 & .92 & .97 & .96\\
Freezing     &  .96 & .85 & .92 & .86 & .95 & .78 & .95 & .91 & .92 & .77 & .89 & .83 & .96 & .85 & .89 & .82 & .89 & .67 & .89 & .84 & .94 & .91    \\
\hline
Temporal     &      .95&.82 & .92&.87 & .95&.80 & \abs{\underline{.95}}&\rel{\underline{.90}} & .87&.62 & .81&.71 & .93&.75 & .88&.80 & .87&.60 & .85&.70 & \abs{\textbf{.97}}&\rel{\textbf{.95}} \\ 
\hline
\end{tabular}}
\caption{Performance of all the models for various perturbation categories when trained from scratch and using pre-trained weights on HMDB-51P dataset. S indicates training from scratch and P indicates using pre-trained weights. We use pre-trained weights from Kinetics-400 for all the models. Red values are for the absolute robustness while blue values are for relative robustness. Bold values are the best models while underlined are the second best models. 
}
\label{table_hmdb}
\end{table*}

\begin{figure*}
\centering
\begin{center}
\centering
\includegraphics[width=.75\linewidth]{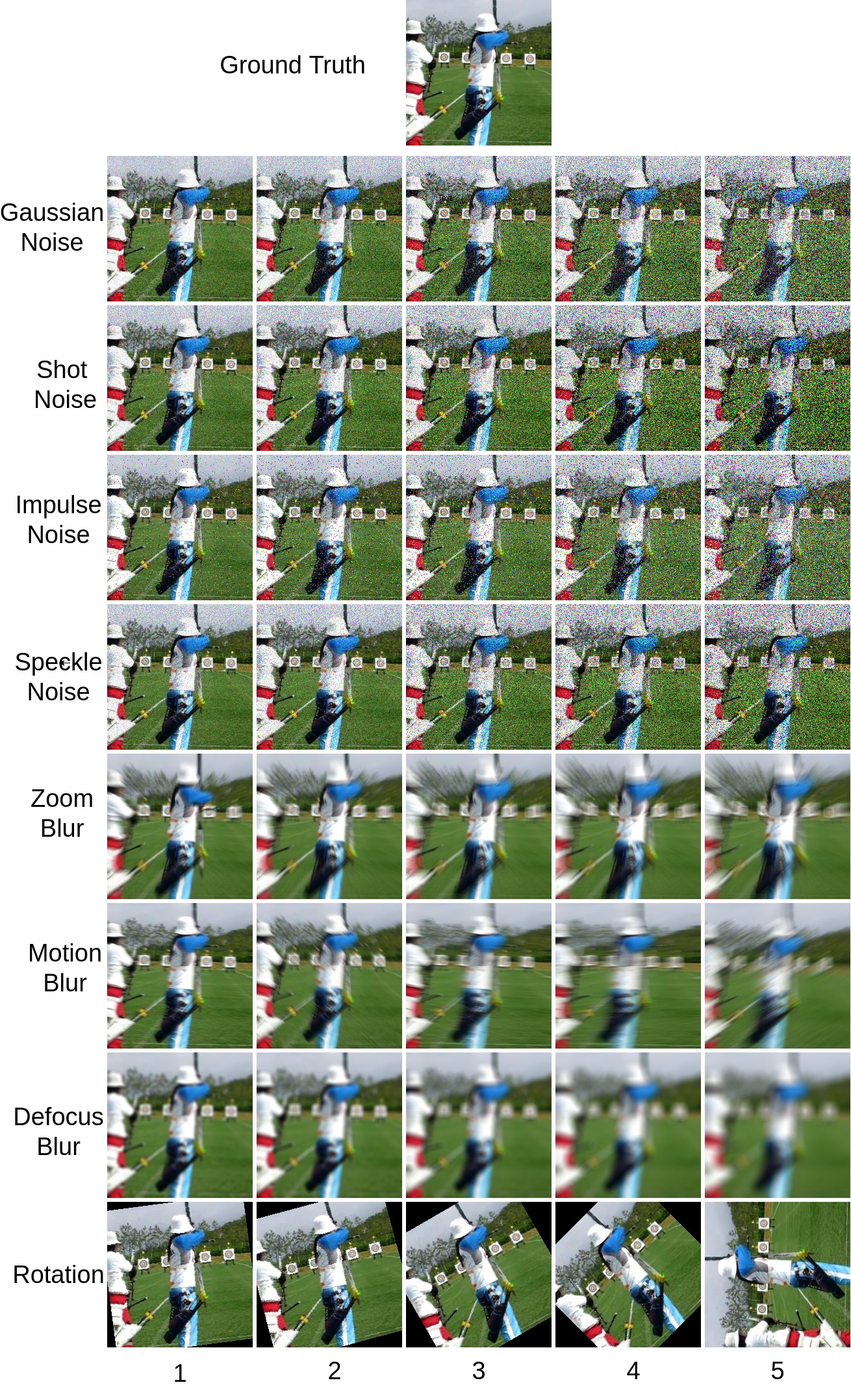}
\end{center}
  \caption{
 Sample video frames from Kinetics-400P showing different severity levels of some spatial perturbations(severity increases from left to right) .  
  }
\label{fig:samlple_kin_sev}
\end{figure*}

\begin{figure*}[t!]
\centering
\begin{center}
\centering
\includegraphics[width=.9\linewidth]{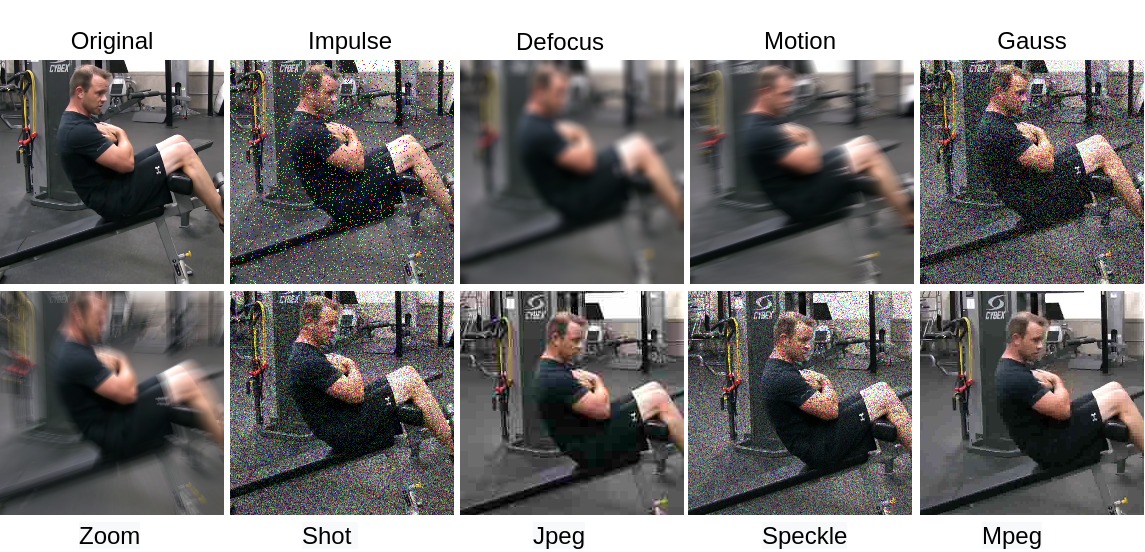}
\end{center}
  \caption{
 Sample video frame from Kinetics-400P showing different perturbations using blur and noise.  
  }
\label{fig:samlple_kin}
\end{figure*}


\section{Implementation details}
\label{sec:implementation_details}
In Figure \ref{fig:samlple_kin_sev}, we show some sample video frames from Kinetics-400P showing different severity levels for some of the perturbations. We can see clearly that the level of perturbations increase as we move from 1 to 5. More sample videos are available in the project webpage: \url{bit.ly/3TJLMUF}.

The implementations of the various perturbations is provided in the project page. Following are brief details of severity levels for the perturbations:

\paragraph{Noise Perturbations:}
For gaussian noise,  we increase the standard deviation of the gaussian distribution(from where we sample noise which is added) and similarly for speckle noise, we increase the standard deviation of the gaussian distribution(from where we sample noise which is multiplied with pixelwise itensities and then added) as we increase the severity levels. 
For impulse noise, we increase the proportion of image  pixels to be replaced in the original frame with noise . For shot noise, we increase the proportion of image  pixels to be replaced in the original frame with noise as we increase the severity levels. 

\paragraph{Blur Perturbations:}
In defocus blur, we increase the radius of the disk which is convolved over the image to create defocus blurring effect. In motion blur, we increase the radius and sigma of the kernel which is used to create the motion blurring effect. In zoom blur, we increase  as we increase the severity levels. 

\paragraph{Digital Perturbations:}
For JPEG, MPEG1 and MPEG2 the amount of compression is increased as we increase the severity levels.

\paragraph{Camera Perturbations:}
For rotation, we rotate each frame by angle and the angle increases over the severity levels. For random rotation, each frame  is rotated by a random angle from a range , which is increased as we increase the severity level. 
For translation, we randomly choose the center while cropping the image from 256x256 to 224x224 resolution,

\paragraph{Temporal Perturbations:}
For jumbling, we divide video into segments and randomly shuffle frames of those segments, the segment size increases from 4 to 64 from level 1 to 5. For box jumbling, the divided segments are jumbled, the segment size decreases from 64 to 4 as we increase the severity levels. For freezing, we increase the threshold value below which we freeze the frame(keep the previous frame for that index) as we increase the severity levels. For implementation of these, we save the perturbed frame indices and use that while loading the data for the model.

For implementation of model evaluations and to get pretrained weights for the models, we used the open source video understanding codebase PySlowfast \footnote{\url{https://github.com/facebookresearch/SlowFast}} . We also used the code provided by them to finetune models on UCF101 and HMDB51 dataset. Also various examples sample videos of Kinetics-400P are provided in the project webpage. Figure \ref{fig:samlple_kin} shows sample video frame from Kinetics-400P showing different perturbations using blur and noise. \\

\subsection{Training on Perturbations}
\label{training_perturbations}
We trained one CNN-based model, ResNet50, and one transformer-based model, MViT. Both models are pre-trained on the Kinetics400 dataset. 
In order to understand how training on different type of perturbations may impact overall performance, we train the ResNet50 and MViT model on temporal, spatial, mixed and the state-of-the-art PixMix \cite{hendrycks2022pixmix}. Originally from the image-domain, PixMix adds augmentations by mixing a given image with diverse patterns from fractals and feature visualizations. For training on temporal, a perturbation is randomly selected from \textit{jumble}, \textit{freeze}, and \textit{sampling}. When evaluation on the temporal category, the perturbations are randomly chosen from \textit{jumble}, \textit{freeze}, \textit{sampling}, \textit{box jumble}, and \textit{reverse sampling}. For training on spatial, a perturbation is randomly selected from \textit{speckle noise}, \textit{gaussian noise}, and \textit{rotate}. For testing on spatial, a pertubation is randomly selected from \textit{shot noise}, \textit{static rotate}, \textit{translate}, and \textit{impulse noise}. For mixed, a perturbation category is first selected then a perturbation type from that category. For temporal, spatial and mixed during training, severities are chosen at random between 1,2, and 3 for training and  4 or 5 for testing. For PixMix \cite{hendrycks2022pixmix}, we apply the augmentation at severity 3 for each frame individually, in which a different fractal image is chosen for each. We trained one CNN-based model, ResNet50, and one transformer-based model, MViT. Both models are pre-trained on the Kinetics400 dataset. 

\section{UCF101-DS dataset}

\begin{figure}
    \centering
    \includegraphics[width=.8\linewidth]{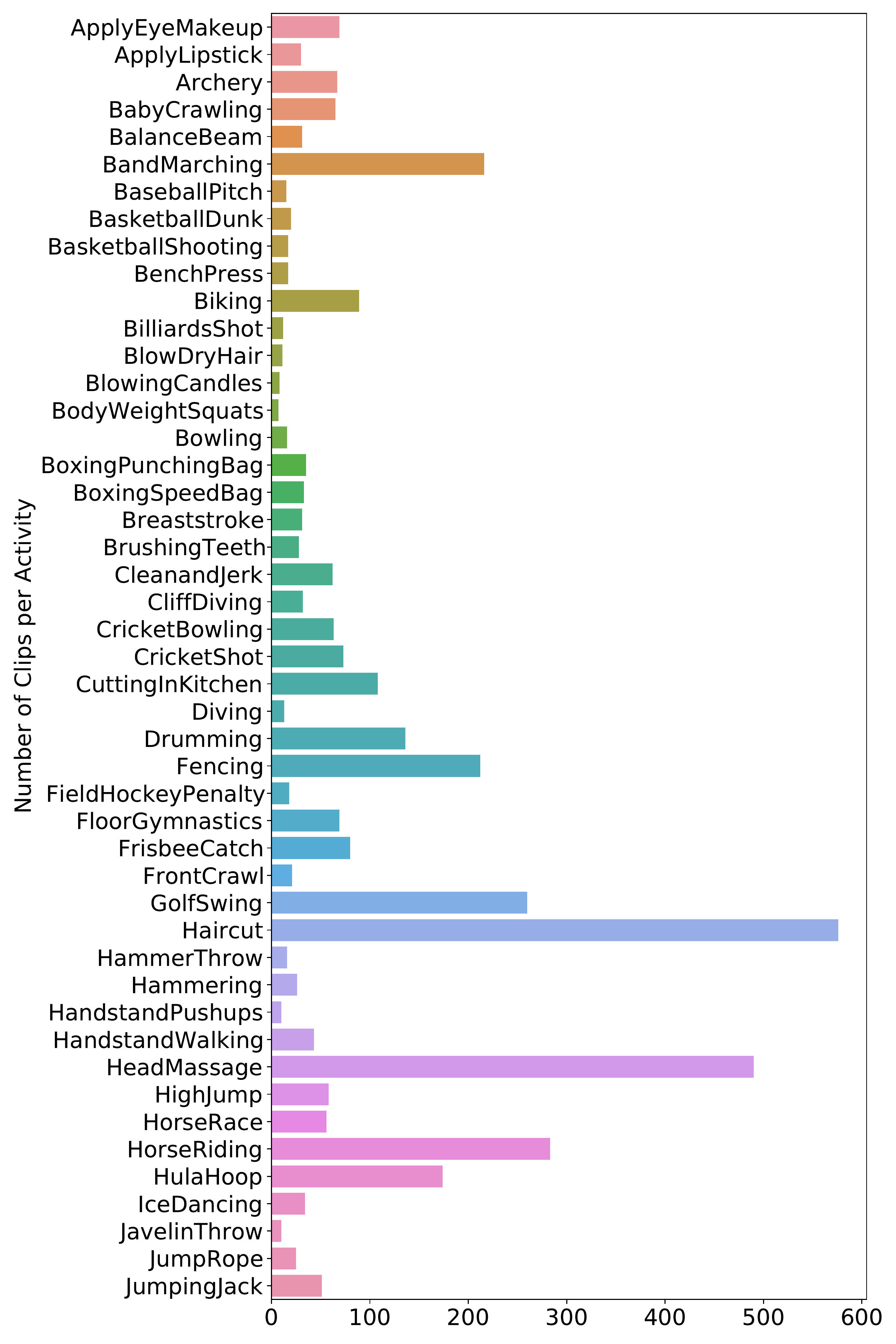}
    \caption{The number of clips per UCF101 activity.}
    \label{fig:clips_per_action}
\end{figure}

\begin{figure}
    \centering
    \includegraphics[width=.6\linewidth]{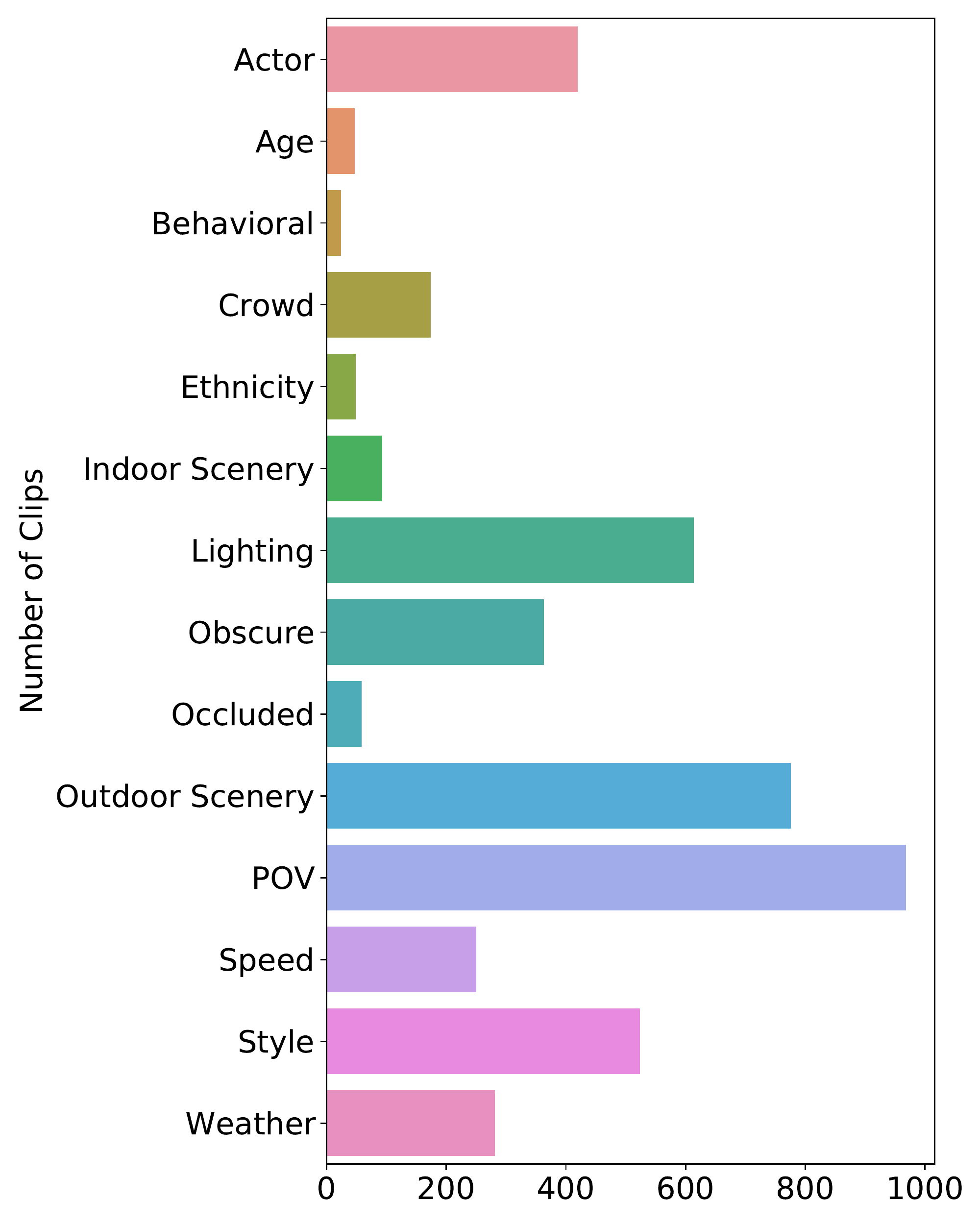}
    \caption{The number of clips per distribution shift category.}
    \label{fig:clips_per_category}
\end{figure}

The UCF101-DS\footnote{For more information and to download this dataset, visit \url{https://www.crcv.ucf.edu/research/projects/ucf101-ds-action-recognition-for-real-world-distribution-shifts/}} dataset consists of distribution shifts for 47 classes (see Figure \ref{fig:clips_per_action}) with 63 different distribution shifts that can be categorized into 15 categories (see Figure \ref{fig:clips_per_category}). Table \ref{tab:category_mappings} shows the defined mappings and the number of clips for each category. A total of 536 unique videos were collected from YouTube and split into a total of 4,708 clips. While there are many clips per some videos, we do confirm that models will give different results for each clip for a long video.   For example on video-id $v4TFEL3lPhg$, models X3D and R3D correctly classify ``HighJump`` 2, MViT 14 and ResNet50 9 of 31 clips. Another example of a longer video $oEm64FFEKnc$, the MViT model classifies the activity ``Haircut'' correctly for 32 of the 76 clips while ResNet50 classified 18. 

\begin{table*}
\caption{A summary of the distribution shifts we used to collect videos for UCF101-DS and their corresponding high-level category. }
\label{tab:category_mappings}
\resizebox{\linewidth}{!}{\begin{tabular}{l|p{18cm}|l}
\hline
       Category &                                                                                        Distribution Shift &  Number of Clips \\
\hline
                   Actor &                                                                                    [animal, costume, toy] &    420 \\
            Age &                                                                                        [kids, old\_person] &     47 \\
     Behavioral &                                                                  [caught\_on\_cam!, prank, reaction, scary] &     24 \\
          Crowd &                                                                                                   [crowd] &    176 \\
      Ethnicity &                                                                     [african, asian, black, indian\_brown] &     50 \\
 Indoor Scenery &                          [at\_home, at\_the\_club, at\_the\_gym, indoor, indoors, in\_court, in\_garage, mirror] &     96 \\
       Lighting &                                          [low\_light, at\_late\_night, at\_night, dark, low\_light\_conditions] &    616 \\
        Obscure &                                                                                         [unsual, unusual] &    378 \\
       Occluded &                                                                             [obstructed, obstructed\_view] &     59 \\
Outdoor Scenery & [at\_the\_beach, desert, in\_backyard, in\_garden, in\_the\_fields, on\_the\_road, outdoors, outside, underwater] &    777 \\
            POV &             [camera\_angle, camera\_angles, go\_pro, on\_TV, pov, pov\_at\_night, shaky, tutorial, upside\_down] &    992 \\
          Speed &                                                                   [alow\_mo, fastest, slowmotion, slow\_mo] &    251 \\
          Style &                                                    [animated, animation, filter, text\_on\_screen, vintage] &    535 \\
        Weather &                                                                         [fog, in\_rain, muddy, rain, snow] &    287 \\
\bottomrule
\end{tabular}}
\end{table*}

At most two distribution-shift specific search terms are concatenated to the class names at random to form a search query, which is then used to search YouTube nd retrieve all the search results. Miscellaneous search terms such as ``prank", ``reaction", ``unusual", etc have also been added at random to the search queries. These search results are then filtered to only download the videos with length less than 60 seconds and height and width dimensions of at least 256x256. 
These videos have been manually analysed and cleaned
These videos are then trimmed into smaller videos that are less than 10 seconds each. Additional caution has been taken by manually verifying each video to consist the ground truth data. 

In Figure \ref{fig:all_models_ucf101DS}, we have presented the performance of different models trained with and without augmentation on this real-world dataset. We observe that while CNN models benefits from augmentations, transformer based model MViT trained on clean videos performs the best, not showing any benefits from data augmentations. 

\end{document}